\documentclass[10pt,twocolumn,letterpaper]{article}
\usepackage{subfiles}

\usepackage{xr-hyper}       
\usepackage{xr}
\makeatletter
\newcommand*{\addFileDependency}[1]{
  \typeout{(#1)}
  \@addtofilelist{#1}
  \IfFileExists{#1}{}{\typeout{No file #1.}}
}
\makeatother

\newcommand*{\myexternaldocument}[1]{%
    \externaldocument{#1}%
    \addFileDependency{#1.tex}%
    \addFileDependency{#1.aux}%
}
\myexternaldocument{appendix_2425}

\newcommand{\myeqref}[1]{Eq.~(\ref{#1})}

\usepackage{iccv}
\usepackage{times}
\usepackage{epsfig}
\usepackage{graphicx}
\usepackage{amsmath}
\usepackage{amssymb}
\usepackage{subcaption}
\usepackage{CJKutf8}
\usepackage{booktabs}       
\usepackage{multirow}
\usepackage{wrapfig}
\usepackage{minitoc}
\usepackage{multirow, makecell}
\usepackage[normalem]{ulem}

\usepackage[pagebackref=true,breaklinks=true,letterpaper=true,colorlinks,bookmarks=false]{hyperref}

\iccvfinalcopy 

\def\iccvPaperID{2425} 

\ificcvfinal\fi

\usepackage{amsmath,amsfonts,bm}

\def\eqref#1{equation~\ref{#1}}

\def\1{\bm{1}}

\def\rg{{\textnormal{g}}}

\def\vs{{\bm{s}}}

\def\vx{{\bm{x}}}
\def\vy{{\bm{y}}}
\def\vz{{\bm{z}}}

\def\evs{{s}}

\def\mM{{\bm{M}}}

\def\mV{{\bm{V}}}

\DeclareMathAlphabet{\mathsfit}{\encodingdefault}{\sfdefault}{m}{sl}
\SetMathAlphabet{\mathsfit}{bold}{\encodingdefault}{\sfdefault}{bx}{n}

\def\sI{{\mathbb{I}}}

\def\sS{{\mathbb{S}}}

\newcommand{\E}{\mathbb{E}}

\begin{document}

\title{Omni-GAN: On the Secrets of cGANs and Beyond}


\author{%
Peng Zhou$^1$, Lingxi Xie$^2$, Bingbing Ni$^1$, Cong Geng$^1$, Qi Tian$^2$\\
$^1$Shanghai Jiao Tong University,\quad$^2$Huawei Inc.\\
{\tt\small zhoupengcv@sjtu.edu.cn}, {\tt\small 198808xc@gmail.com}, {\tt\small nibingbing@sjtu.edu.cn},\\ {\tt\small gengcong@sjtu.edu.cn}, {\tt\small tian.qi1@huawei.com}
}

\maketitle
\ificcvfinal\thispagestyle{empty}\fi

\begin{abstract}
   The conditional generative adversarial network (cGAN) is a powerful tool of generating high-quality images, but existing approaches mostly suffer unsatisfying performance or the risk of mode collapse. This paper presents \textbf{Omni-GAN}, a variant of cGAN that reveals the devil in designing a proper discriminator for training the model. The key is to ensure that the discriminator receives strong supervision to perceive the concepts and moderate regularization to avoid collapse. Omni-GAN is easily implemented and freely integrated with off-the-shelf encoding methods (\textit{e.g.}, implicit neural representation, INR). Experiments validate the superior performance of Omni-GAN and Omni-INR-GAN in a wide range of image generation and restoration tasks. In particular, Omni-INR-GAN sets new records on the ImageNet dataset with impressive Inception scores of \textbf{262.85} and \textbf{343.22} for the image sizes of 128 and 256, respectively, surpassing the previous records by \textbf{100+} points. Moreover, leveraging the generator prior, Omni-INR-GAN can extrapolate low-resolution images to arbitrary resolution, even up to $\times60+$ higher resolution. Code will be available\footnote{\url{https://github.com/PeterouZh/Omni-GAN-PyTorch}}.

\end{abstract}


\section{Introduction}



\begin{figure}[t]
   \footnotesize
   \centering
   \renewcommand{\tabcolsep}{1pt} \renewcommand{\arraystretch}{0.4}
   \begin{tabular}{cc}
      \includegraphics[width=0.5\linewidth,height=3.3cm]{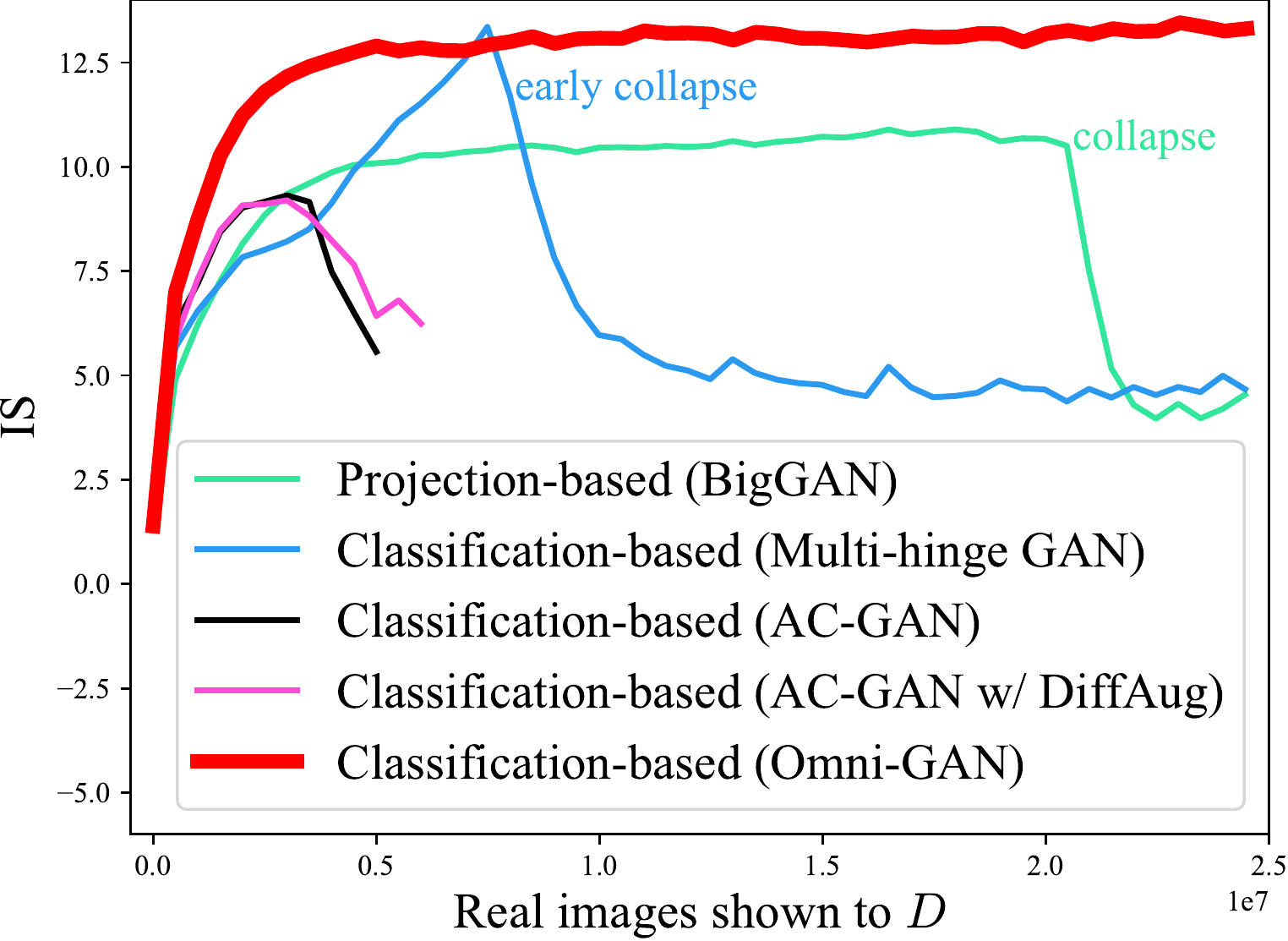}
          & \includegraphics[width=0.5\linewidth,height=3.3cm]{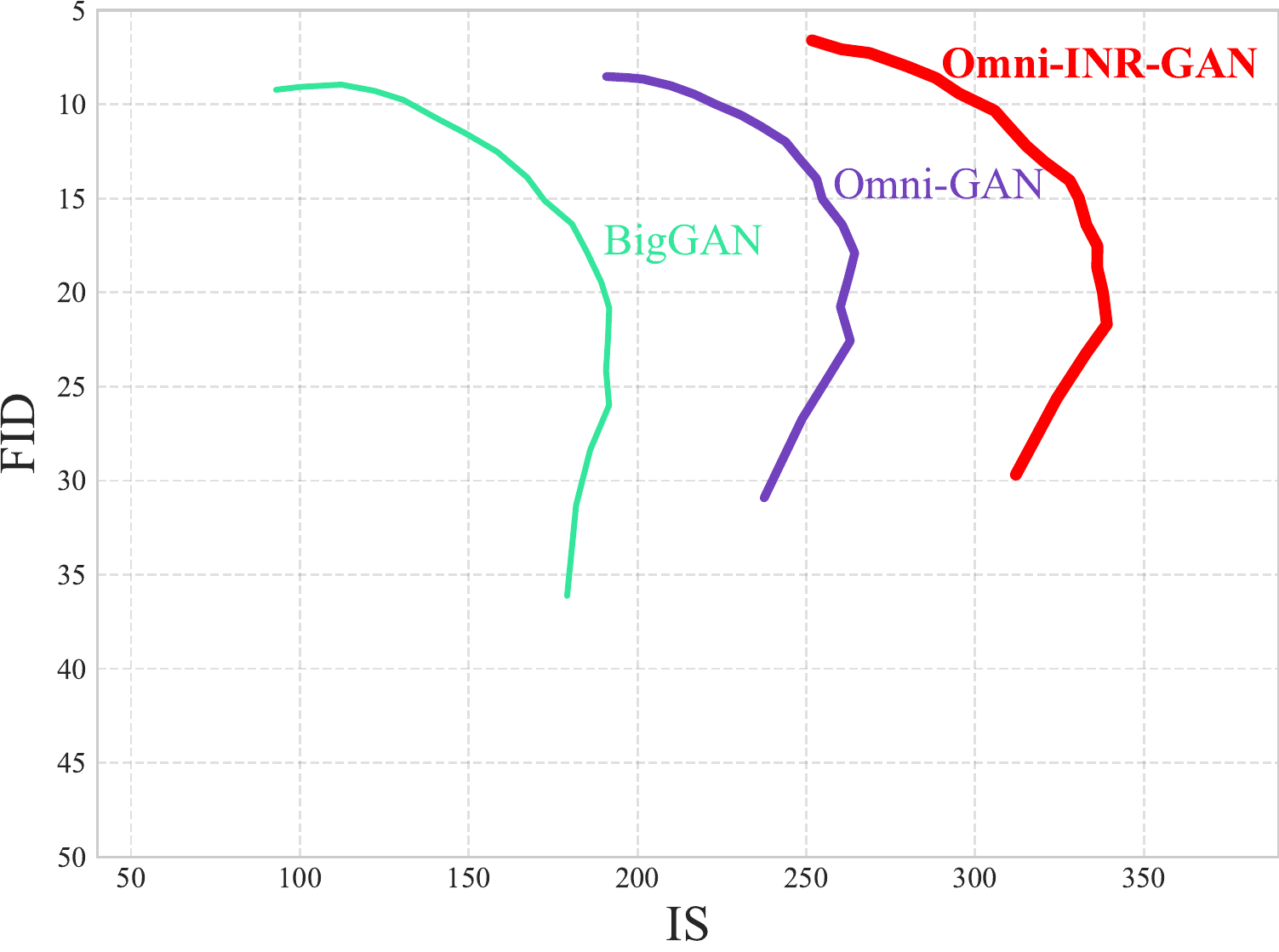} \\
      (a) &
      (b)                                                                                                     \\
   \end{tabular}
   \vspace{-5pt}
   \caption{(a) The trend of Inception scores along the training procedure on CIFAR100, showing that Omni-GAN enjoys both high performance and a lower risk of mode collapse. (b) The tradeoff curves using the truncation trick to generate $128\times128$ images on ImageNet, where Omni-GAN and Omni-INR-GAN outperform BigGAN.}
   \label{fig:introduction}
   \vspace{-5pt}
\end{figure}

The Generative Adversarial Network (GAN)~\cite{goodfellow2014Generative} is a powerful tool for image generation~\cite{arjovsky2017Wasserstein,karras2017Progressive,miyato2018Spectral} and domain adaptation~\cite{chen2020AdversarialLearned,jiang2020Implicit,long2018Conditional,tzeng2017Adversarial}. The big family of GAN can be roughly divided into two parts, \textit{i.e.}, the unconditional GANs~\cite{karras2019StyleBased,karras2019Analyzing} and conditional GANs (cGANs)~\cite{mirza2014Conditional,brock2018Large}, differing from each other in whether the class labels (\textit{e.g.}, cat, car, flower, \textit{etc.}) are used for image generation.
In practice, cGAN often suffers severe collapse when the number of categories is large. As shown in Fig.~\ref{fig:introduction}, all of BigGAN~\cite{brock2018Large}, Multi-hinge GAN~\cite{kavalerov2019cGANs}, and AC-GAN~\cite{odena2017Conditional} achieve high Inception scores, but the curves drop dramatically at some point of training. This makes the cGAN training procedure unstable and thus early termination trick is used by the community~\cite{brock2018Large}.


\begin{figure*}[t]
   \begin{center}
      \includegraphics[width=\linewidth]{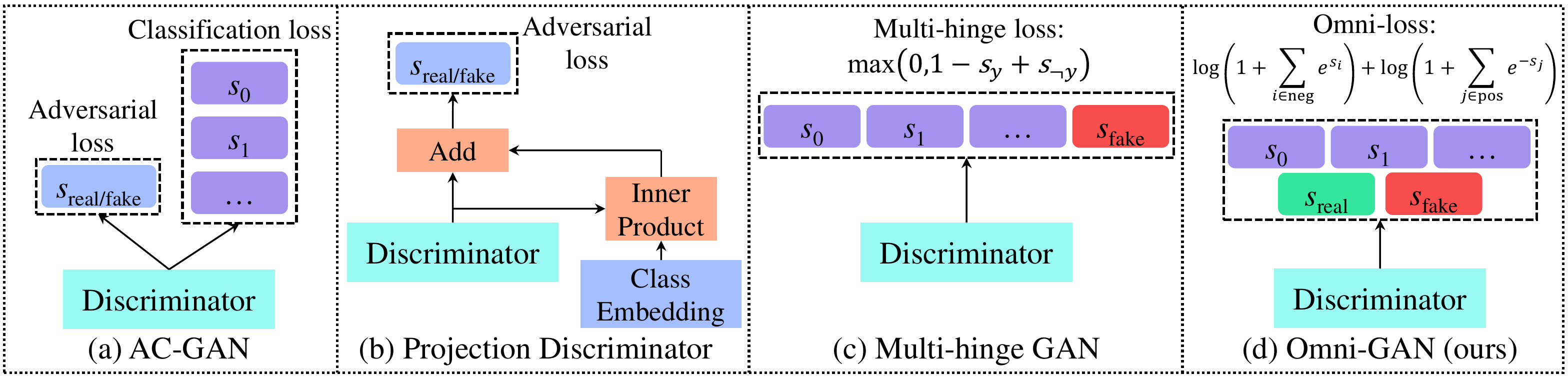}
   \end{center}
   \vspace{-0.5cm}
   \caption{Different discriminator models for training a cGAN. Omni-loss  supports to implement a classification-based cGAN or a projection-based cGAN, enabling us to fairly and intuitively explore the secrets behind them. Please refer to the texts in Sec.~\ref{sec:conditionalGANs} and Sec.~\ref{sec:omni_loss} for details.}
   \vspace{-5pt}
   \label{fig:cgans}
\end{figure*}

It has been noticed~\cite{karras2020Training} that the instability of the training procedure is highly related to the discriminator, \textit{i.e.}, the module that outputs a value indicating the reality of the generated image. The existing cGAN discriminators are roughly categorized into two types, namely, the projection-based~\cite{miyato2018cGANs,brock2018Large} and classification-based~\cite{odena2017Conditional,kavalerov2019cGANs} ones, according to whether the discriminator is required to output an explicit class label for each image. We find that, although the former choice (\textit{i.e.}, a projection-based, with a weaker, implicit discriminator) is inferior to the latter in terms of the Inception score, the latter is prone to collapse (\textit{e.g.}, in Fig.~\ref{fig:introduction}, Multi-hinge GAN achieves a high Inception score but collapses earlier).

This paper investigates the reason behind this phenomenon. We formulate the classification-based and projection-based discriminator into a multi-label classification framework, which offers us an opportunity to observe the advantages and disadvantages of them. As a result, we find that combining strong supervision (classification loss)
and \textbf{moderate regularization} (to prevent it from quickly memorizing the training image set) is the best choice, where the GAN model enjoys high quality in image generation yet has a low risk of mode collapse. In practice, we use weight decay as the choice of regularization, and our algorithm, named \textbf{Omni-GAN}, is easily implemented in any deep learning framework (adding a few lines of code beyond BigGAN). To show that our discovery generalizes to a wide range of cGAN models, we further integrate implicit neural representation (INR)~\cite{park2019DeepSDF,mescheder2019Occupancy,chen2019Learninga}, an off-the-shelf encoding method, into Omni-GAN. This not only improves the generation quality of Omni-GAN, but also enables it to generate images of any aspect ratio and any resolution, facilitating its application to downstream tasks.

We validate the advantages of our approach with extensive experiments on image generation and restoration. The image generation task is performed on CIFAR10, CIFAR100~\cite{krizhevsky2009learning}, and ImageNet~\cite{deng2009ImageNet}, three popular datasets. Omni-GAN surpasses the baselines in terms of the Fréchet Inception distance~\cite{heusel2017GANs} and Inception score~\cite{salimans2016Improved}. In particular, in generating $128\times128$ and $256\times256$ images on ImageNet, Omni-GAN achieves surprising Inception scores of $262.85$ and $343.22$, respectively, both of which surpassing the previous records by more than $100$ points.
The image restoration part involves colorization and single image super-resolution, where Omni-INR-GAN is more flexible than BigGAN and significantly outperforms other restoration methods like DIP~\cite{ulyanov2018Deep} and LIIF~\cite{chen2020Learning}, arguably because the prior learned by the generator is stronger.


We highlight the contributions of this paper as follows:
\begin{itemize}
   \item The core discovery is that combining strong supervision and moderate regularization is the key to cGAN optimization. We achieve this goal easily with the proposed Omni-GAN framework.
   \item We integrate Omni-GAN into a recently published encoding named INR to validate its generalized ability and extend its range of applications.
   \item Omni-GAN achieves the state-of-the-art on the task of image generation on ImageNet, surpassing the prior best results by significant margins. It also requires fewer computational costs to gain the ability of high-quality image generation.
\end{itemize}
We will release the code to the public. We hope that our discovery can inspire the community in studying the principle of generative models and designing powerful algorithms.

\section{Preliminaries}
\label{sec:preliminaries}

\subsection{Conditional GANs}
\label{sec:conditionalGANs}

Conditional GAN (cGAN)~\cite{mirza2014Conditional} adds conditional information to the generator and discriminator of GANs~\cite{kang2020ContraGAN,jolicoeur-martineau2019relativistic,arjovsky2017Principled,arjovsky2017Wasserstein,mescheder2017Numerics,roth2017Stabilizing,heusel2017GANs,karnewar2019MSGGANa,bau2018GAN,schonfeld2020UNet,tian2020AlphaGAN,choi2020StarGAN,gong2019AutoGAN,mao2016Least,qi2017LossSensitive,chen2021UltraDataEfficienta,shim2020CircleGAN}. There are some ways to incorporate class information into the generator, such as conditional batch normalization (CBN)~\cite{devries2017Modulating}, conditional instance normalization (CIN)~\cite{dumoulin2017Learned,huang2017Arbitrary}, class-modulated convolution (CMConv)~\cite{zhou2020Searching}, \etc. There are also many ways to add class information to the discriminator. A simple way is to directly concatenate the class information with the input or features from some middle layers~\cite{denton2015Deep,reed2016Generativea,zhang2017StackGAN,perarnau2016Invertible,saito2017Temporal}. Next, we expound on several slightly complicated methods.

\noindent
\textbf{AC-GAN}\quad
Auxiliary classifier GAN (AC-GAN)~\cite{odena2017Conditional} uses an auxiliary classifier to enhance the standard GAN model (see Fig.~\ref{fig:cgans}a). In particular, the objective function consists of tow parts: the GAN loss, $\mathcal{L}_{\text{GAN}}$, and the classification loss, $\mathcal{L}_{\text{cls}}$:
\begin{equation}
   \begin{aligned}
      \mathcal{L}_{\text{GAN}}=
       & \E \left[ \log P\left(\rg=\text{real} \mid \vx_{\text{real}}\right)\right] + \\
       & \E \left[ \log P\left(\rg=\text{fake} \mid \vx_{\text{fake}}\right)\right],
   \end{aligned}
\end{equation}
\begin{equation}
   \begin{array}{r}
      \mathcal{L}_{\text{cls}}=
      \E \left[ \log P\left(\rg=c \mid \vx_{\text{real}}\right) \right]+
      \E \left[ \log P\left(\rg=c \mid \vx_{\text{fake}}\right) \right],
   \end{array}
\end{equation}
where $\rg$ denotes the label of $\vx$. $\vx_{\text{real}}$ and $\vx_{\text{fake}}$ represent a real image and a generated image respectively. The discriminator $D$ of AC-GAN is trained to maximize $\mathcal{L}_\text{GAN}+\mathcal{L}_\text{cls}$, and the generator is trained to maximize $\mathcal{L}_{\text{cls}}-\mathcal{L}_\text{GAN}$. We will show that the discriminator loss of AC-GAN  is not optimal (see Sec.~\ref{sec:other_tech}).

\noindent
\textbf{Projection Discriminator}\quad
Projection discriminator~\cite{miyato2018cGANs} incorporates class information into the discriminator of GANs in a projection-based way (see Fig.~\ref{fig:cgans}b). The mathematical form of the projection discriminator is given by
\begin{equation}
   D(\vx, \vy)=
   \vy^{\mathrm{T}} \mV f_1\left(\boldsymbol{x} ; \theta_{1}\right) +
   f_2\left( f_1\left(\vx ; \theta_1 \right) ; \theta_2 \right),
   \label{equ:projection_discriminator}
\end{equation}
where $\vx$ and $\vy$ denote the input image and one-hot label vector respectively. $\mV$ is a class embedding matrix, $f_1\left(\cdot; \theta_1\right)$ is a vector function, and $f_2\left(\cdot; \theta_2\right)$ is a scalar function. $\mV, \theta_1, \theta_2$ are learned parameters of $D$. The discriminator $D$ only outputs a scalar for each pair of $\vx$ and $\vy$.

\noindent
\textbf{Multi-hinge GAN}\quad
Multi-hinge GAN~\cite{kavalerov2019cGANs} belongs to classification-based cGANs. It uses a $C+1$ dimensional classifier as the discriminator, which is trained by a multi-class hinge loss (see Fig.~\ref{fig:cgans}c).

\subsection{Implicit Neural Representation}

Images are usually represented by a set of pixels with fixed resolution. A popular method named implicit neural representation (INR) is prevalent in the 3D field~\cite{park2019DeepSDF,mescheder2019Occupancy,chen2019Learninga,atzmon2020SAL,gropp2020Implicit,chabra2020Deep,peng2020Convolutional,jiang2020Local}. Recently, people introduced the INR method to 2D images~\cite{chen2020Learning,skorokhodov2020Adversarial,sitzmann2020Implicita,bemana2020XFields,stanley2007Compositional}. The INR of an image directly maps ($x$, $y$) coordinates to image's RGB pixel values. Since the coordinates are continuous, once we get the INR of an image, we can get images of arbitrary resolutions by sampling different numbers of coordinates.

\subsection{Unified Loss for Feature Learning}

There is a unified perspective for classification tasks. We denote the positive score set as $\sS_{\text{pos}}=\{\evs_1^{(p)}, \cdots, \evs_K^{(p)}\}$, and negative score set as $\sS_{\text{neg}}=\{\evs_1^{(n)}, \cdots, \evs_L^{(n)}\}$, respectively. Sun \etal~\cite{sun2020Circle} proposed a unified loss to maximize $s^{(p)}$ as well as to minimize $s^{(n)}$. The loss is defined as

\begin{equation}
   \resizebox{\hsize}{!}{$
         \begin{aligned}
            \mathcal{L}_{\text {uni}} & =\log \left[1+\sum_{s_{i}^{(n)} \in \sS_{\text{neg}}} \sum_{s_{j}^{(p)} \in \sS_{\text{pos}}} e^ {\left(\gamma\left(s_{i}^{(n)}-s_{j}^{(p)}+m\right)\right)} \right]                                         \\
                                      & =  \log \left[1+\sum_{s_{i}^{(n)} \in \sS_{\text{neg}}} e^ {\left(\gamma\left(s_{i}^{(n)}+m\right)\right)} \sum_{s_{j}^{(p)} \in \sS_{\text{pos}}} e^ {\left(\gamma\left(-s_{j}^{(p)}\right)\right)}\right],
         \end{aligned}
      $}
   \label{equ:unified_loss}
\end{equation}
where $\gamma$ stands for a scale factor, and $m$ for a margin between positive and negative scores.
\myeqref{equ:unified_loss} can be converted into triplet loss~\cite{schroff2015FaceNet} or softmax with the cross-entropy loss~\cite{sun2020Circle}.


\section{Omni-GAN}


\subsection{Omni-GAN and One-sided Omni-GAN}
\label{sec:omni_loss}

We commence from defining the omni-loss. Based on this loss, we design two versions of cGANs: Omni-GAN, being a classification-based cGAN, and one-sided Omni-GAN, being a projection-based cGAN. These two cGANs enable us to fairly and intuitively explore the secrets behind classification-based cGANs and projection-based cGANs.

Let $\vx$ and $\vy$ denote an image and its multi-label vector respectively. $S$ is a classifier. Suppose that there are $K$ positive labels and $L$ negative labels. Then $\vs=S(\vx)$ is a $K+L$ dimensional score vector. The omni-loss is defined as
\begin{equation}
   \resizebox{\hsize}{!}{$
         \begin{aligned}
            \mathcal{L}_{\text {omni}}\left(\vx, \vy\right)
            = & \log \left(1 + \sum_{i \in \sI_{\text{neg}}} e^{\evs_i(\vx)} \right)
            + \log \left(1 + \sum_{j \in \sI_{\text{pos}}} e^{-\evs_j(\vx)} \right),
         \end{aligned}
      $}
   \label{equ:multi_label_loss}
\end{equation}
where $\sI_{\text{neg}}$ is a set consisting of indexes of negative scores (\ie, $|\sI_{\text{neg}}|=L$), and $\sI_{\text{pos}}$ consists of indexes of positive scores (\ie, $|\sI_{\text{pos}}|=K$). $\evs_k(\vx)$ represents the element $k$ of vector $\vs$. \cite{su2020multilabelloss} shows that \myeqref{equ:multi_label_loss} is a special case of \myeqref{equ:unified_loss}. We provide a detailed derivation from \myeqref{equ:unified_loss} to \myeqref{equ:multi_label_loss} in Appendix~\ref{apx:derivation_omniloss}. Next, we introduce two versions of Omni-GAN by setting different labels for the omni-loss.

\noindent
$\bullet$\quad\textbf{The classification-based Omni-GAN.}\quad

We first elucidate the loss of the discriminator for Omni-GAN. The discriminator loss consists of two parts, one for $\vx_{\text{real}}$ (drawn from the training data), and the other for $\vx_{\text{fake}}$ (drawn from the generator). For $\vx_{\text{real}}$, its multi-label vector is given by
\begin{equation}
   \vy_{\text{real}} = [\underbrace{-1, \dots, 1_{\text{gt}}, \dots, -1}_{C}, \underbrace{1_{\text{real}}, -1}_{2}],
   \label{equ:multi_label_real}
\end{equation}
whose dimension is $C+2$, with $C$ being the number of classes of the training dataset. $1_{\text{gt}}$ is $1$ if its index in the vector is equal to the ground truth label of $\vx_{\text{real}}$, otherwise $-1$. We use $1$ to denote the corresponding score belongs to the positive set, and $-1$ to the negative set. The multi-label vector of $\vx_{\text{fake}}$ is also a $C+2$ dimensional vector:
\begin{equation}
   \vy_{\text{fake}} = [\underbrace{-1, \dots, -1, \dots, -1}_{C}, \underbrace{-1, 1_{\text{fake}}}_{2}],
   \label{equ:multi_label_fake}
\end{equation}
where only the last element is $1$.

According to \myeqref{equ:multi_label_loss}, (\ref{equ:multi_label_real}), and (\ref{equ:multi_label_fake}), we define the discriminator loss as
\begin{equation}
   \begin{aligned}
      \mathcal{L}_{D}
      = & \E_{\vx_{\text{real}} \sim p_{\text{d}}} \left[\mathcal{L}_{\text {omni}}\left(\vx_{\text{real}}, \vy_{\text{real}}\right)\right]    \\
        & + \E_{\vx_{\text{fake}} \sim p_{\text{g}}} \left[\mathcal{L}_{\text {omni}}\left(\vx_{\text{fake}}, \vy_{\text{fake}}\right)\right],
   \end{aligned}
   \label{equ:D_loss}
\end{equation}
where $p_{\text{d}}$ is the training data distribution, and $p_{\text{g}}$ is the generated data distribution. It is obvious that the discriminator $D$ actually acts as a multi-label classifier, which takes as input $\vx$, and outputs a score vector $\vs=D(\vx)$.

The generator attempts to fool the discriminator into believing its samples are real. To this end, its multi-label is set to be
\begin{equation}
   \vy_{\text{fake}}^{(\text{G})} = [\underbrace{-1, \dots, 1_{\text{G}}, \dots, -1}_{C}, \underbrace{1_{\text{real}}, -1}_{2}],
   \label{equ:multi_label_G}
\end{equation}
which is the same as $\vy_{\text{real}}$ defined in \myeqref{equ:multi_label_real}. $1_{\text{G}}$ is $1$ if its index in the vector is equal to the label adopted by the generator to generate $\vx_{\text{fake}}$, otherwise $-1$. The generator loss is then given by
\begin{equation}
   \begin{aligned}
      \mathcal{L}_{G}
      = \E_{\vx_{\text{fake}} \sim p_{\text{g}}} \left[\mathcal{L}_{\text {omni}}\left(\vx_{\text{fake}}, \vy_{\text{fake}}^{(\text{G})}\right)\right].
   \end{aligned}
   \label{equ:G_loss}
\end{equation}

\noindent
$\bullet$\quad\textbf{The projection-based (one-sided) Omni-GAN.}

We imitate the way how the projection-based discriminator~\cite{miyato2018cGANs} utilizes class labels (see \myeqref{equ:projection_discriminator}), and design a projection-based variant of Omni-GAN, named one-sided Omni-GAN, which does not fully utilize the class supervision.

It is easy to implement one-sided Omni-GAN: only slightly modify the multi-label vector, $\vy$. Following the setting above, the multi-label vector for $\vx_{\text{real}}$ is set to be
\begin{equation}
   \vy_{\text{real}} = [\underbrace{0, \dots, 1_{\text{gt}}, \dots, 0}_{C}, \underbrace{1, 0}_{2}],
   \label{equ:oneside-multi_label_real}
\end{equation}
where $1_{\text{gt}}$ is $1$ if its index in the vector is equal to the ground truth label of $\vx_{\text{real}}$, otherwise $0$. And $0$ means that the corresponding score will be ignored when calculating the omni-loss. The multi-label vector for $\vx_{\text{fake}}$ is given by
\begin{equation}
   \vy_{\text{fake}} = [\underbrace{0, \dots, -1_{\text{G}}, \dots, 0}_{C}, \underbrace{-1, 0}_{2}],
   \label{equ:oneside-multi_label_fake}
\end{equation}
where $-1_{\text{G}}$ is $-1$ if its index in the vector is equal to the label adopted by the generator to generate $\vx_{\text{fake}}$, otherwise $0$. The discriminator loss is the same as that defined in \myeqref{equ:D_loss}.

For generator, its multi-label vector for $\vx_{\text{fake}}$ is
\begin{equation}
   \vy_{\text{fake}}^{(\text{G})} = [\underbrace{0, \dots, 1_{\text{G}}, \dots, 0}_{C}, \underbrace{1, 0}_{2}],
   \label{equ:oneside-multi_label_G}
\end{equation}
where $1_{\text{G}}$ is $1$ if its index in the vector is equal to the label adopted by the generator to generate $\vx_{\text{fake}}$, otherwise $0$. The generator loss is the same as that defined in \myeqref{equ:G_loss}.

In summary, we introduce two versions of Omni-GAN by setting different multi-label vector for the omni-loss (defined in \myeqref{equ:multi_label_loss}). It is easy to implement these two GANs in practice: as shown in Fig.~\ref{fig:cgans}d, first, let the discriminator output a vector instead of a scalar; second, apply the omni-loss to the output vector.

\begin{figure}[t]
   \begin{center}
      \includegraphics[width=.9\linewidth,height=5cm]{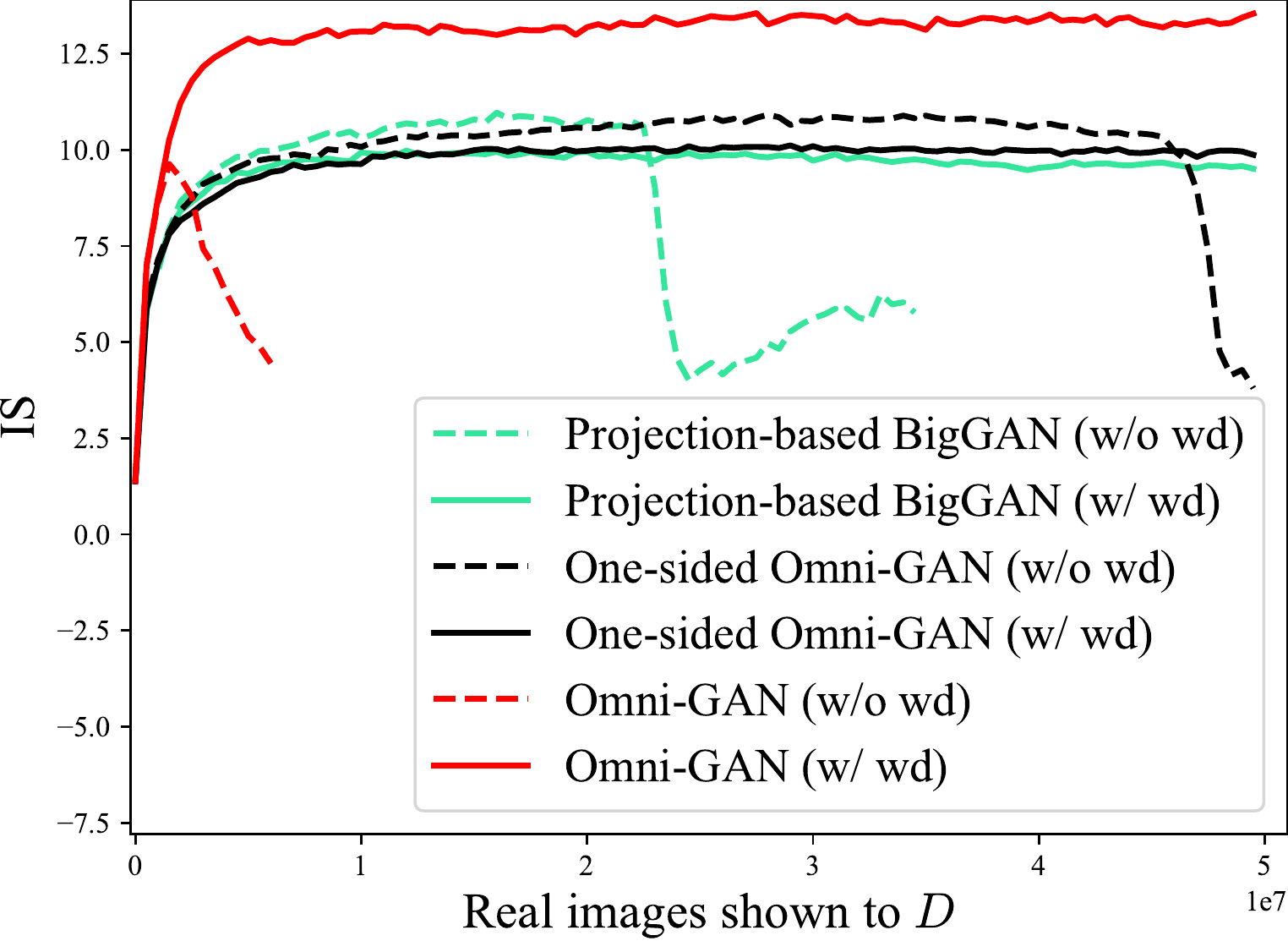}
   \end{center}
   \vspace{-.7cm}
   \caption{IS on CIFAR100. ``wd'' stands for weight decay. The combination of strong supervision and weight decay is crucial. Weight decay effectively alleviates the collapse problem of strongly supervised cGANs (Omni-GAN) so that the cGANs enjoy superior performance from strong supervision. On the other hand, weight decay may slightly impair the performance of weakly supervised cGANs (\eg, projection-based BigGAN, one-sided Omni-GAN).}
   \vspace{-.4cm}
   \label{fig:OmniGAN_onesidedOmniGAN_BigGAN_wd}
\end{figure}

\subsection{The Devil Lies in Combination of Strong Supervision and Moderate Regularization}
\label{sec:early_collapse}

We conducted control experiments on CIFAR100~\cite{krizhevsky2009learning}, and compared Omni-GAN and one-sided Omni-GAN with a projection-based cGAN, namely BigGAN~\cite{brock2018Large}.
As shown in Fig.~\ref{fig:OmniGAN_onesidedOmniGAN_BigGAN_wd}, One-sided Omni-GAN is on par with BigGAN in terms of IS, indicating that one-sided Omni-GAN is indeed a projection-based cGAN. We can see from \myeqref{equ:oneside-multi_label_real}, (\ref{equ:oneside-multi_label_fake}) and (\ref{equ:oneside-multi_label_G}) that one-sided Omni-GAN does not make full use of the class label's supervision. Those elements with label $0$ in the output vector of the discriminator are ignored (\ie, they are not used to calculate the omni-loss, meaning these elements will not get gradients when backpropagation). Therefore, we think that the projection-based cGAN is an implicit and weaker cGAN in the sense that it does not make full use of the supervision of the class label.

On the other hand, as a classification-based cGAN, Omni-GAN makes full use of class supervision (strong supervision). However, it suffers a severe collapse in the initial stages of training. As shown in Fig.~\ref{fig:OmniGAN_onesidedOmniGAN_BigGAN_wd}, the IS of Omni-GAN shows a significant upward compared to the projection-based cGAN but drops dramatically when about $1$M real images ($20$ epoch) are shown to the discriminator.

Our core discovery is that a moderate regularization effectively prevents the early collapse of classification-based cGANs. In practice,  we use weight decay~\cite{krogh1992Simple} as the choice of regularization. We call weight decay moderate regularization in that it does not introduce considerable computational overhead as other regularizations do, such as gradient penalty~\cite{gulrajani2017Improved,mescheder2018Which,wu2018Wasserstein}. Therefore, Omni-GAN can be trained efficiently on large-scale datasets such as ImageNet.
As shown in Fig.~\ref{fig:OmniGAN_onesidedOmniGAN_BigGAN_wd}, combined with weight decay, Omni-GAN has greatly improved its IS compared to BigGAN. Moreover, we observe the projection-based cGAN, BigGAN, also collapses after long training.
Weight decay is also effective for alleviating the collapse of BigGAN.

Note that we are not the first to use weight decay in GANs. \cite{zhou2018Don} applys weight decay to unconditional GANs. However, our main contribution is to emphasize that the combination of strong supervision and weight decay is the key to cGANs. In fact, combining weight decay with weakly supervised cGANs even hurts performance. As shown in Fig.~\ref{fig:OmniGAN_onesidedOmniGAN_BigGAN_wd}, both projection-based BigGAN and one-sided Omni-GAN suffer performance degradation after combined with weight decay.

In summary, we claim that the combination of strong supervision and weight decay is critical for cGANs. Strong supervision helps boost the performance of cGANs but causes severe early collapse. Weight decay effectively alleviates early collapse, so that cGANs can fully enjoy the benefits of strong supervision.

\subsection{Comparison to Previous Approaches}
\label{sec:other_tech}

\begin{figure}[!t]
   \begin{subfigure}{0.49\linewidth}
      \centering
      \includegraphics[width=\linewidth]{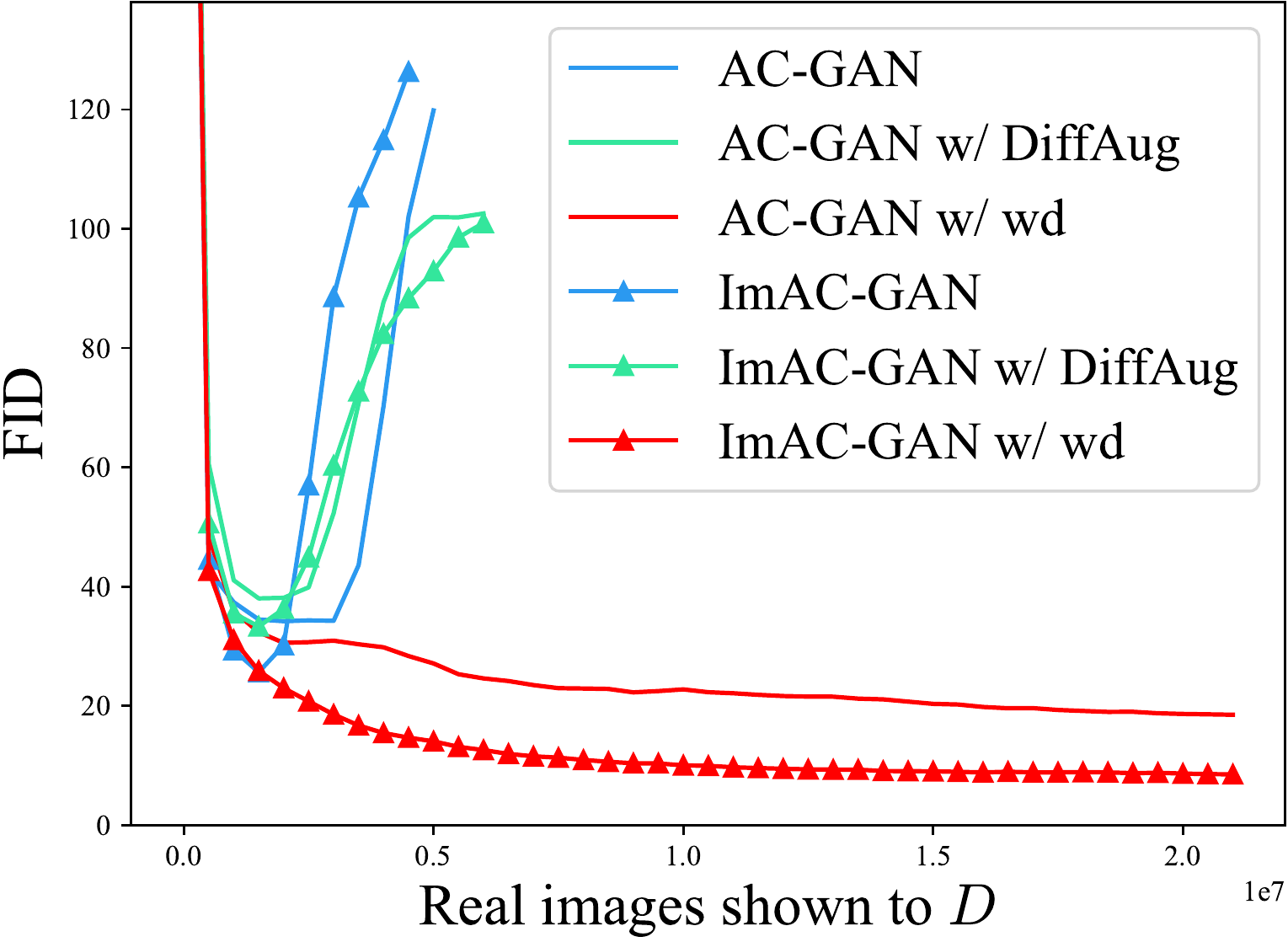}
      \vspace{-.6cm}
      \caption{FID on CIFAR100}
   \end{subfigure}
   \begin{subfigure}{0.49\linewidth}
      \centering
      \includegraphics[width=\linewidth]{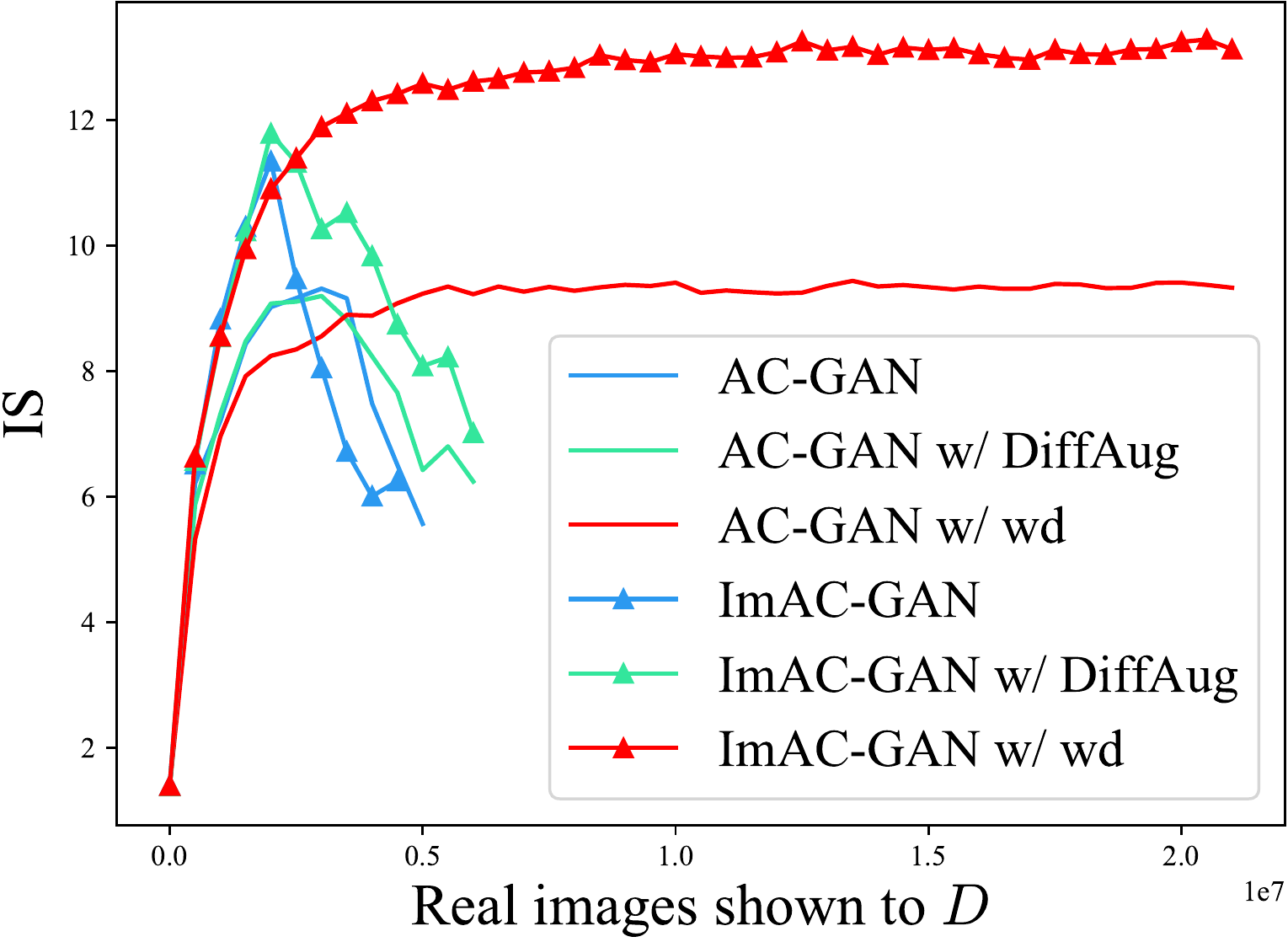}
      \vspace{-.6cm}
      \caption{IS on CIFAR100}
   \end{subfigure}
   \vspace{-.3cm}
   \caption{AC-GAN suffers a severe collapse in the initial stage of training. ``wd'' stands for weight decay, DiffAug for differentiable data augmentation. Weight decay effectively alleviates early collapse, but data augmentation does not. ImAC-GAN means an improved version of AC-GAN. Please refer to Sec.~\ref{sec:other_tech} for details.}
   \vspace{-.4cm}
   \label{fig:acgan_imacgan_diffaug_wd}
\end{figure}

We study another well-known classification-based cGAN, AC-GAN~\cite{odena2017Conditional}, and show its results in Fig.~\ref{fig:acgan_imacgan_diffaug_wd}. AC-GAN also suffers severe early collapse like Omni-GAN does. One possible explanation for the early collapse of classification-based cGANs is that the discriminator overfits the training data~\cite{karras2020Training}. Therefore, we study whether data augmentation is effective to alleviate the early collapse. As shown in Fig.~\ref{fig:acgan_imacgan_diffaug_wd}, AC-GAN combined with differentiable data augmentation (DiffAug)~\cite{zhao2020Differentiable}\footnote{We did not choose ADA augmentation because ADA needs to design different overfitting heuristics for different losses.} still cannot avoid early collapse. However, weight decay is still very effective in alleviating the early collapse of AC-GAN.

We emphasize that weight decay prevents overfitting of the discriminator at the model level, and data augmentation does at the data level. We will show in experiments that both methods are effective for improving the performance of cGANs. However, we empirically find that weight decay is almost $100\%$ effective in preventing the early collapse of classification-based cGANs, but data augmentation is not always effective.

Next, we investigate whether spectral normalization (SN)~\cite{miyato2018Spectral} and gradient penalty~\cite{gulrajani2017Improved,mescheder2018Which,wu2018Wasserstein} will alleviate early collapse. Because the network architectures of the generator and discriminator we used in our experiments employ SN by default, Omni-GAN and AC-GAN still suffer severe early collapse, indicating that SN cannot alleviate the collapse problem of classification-based cGANs. In addition, we have empirically found that gradient penalty cannot help classification-based cGANs avoid early collapse (refer to Appendix~\ref{apx:sec:gp} for experimental results).

Finally, by comparing AC-GAN and Omni-GAN's loss functions, we found that the original AC-GAN still has room for improvement. Due to space limitations, we put the details in Appendix~\ref{apx:sec:imacgan}. We name the improved AC-GAN ImAC-GAN. As shown in Fig.~\ref{fig:acgan_imacgan_diffaug_wd}, the performance of ImAC-GAN is significantly better than that of AC-GAN, and is comparable to that of Omni-GAN (refer to Sec.~\ref{sec:exp:cifar}).

To sum up, our results reveals that fully utilizing the supervision can improve performance of cGANs, but at the risk of early collapse. This work offers a practical way (weight decay) to overcome the collapse issue, so that the trained model enjoys both superior performance and safe optimization.

\begin{figure*}[!t]
   \footnotesize
   \centering
   \renewcommand{\tabcolsep}{1pt} \renewcommand{\arraystretch}{0.4}
   \resizebox{\linewidth}{!}{%
      \centering
      \begin{tabular}{cccccc}
         \includegraphics[height=2.5cm]{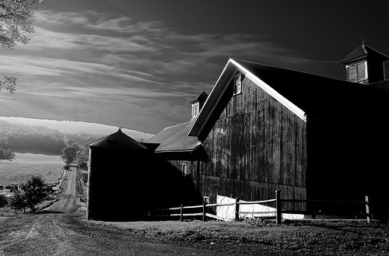}              &
         \includegraphics[height=2.5cm]{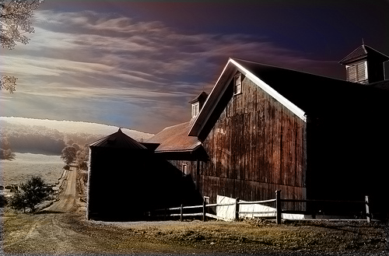}                                &
         \includegraphics[height=2.5cm]{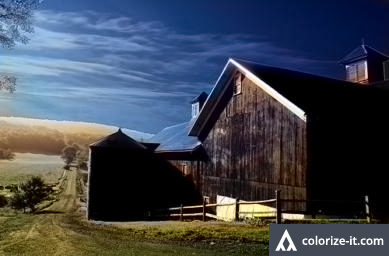}                             &
         \includegraphics[height=2.5cm]{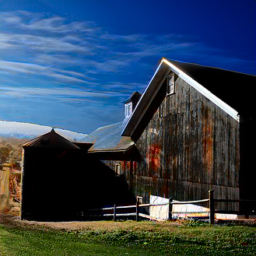} &
         \includegraphics[height=2.5cm]{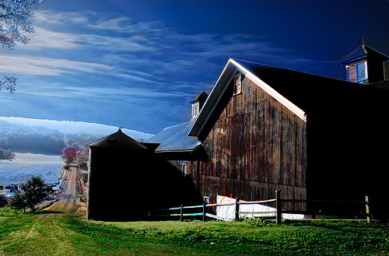}        &
         \includegraphics[height=2.5cm]{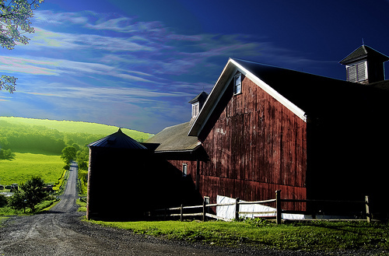}
         \\
         \multirowcell{2}{(a) Grayscale image}                                                                                   & \multirowcell{2}{(b) Autocolorize~\cite{larsson2016Learning}} & \multirowcell{2}{(c) Zhang \etal~\cite{zhang2016Colorful}} & \multirowcell{2}{(d) Restored image \\(DGP w/ BigGAN)} &  \multirowcell{2}{(e) Restored image\\(DGP w/ Omni-INR-GAN)} & \multirowcell{2}{(f) Ground truth} \\
      \end{tabular}

   }
   \vspace{0.1cm}
   \caption{Colorization. (a) degraded input. (b) and (c) automatic colorization algorithms. (d) and (e) GAN inversion-based methods. BigGAN only colorizes a square image patch. Omni-INR-GAN directly colorizes the entire image.}
   \label{fig:dgp_color_limitation}
   \vspace{-5pt}
\end{figure*}

\subsection{Omni-INR-GAN: Being Friendly to Downstream Tasks }

Omni-GAN is easily implemented and freely integrated with off-the-shelf encoding methods. We derive Omni-INR-GAN by integrating implicit neural representation (INR)~\cite{park2019DeepSDF,mescheder2019Occupancy,chen2019Learninga} into Omni-GAN. Omni-INR-GAN employs INR to enhance the generator's output layer. Due to limited space, we detail technical details in Appendix~\ref{apx:sec:omni_inr_gan}.

Omni-INR-GAN has the ability to output images with any aspect ratio and any resolution. Thus it is friendly to downstream tasks like image restoration. Fig.~\ref{fig:dgp_color_limitation} shows an example of combining Omni-INR-GAN with DGP~\cite{pan2020Exploiting} for colorization. As shown in Fig.~\ref{fig:dgp_color_limitation}~(d), the DGP with BigGAN model only colorize a square patch in the original image. This is because BigGAN only generates fixed-size images, which is inflexible for downstream tasks. On the other hand, Omni-INR-GAN is able to generate images with any aspect ratio and any resolution easily so as to directly handle the entire degraded image. Because the generator has seen considerable numbers of natural images, it owns a wealth of prior knowledge. As shown in Fig.~\ref{fig:dgp_color_limitation}~(e), utilizing the generator prior helps get a plausible color image.

Another benefit is that just adding a simple INR network, whose parameters can be neglected compared to the backbone of the generator, greatly boost Omni-GAN's performance on ImageNet. For example, Omni-INR-GAN improves the IS of Omni-GAN from $190.94$ to $262.85$ on ImageNet $128\times128$, and from $304.05$ to $343.22$ on ImageNet $256\times256$, respectively (refer to experiments for details).

\section{Experiments}


\begin{table}[t]
   \centering
   \resizebox{\linewidth}{!}{%
      \begin{tabular}{clcccccc}
         \toprule
         \multirow{2}{*}{No.} & \multicolumn{1}{l}{\multirow{2}{*}{Method}}   & \multicolumn{3}{c}{CIFAR10} & \multicolumn{3}{c}{CIFAR100}                                                               \\
                              & \multicolumn{1}{c}{}                          & FID $\downarrow$            & IS $\uparrow$                & Collapse? & FID $\downarrow$ & IS $\uparrow$    & Collapse? \\ \midrule
         -                    & FQ-GAN~\cite{zhao2020Feature}                 & $6.16$                      & $9.16$                       & No        & $8.23$           & $10.62$          & No        \\
         -                    & Multi-hinge~\cite{kavalerov2019cGANs}         & $6.22$                      & $9.55$                       & No        & $14.62$          & $13.35$          & Yes       \\
         -                    & ADA~\cite{karras2020Training}                 & $\mathbf{2.67}^\dagger$     & $10.06^\dagger$              & -         & -                & -                & -         \\
         \midrule
         0                    & BigGAN, \sout{wd}~\cite{brock2018Large}       & $7.27$                      & $9.19$                       & Yes       & $10.12$          & $10.96$          & Yes       \\
         1                    & BigGAN, wd                                    & $7.77$                      & $9.29$                       & No        & $13.73$          & $9.98$           & No        \\
         2                    & BigGAN, DiffAug, \sout{wd}                    & $5.46$                      & $9.28$                       & No        & $8.63$           & $10.69$          & No        \\
         3                    & BigGAN, DiffAug, wd                           & $7.97$                      & $9.61$                       & No        & $13.97$          & $9.79$           & No        \\
         \midrule
         4                    & AC-GAN, \sout{wd}~\cite{odena2017Conditional} & $7.11$                      & $9.43$                       & Yes       & $34.19$          & $9.31$           & Yes       \\
         5                    & AC-GAN, wd                                    & $8.03$                      & $9.58$                       & No        & $16.41$          & $9.74$           & No        \\
         6                    & AC-GAN, DiffAug, \sout{wd}                    & $5.71$                      & $9.76$                       & No        & $38.01$          & $9.19$           & Yes       \\
         7                    & AC-GAN, DiffAug, wd                           & $8.98$                      & $9.75$                       & No        & $14.91$          & $11.01$          & No        \\
         \midrule
         8                    & ImAC-GAN, \sout{wd}                           & $6.62$                      & $9.46$                       & Yes       & $25.64$          & $11.34$          & Yes       \\
         9                    & ImAC-GAN, wd                                  & $5.63$                      & $9.72$                       & No        & $8.11$           & $13.38$          & No        \\
         10                   & ImAC-GAN, DiffAug, \sout{wd}                  & $4.57$                      & $9.85$                       & No        & $33.30$          & $11.78$          & Yes       \\
         11                   & ImAC-GAN, DiffAug, wd                         & $6.66$                      & $10.02$                      & No        & $8.45$           & $14.54$          & No        \\
         \midrule
         12                   & Omni-GAN, \sout{wd}                           & $7.75$                      & $9.74$                       & Yes       & $26.51$          & $11.45$          & Yes       \\
         13                   & Omni-GAN, wd                                  & $5.57$                      & $9.79$                       & No        & $8.41$           & $13.30$          & No        \\
         14                   & Omni-GAN, DiffAug, \sout{wd}                  & $7.13$                      & $9.86$                       & No        & $39.90$          & $10.66$          & Yes       \\
         15                   & Omni-GAN, DiffAug, wd                         & $7.83$                      & $\textbf{10.37}$             & No        & $11.39$          & $\textbf{15.37}$ & No        \\
         \midrule
         16                   & Omni-INR-GAN, \sout{wd}                       & $8.59$                      & $9.74$                       & Yes       & $53.29$          & $9.05$           & Yes       \\
         17                   & Omni-INR-GAN, wd                              & $5.25$                      & $9.74$                       & No        & $7.63$           & $13.90$          & No        \\
         18                   & Omni-INR-GAN, DiffAug, \sout{wd}              & $75.75$                     & $5.80$                       & Yes       & $53.88$          & $11.06$          & Yes       \\
         19                   & Omni-INR-GAN, DiffAug, wd                     & $4.32$                      & $10.03$                      & No        & $\textbf{6.70}$  & $14.15$          & No        \\
         \bottomrule
      \end{tabular}%
   }
   \vspace{-0.2cm}
   \caption{FID and IS on CIFAR10 and CIFAR100. $\dagger$ indicates quoted from the paper. ``wd'' stands for applying weight decay, and ``\sout{wd}'' for not applying weight decay. DiffAug means differentiable data augmentation. FID and IS are computed using $50$K training and $50$K generated images with the TensorFlow-based pre-trained Inception-V3 model. Please refer to Sec.~\ref{sec:exp:cifar} for detailed analysis.}
   \vspace{-.4cm}
   \label{tab:cifar10_cifar100}
\end{table}

\begin{table*}[t]
   \centering
   \resizebox{\linewidth}{!}{%
      \begin{tabular}{llllc|lllc}
         \toprule
         \multicolumn{1}{l}{\multirow{2}{*}{Method}}   & \multicolumn{4}{c}{ImageNet $128\times128$}        & \multicolumn{4}{c}{ImageNet $256\times256$}                                                                                                                                                                                                                                                      \\
         \multicolumn{1}{c}{}                          & FID (train) $\downarrow$                           & FID (val) $\downarrow$                             & IS $\uparrow$                                        & G Params & FID (train) $\downarrow$                           & FID (val) $\downarrow$                             & IS $\uparrow$                                        & G Params \\
         \midrule
         CR-BigGAN$^\dag$~\cite{zhang2020Consistencya} & $6.66$                                             & -                                                  & -                                                    & -        & -                                                  & -                                                  & -                                                    & -        \\
         S3GAN$^\dagger$~\cite{lucic2019HighFidelitya} & $7.70$                                             & -                                                  & $83.10$                                              & -        & -                                                  & -                                                  & -                                                    & -        \\
         BigGAN$^\ddagger$~\cite{brock2018Large}       & $9.77$                                             & $9.96$                                             & $93.09$                                              & $70.43$M & -                                                  & -                                                  & -                                                    & -        \\
         \midrule
         BigGAN$^\star$                                & $9.19$                                             & $9.18$                                             & $104.57$                                             & $70.43$M & $9.95$                                             & $9.88$                                             & $187.60$                                             & $82.10$M \\
         Omni-GAN                                      & $8.30\textcolor{green}{(0.89\downarrow)}$          & $8.93\textcolor{green}{(0.25\downarrow)}$          & $190.94\textcolor{green}{(86.37\uparrow)}$           & $70.43$M & $6.03\textcolor{green}{(3.92\downarrow)}$          & $6.83\textcolor{green}{(3.05\downarrow)}$          & $304.05\textcolor{green}{(116.45\uparrow)}$          & $82.10$M \\
         Omni-INR-GAN                                  & $\mathbf{6.53\textcolor{green}{(2.66\downarrow)}}$ & $\mathbf{7.99\textcolor{green}{(1.19\downarrow)}}$ & $\mathbf{262.85\textcolor{green}{(158.28\uparrow)}}$ & $70.52$M & $\mathbf{4.93\textcolor{green}{(5.02\downarrow)}}$ & $\mathbf{6.36\textcolor{green}{(3.52\downarrow)}}$ & $\mathbf{343.22\textcolor{green}{(155.62\uparrow)}}$ & $82.19$M \\
         \bottomrule
      \end{tabular}%
   }
   \vspace{-0.2cm}
   \caption{FID and IS on ImageNet dataset. Omni-GAN achieves consistent improvements in terms of FID and IS compared to BigGAN. Omni-INR-GAN improves the IS to $2.5$ times compared with BigGAN on ImageNet $128\times128$, with almost the same number of parameters. $\dag$ stands for quoting from the paper, $\ddag$ for using the model provided by the author, and $\star$ for reproducing BigGAN by us. FID and IS are computed using $50$K generated images. The training and validation data are utilized as the reference distribution for the computing of FID, respectively.}
   \vspace{-10pt}
   \label{tab:imagenet}
\end{table*}

\subsection{Evaluation on CIFAR}
\label{sec:exp:cifar}

We compare a projection-based cGAN, BigGAN, and several classification-based cGANs, including AC-GAN, ImAC-GAN, Omni-GAN, and Omni-INR-GAN. We also study the effects of data augmentation and weight decay on these methods. Results are summarized in Table~\ref{tab:cifar10_cifar100}.

First, let us focus on the projection-based cGAN, BigGAN. We found that data augmentation helps improve the performance of BigGAN, but weight decay cannot, even being harmful. This is reasonable because BigGAN belongs to projection-based cGANs, whose discriminator is essentially a weak implicit classifier. Weight decay constrains the fitting ability of the discriminator (\ie, a weak network). Moreover, the supervision for the discriminator is too weak (\ie, a weak supervision). As such, the performance is naturally not good.

Next, we compare AC-GAN and ImAC-GAN, which belong to classification-based cGANs. ImAC-GAN has a clear and consistent improvement over AC-GAN in all experiments. This improvement is achieved by slightly modifying the loss function of AC-GAN to be consistent with Omni-GAN (refer to Appendix~\ref{apx:sec:imacgan} for technical details).

Third, ImAC-GAN and Omni-GAN are both superior classification-based cGANs, and their performance is comparable. ImAC-GAN uses cross-entropy as the classification loss, so it only supports single-label classification. However, the omni-loss used by Omni-GAN naturally supports multi-label classification. Therefore when the image owns multiple positive labels, Omni-GAN is more flexible than ImAC-GAN. We give an example of using Omni-GAN to generate images with multiple positive labels in Appendix~\ref{apx:sec:multi_label_D}.

We find that training collapse is more likely to appear on CIFAR100 that owns more classes. For example, all classification-based cGANs, including AC-GAN, ImAC-GAN, Omni-GAN, and Omni-INR-GAN, collapsed on CIFAR100 (shown in Table~\ref{tab:cifar10_cifar100}, No. \#4, \#8, \#12, \#16). Even if equipped with data augmentation, they still collapsed (No. \#6, \#10, \#14, \#18). However, weight decay effectively alleviates the training collapse of these methods. On CIFAR10, data augmentation can alleviate the collapse issue of classification-based cGANs (No. \#6, \#10, \#14). We think a possible reason is that CIFAR10 has a relatively small number of categories, and each category has $5,000$ training images. Combining with data augmentation can effectively prevent the discriminator from overfitting the training data. An anomaly is that data augmentation cannot prevent the collapse of Omni-INR-GAN on CIFAR10 (\#18). We found in our experiments that data augmentation seems to be somewhat exclusive with Omni-INR-GAN. We guess that one possible reason is that some random shift operations in data augmentation destroy some prior information to be learned by Omni-INR-GAN from the coordinates.

Last but not least, we have tried to combine Omni-GAN with the network architecture of StyleGAN2~\cite{karras2019Analyzing}, but failed. StyleGAN2 employs some technologies like equalized learning rate~\cite{karras2017Progressive} (explicitly scales the weights at runtime) during the training process. We found that these techniques seem to conflict with weight decay, which is the key for classification-based cGANs to avoid early collapse. Combining strong classification loss with StyleGANs while avoiding early collapse needs further exploration.

\subsection{Evaluation on ImageNet}

\begin{figure}[t]
   \begin{subfigure}{0.49\linewidth}
      \centering
      \includegraphics[width=\linewidth,height=3.3cm]{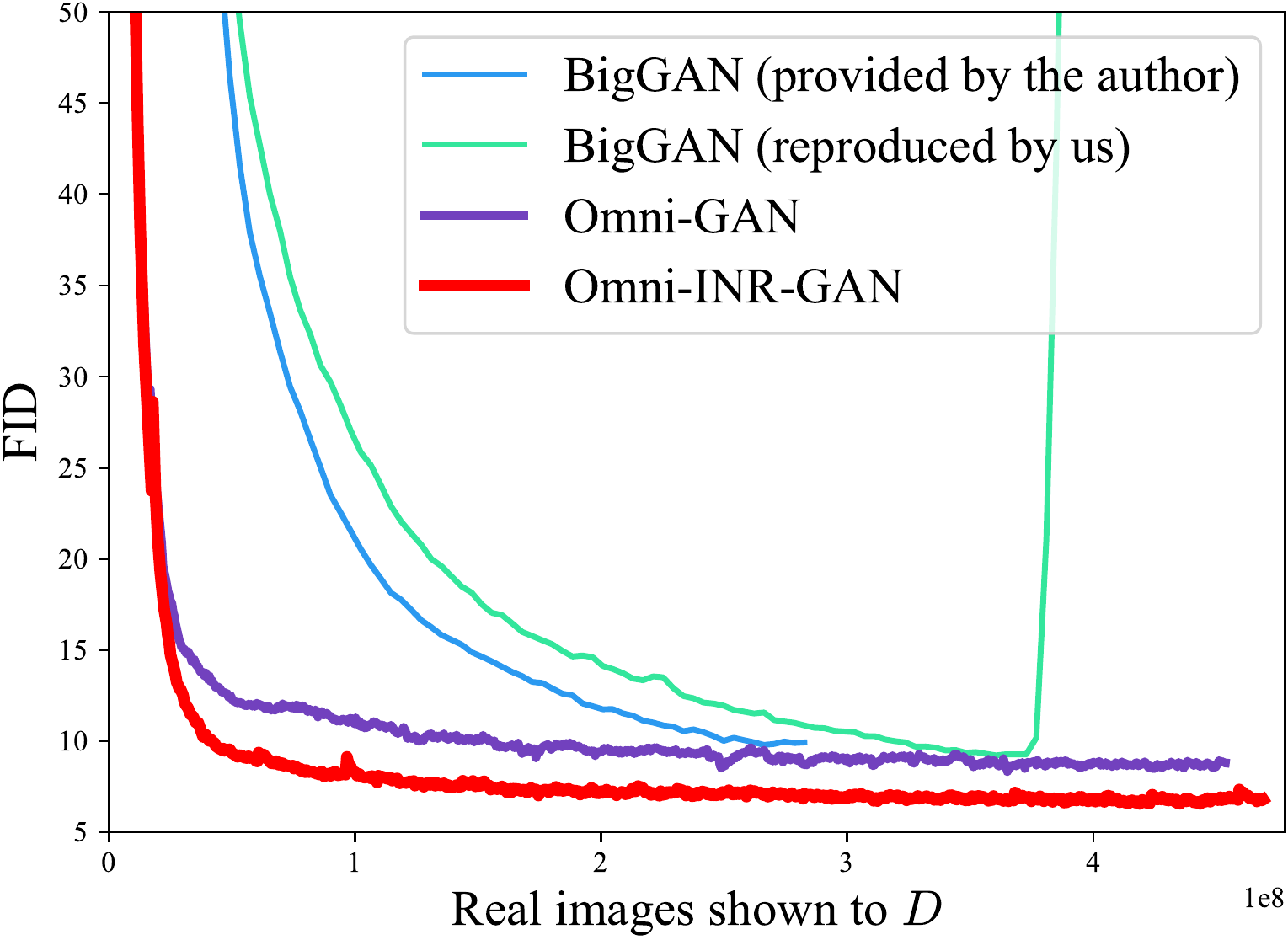}
      \caption{FID on ImageNet $128\times128$}
   \end{subfigure}
   \begin{subfigure}{0.49\linewidth}
      \centering
      \includegraphics[width=\linewidth,height=3.3cm]{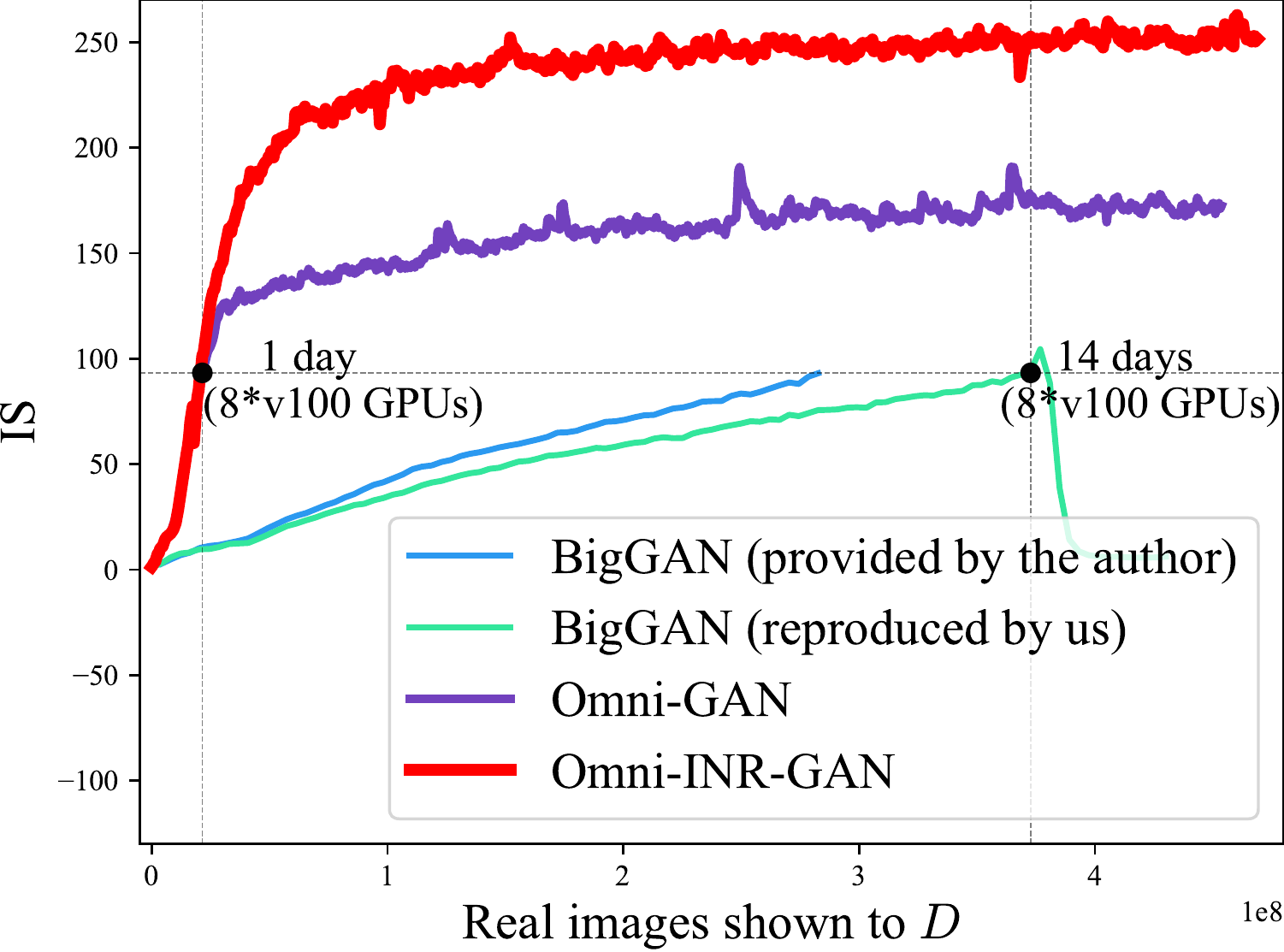}
      \centering
      \caption{IS on ImageNet $128\times128$}
   \end{subfigure}
   \vspace{-0.2cm}
   \caption{Convergence curves on ImgeNet. Both Omni-GAN and Omni-INR-GAN converge faster than the projection-based cGAN, BigGAN. In particular, Omni-GAN only took one day to reach the IS of BigGAN trained for $14$ days. Omni-INR-GAN consistently outperforms BigGAN and Omni-GAN. Its IS is $2.5$ times higher than that of BigGAN (namely $262.85$ \textit{vs.} $104.57$).}
   \vspace{-.4cm}
   \label{fig:imagenet128}
\end{figure}

ImageNet~\cite{deng2009ImageNet} is a large dataset with $1000$ number of classes and approximate $1.2$M training data. We trained BigGAN\footnote{https://github.com/ajbrock/BigGAN-PyTorch.}, Omni-GAN, and Omni-INR-GAN on ImageNet $128\times128$ and $256\times256$, respectively. The results are shown in Table~\ref{tab:imagenet}, and the convergence curves are shown in Fig.~\ref{fig:imagenet128} (please refer to Appendix~\ref{apx:sec:curves_imagenet} for more results).

As shown in Table~\ref{tab:imagenet} and Fig.~\ref{fig:imagenet128}, Omni-GAN shows significant advantages over BigGAN, both in terms of convergence speed and final performance. For example, Omni-GAN only took one day to reach the IS of BigGAN trained for $14$ days on ImageNet $128\times128$. Besides, its IS is almost twice that of BigGAN, namely $190.94$ vs. $104.57$.

Omni-INR-GAN consistently outperforms BigGAN and Omni-GAN. As shown in Table~\ref{tab:imagenet}, on ImageNet $128\times128$, the IS of Omni-INR-GAN is $2.5$ times that of BigGAN. We also show the number of parameters of the generator in Table~\ref{tab:imagenet}. The number of parameters of Omni-INR-GAN is on par with that of BigGAN and Omni-GAN, indicating that the improvement does not lie in the number of parameters. We think the possible reason is that Omni-INR-GAN introduces coordinates as input, which helps the generator learn some prior knowledge of the natural images (for example, the sky often appears in the upper part of an image, and the grass often appears in the lower part on the contrary).

In summary, the significant improvement of Omni-GAN lies in the combination of strong supervision and weight decay. Strong supervision helps boost the performance of cGANs, but it causes the training to collapse earlier. Weight decay effectively alleviates the collapse problem, so that cGANs fully enjoy the benefits from strong supervision.


\subsection{Application to Image-to-Image Translation}

We improve the mIoU of SPADE~\cite{park2019Semantic} from $62.21$ to $65.07$ by only using Omni-GAN loss and weight decay for the discriminator (please refer to Appendix~\ref{apx:sec:imagetoimage} for details). We believe that the improvement comes from the improved ability of the discriminator in distinguishing different classes, so that the generator receives better guidance and thus produces images with richer semantic information.

\begin{table}[t]
   \centering
   \resizebox{0.8\linewidth}{!}{%
      \begin{tabular}{cccc}
         \toprule
                        & BigGAN  & Omni-GAN & Omni-INR-GAN     \\
         \midrule
         PSNR$\uparrow$ & $25.68$ & $26.35$  & $\mathbf{29.36}$ \\
         SSIM$\uparrow$ & $85.17$ & $89.36$  & $\mathbf{92.74}$ \\
         \bottomrule
      \end{tabular}%
   }
   \vspace{-.2cm}
   \caption{Image reconstruction results using pre-trained GAN models. High performance shows the potential to be applied to downstream tasks.}
   \vspace{-.2cm}
   \label{tab:dgp_reconst}
\end{table}

\begin{figure*}[!t]
   \footnotesize
   \centering
   \renewcommand{\tabcolsep}{1pt} \renewcommand{\arraystretch}{0.4}
   \centering
   \resizebox{\linewidth}{!}{%
      \centering
      \begin{tabular}{ccccccccc}
         \includegraphics[width=0.11\linewidth]{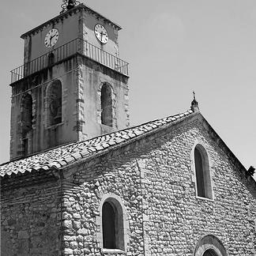}    & \hspace{0.01cm}
         \includegraphics[width=0.11\linewidth]{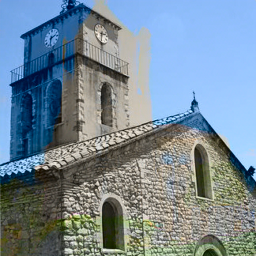}                                  &
         \includegraphics[width=0.11\linewidth]{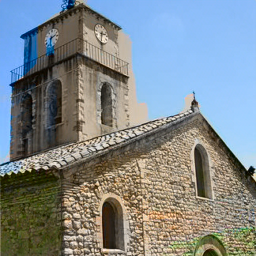}                                  &
         \includegraphics[width=0.11\linewidth]{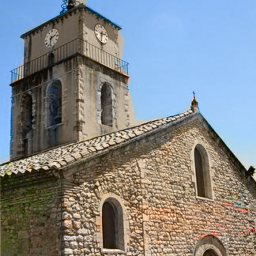}                                  &
         \includegraphics[width=0.11\linewidth]{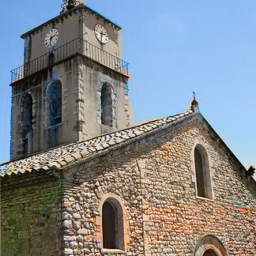}                                  &
         \includegraphics[width=0.11\linewidth]{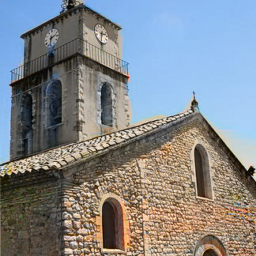}                                  &
         \includegraphics[width=0.11\linewidth]{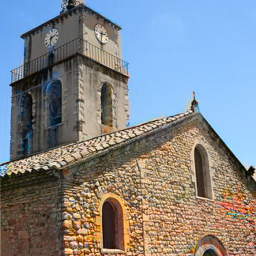}                                  & \hspace{0.01cm}
         \includegraphics[width=0.11\linewidth]{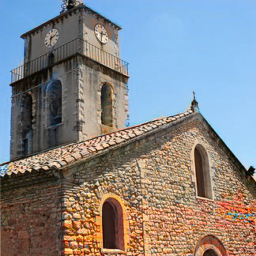}                                  &
         \includegraphics[width=0.11\linewidth]{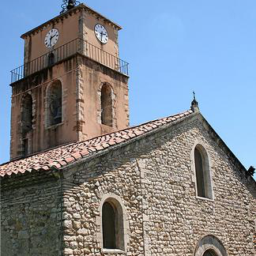}
         \\
         \includegraphics[width=.11\linewidth]{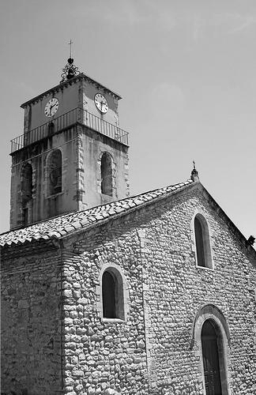} & \hspace{0.01cm}
         \includegraphics[width=.11\linewidth]{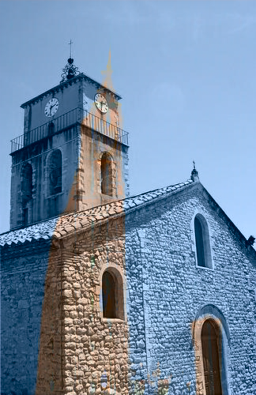}                               &
         \includegraphics[width=.11\linewidth]{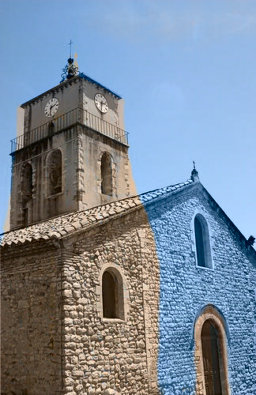}                               &
         \includegraphics[width=.11\linewidth]{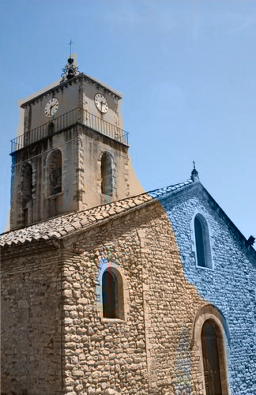}                               &
         \includegraphics[width=.11\linewidth]{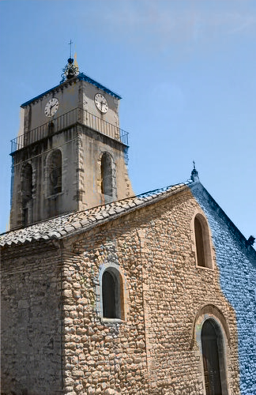}                               &
         \includegraphics[width=.11\linewidth]{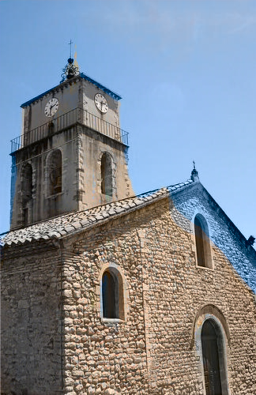}                               &
         \includegraphics[width=.11\linewidth]{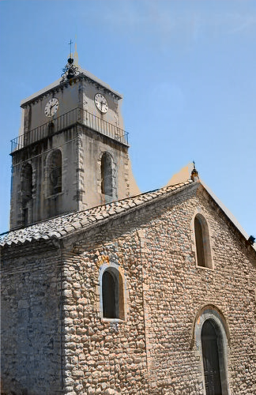}                               & \hspace{0.01cm}
         \includegraphics[width=.11\linewidth]{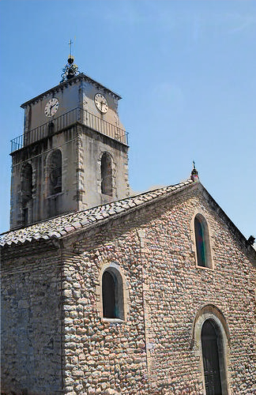}                               &
         \includegraphics[width=.11\linewidth]{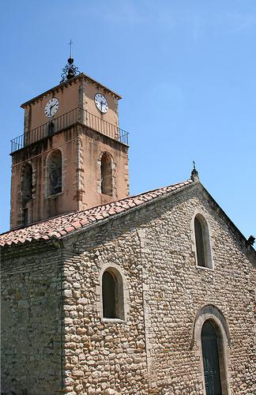}
         \\
         Grayscale image                                                                                                    & \multicolumn{6}{c}{$\longrightarrow$ Finetuning iterations $\longrightarrow$} & Restored image & Ground truth
         \\
      \end{tabular}

   }
   \vspace{-0.3cm}
   \caption{Colorization example. Top: DGP with BigGAN. Bottom: DGP with Omni-INR-GAN. BigGAN only colorizes a square image patch. Omni-INR-GAN directly colorizes the entire image. Omni-INR-GAN gradually overlays colors on the corresponding objects, indicating that the finetuning process is mining GAN's prior information.}
   \label{fig:dgp_colorization}
   \vspace{-5pt}
\end{figure*}

\begin{figure*}[!t]
   \footnotesize
   \centering
   \renewcommand{\tabcolsep}{1pt} \renewcommand{\arraystretch}{0.4}
   \centering
   \resizebox{\linewidth}{!}{%
      \centering
      \begin{tabular}{cc|ccc}
         \includegraphics[width=.19\linewidth]{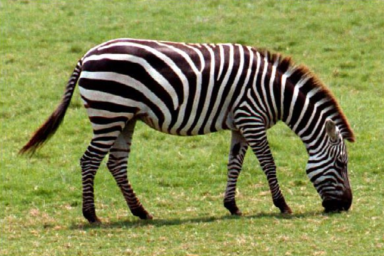}                                        &
         \includegraphics[width=.19\linewidth]{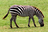}          \hspace{0.01cm}                &
         \hspace{0.01cm}
         \includegraphics[width=.19\linewidth]{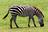}                                   &
         \includegraphics[width=.19\linewidth]{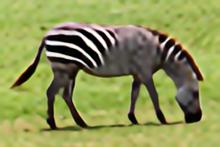}                               &
         \includegraphics[width=.19\linewidth]{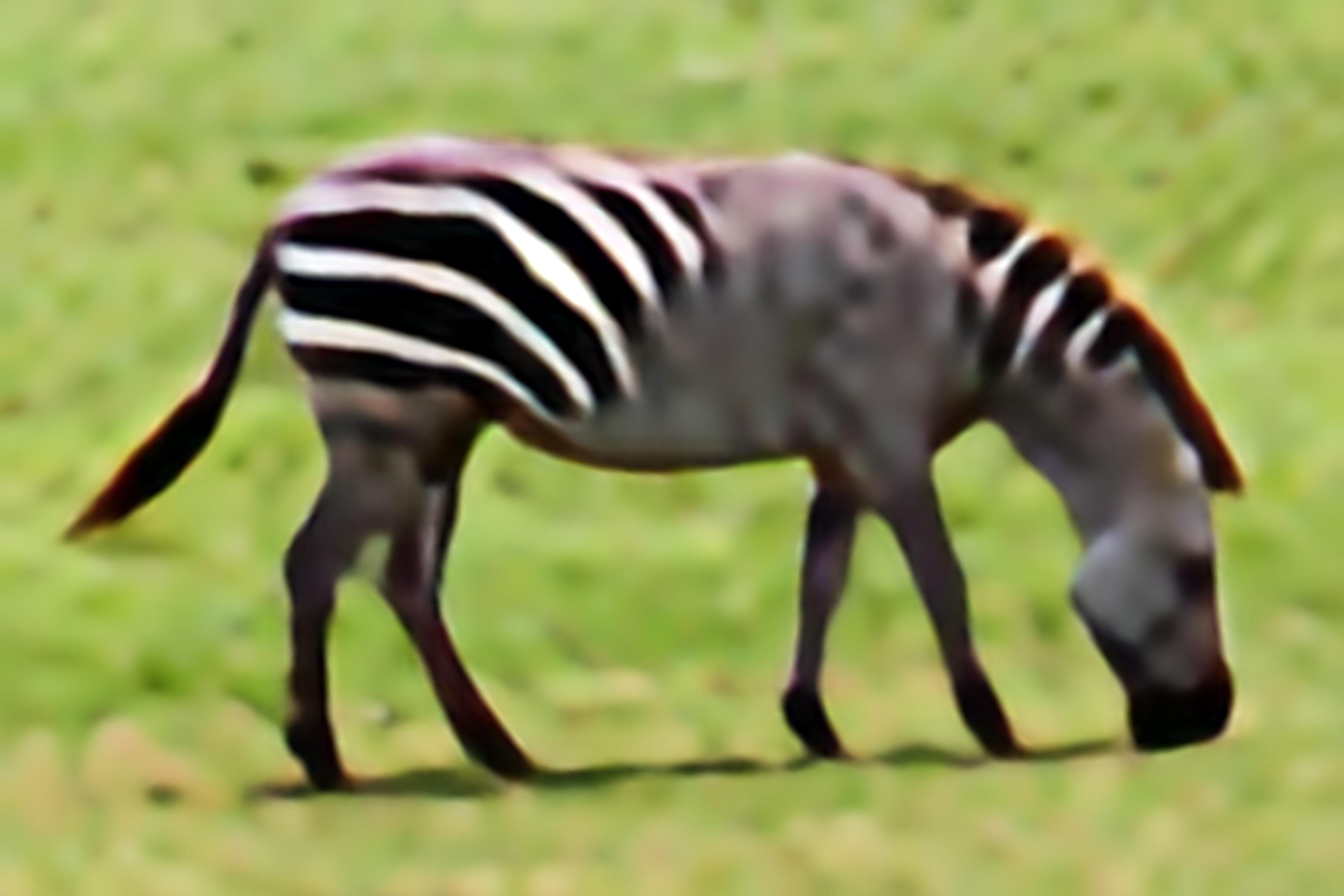}
         \\
         Size ($256\times384$)                                                                                                                       & Size ($32\times48$)          & $\times 1$ ($32\times48$)                          & $\times 4.6$ ($147\times220$) & $\times 63.5$ ($2032\times3048$)
         \\
         (a) Ground truth                                                                                                                            & (b) Input                    & \multicolumn{3}{c}{(c) LIIF}
         \\ \midrule
         \includegraphics[width=.19\linewidth]{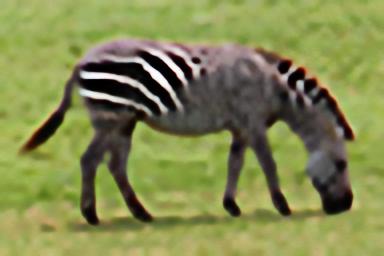}                                      &
         \includegraphics[height=2.19cm]{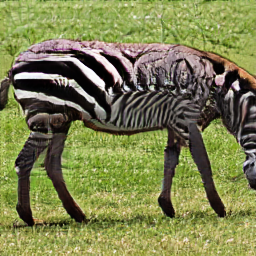}                                                                   &
         \hspace{0.01cm}
         \includegraphics[width=.19\linewidth]{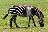}                                         &
         \includegraphics[width=.19\linewidth]{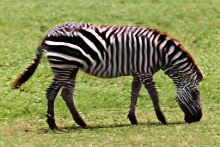}                                       &
         \includegraphics[width=.19\linewidth]{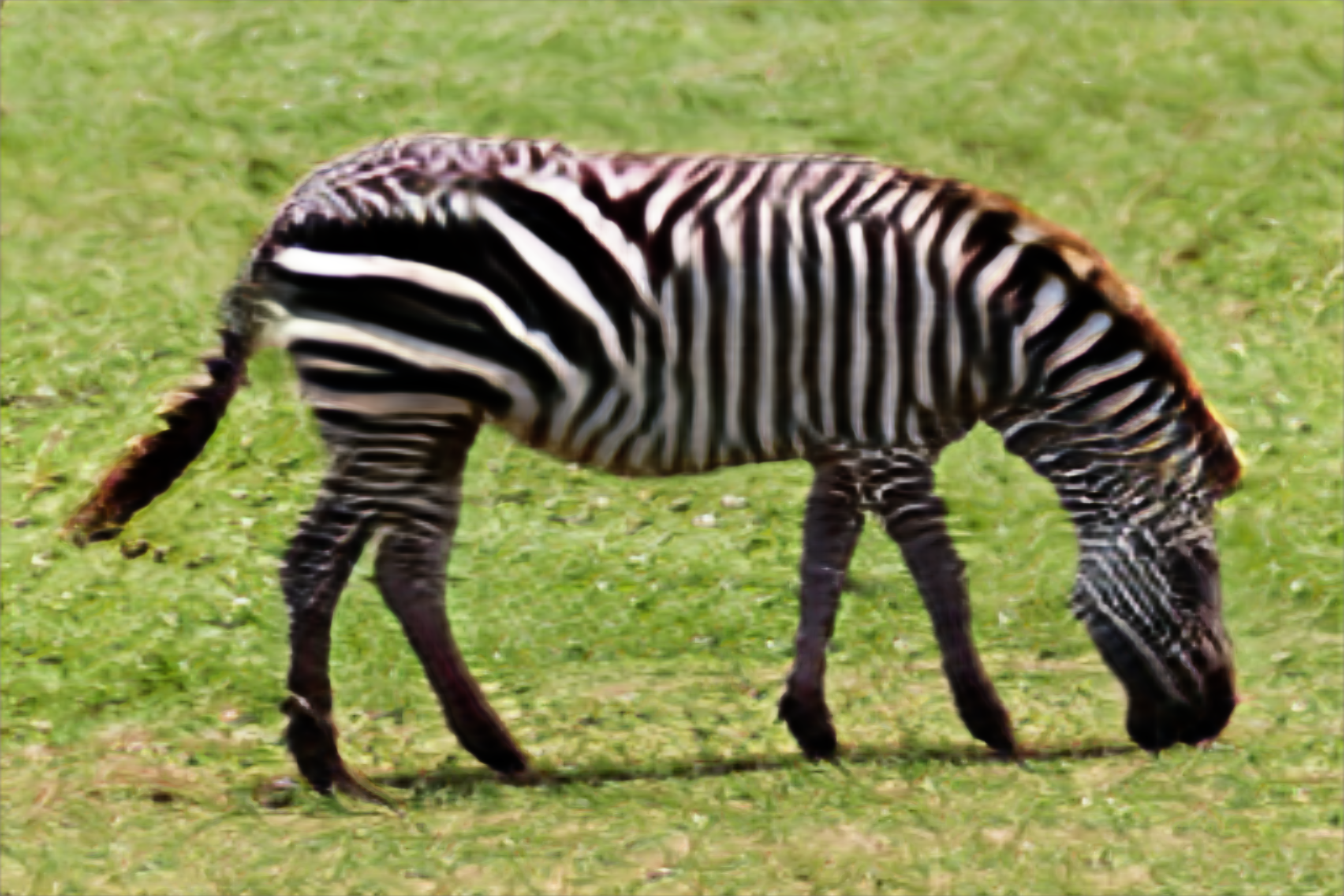}
         \\
         $\times 8$ ($256\times 384$)                                                                                                                & $\times 8$ ($256\times 256$) & $\times 1$ ($32\times48$)                          & $\times 4.6$ ($147\times220$) & $\times 63.5$ ($2032\times3048$)
         \\
         (d) DIP                                                                                                                                     & (e) DGP w/ BigGAN            & \multicolumn{3}{c}{(f) DGP w/ Omni-INR-GAN (ours)}
      \end{tabular}
   }
   \vspace{-0.3cm}
   \caption{Super-resolution using Omni-INR-GAN's prior, at any scale ($\times1$-$\times60+$). (b) input image with low resolution. (c) LIIF~\cite{chen2020Learning} can extrapolate the input image to any scale, but it cannot add semantic details, so the result is still blurred. (d) DIP~\cite{ulyanov2018Deep} also failed because the input image resolution is too low. (e) DGP~\cite{pan2020Exploiting} with BigGAN must crop the input and upsamples the cropped patch to a fixed size, which is inflexible. (f) Omni-INR-GAN has the ability to upsample the input image to any scale and also adds rich semantic details (clearer foreground and background compared to other methods). Please see the video demo in the supplementary material.}
   \label{fig:dgp_SR}
   \vspace{-.5cm}
\end{figure*}


\subsection{Application to Downstream Tasks}

\noindent
\textbf{Colorization.}\quad
Fig.~\ref{fig:dgp_colorization} shows an example of using the pre-trained BigGAN and Omni-INR-GAN to colorize images, respectively. See Appendix~\ref{apx:sec:downstream_tasks} for technical details. Omni-INR-GAN directly colorizes the entire image because it has the ability to output images of any resolution. However, BigGAN cannot do this. Another interesting phenomenon is that Omni-INR-GAN gradually overlays colors on the corresponding objects, indicating that the finetuning process is mining GAN's prior information.

\noindent
\textbf{Prior Enhanced Super-Resolution}\quad
Fig.~\ref{fig:dgp_SR} shows an example of using pre-trained Omni-INR-GAN for super-resolution. We deliver two messages. First, Omni-INR-GAN can extrapolate low-resolution images to any resolution. Fig.~\ref{fig:dgp_SR} (f) shows the results with upsampling scales of $\times1$, $\times4.6$, and even $\times63.5$. Second, Omni-INR-GAN has a wealth of prior knowledge, which helps complement the missing semantics of the input. For example, for the extremely low-resolution zebra image shown in Fig.~\ref{fig:dgp_SR} (b), the image details are severely missing. In this case, although LIIF~\cite{chen2020Learning} has the ability to up-sample images at any scale, it cannot fill in the missing stripes of the zebra, as shown in Fig.~\ref{fig:dgp_SR} (c).

In Fig.~\ref{fig:dgp_SR} (d), DIP~\cite{ulyanov2018Deep} also fails.
DIP only leverages the input image and the structure of a ConvNet as the image prior. It cannot work when the input image's resolution is too low. DGP~\cite{pan2020Exploiting} with BigGAN can only handle fixed-size images, limiting its practical application (Fig.~\ref{fig:dgp_SR} (e)). However, DGP with Omni-INR-GAN can super-resolve the entire image and extrapolate the input at any scale. In Fig.~\ref{fig:dgp_SR} (f), by mining Omni-INR-GAN's prior, the missing zebra stripes are complemented, and a clear foreground and background are obtained. Even the shadow of the zebra is clearly visible.

In Table~\ref{tab:dgp_reconst}, we quantitatively compare different pre-trained GAN models for image reconstruction. Please refer to Appendix~\ref{apx:sec:downstream_tasks} for details. Omni-INR-GAN outperforms BigGAN and Omni-GAN by a large margin, showing its potential for downstream tasks.

\section{Conclusion}

This paper presents an elegant and practical solution to training effective conditional GAN models. The key discovery is that strong supervision can largely improve the upper-bound of image generation quality, but it also makes the model collapse earlier. We design the \textbf{Omni-GAN} algorithm that equips the classification-based loss with regularization (in particular, weight decay) to alleviate collapse. Our algorithm achieves notable performance gain in various scenarios including image generation and restoration. Our research implies that there may be more `secrets' in optimizing cGAN models. We look forward to applying the proposed algorithm and pre-trained models to more scenarios and investigating further properties to improve cGAN.


   {\small
      \bibliographystyle{ieee_fullname}
      \bibliography{egbib}
   }
\clearpage

\appendix

\twocolumn[{%
\centering
\Large\textbf{Omni-GAN and Omni-INR-GAN \\Supplementary Material}
\\
[1.5em]
}]

\begin{center}
  \doparttoc 
  \faketableofcontents 
  \part{}
  \parttoc
  \vspace{2cm}
\end{center}

\begin{figure*}[t]
  \begin{subfigure}{0.5\textwidth}
    \centering
    \includegraphics[width=0.49\linewidth]{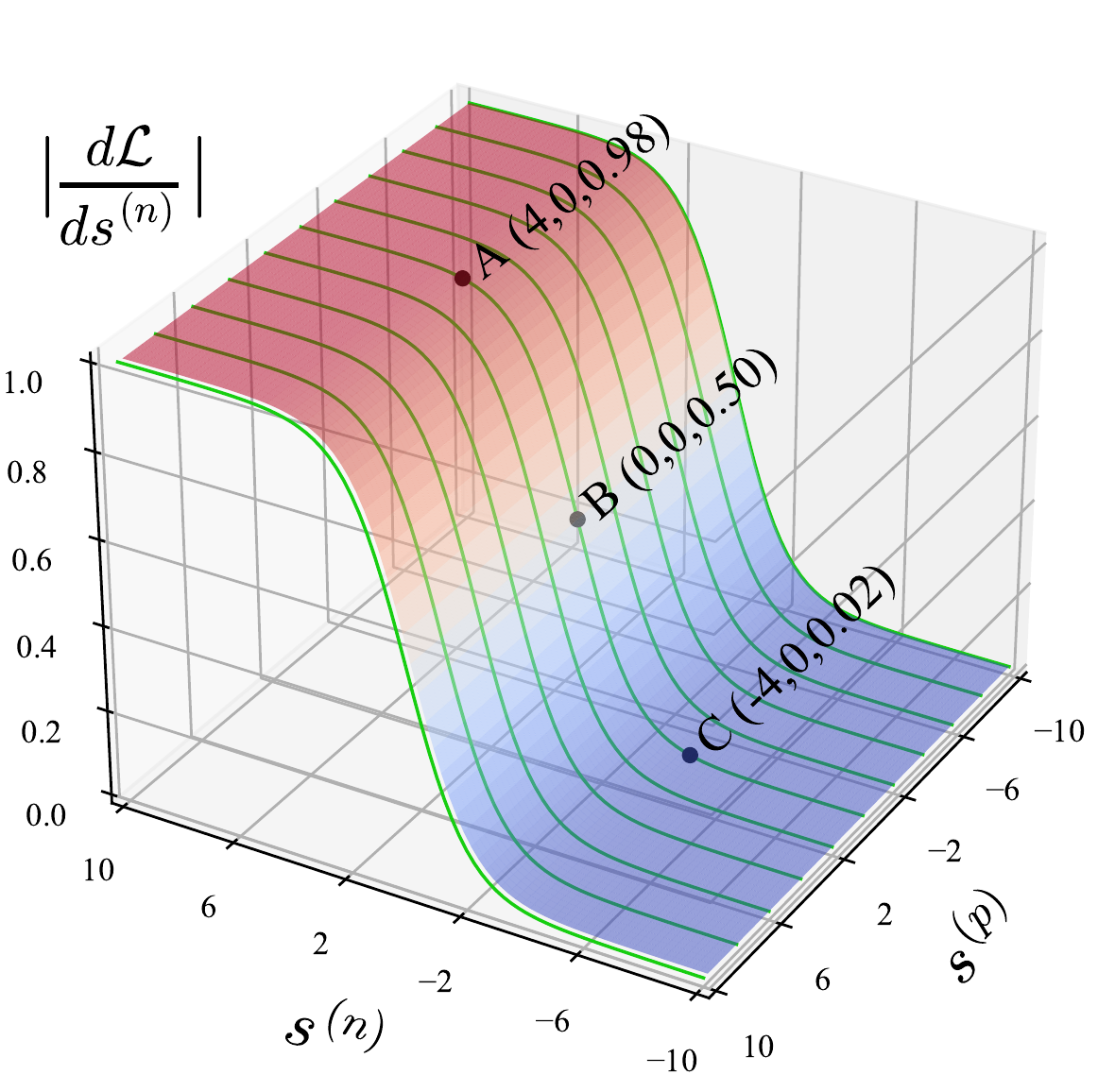}
    \includegraphics[width=0.49\linewidth]{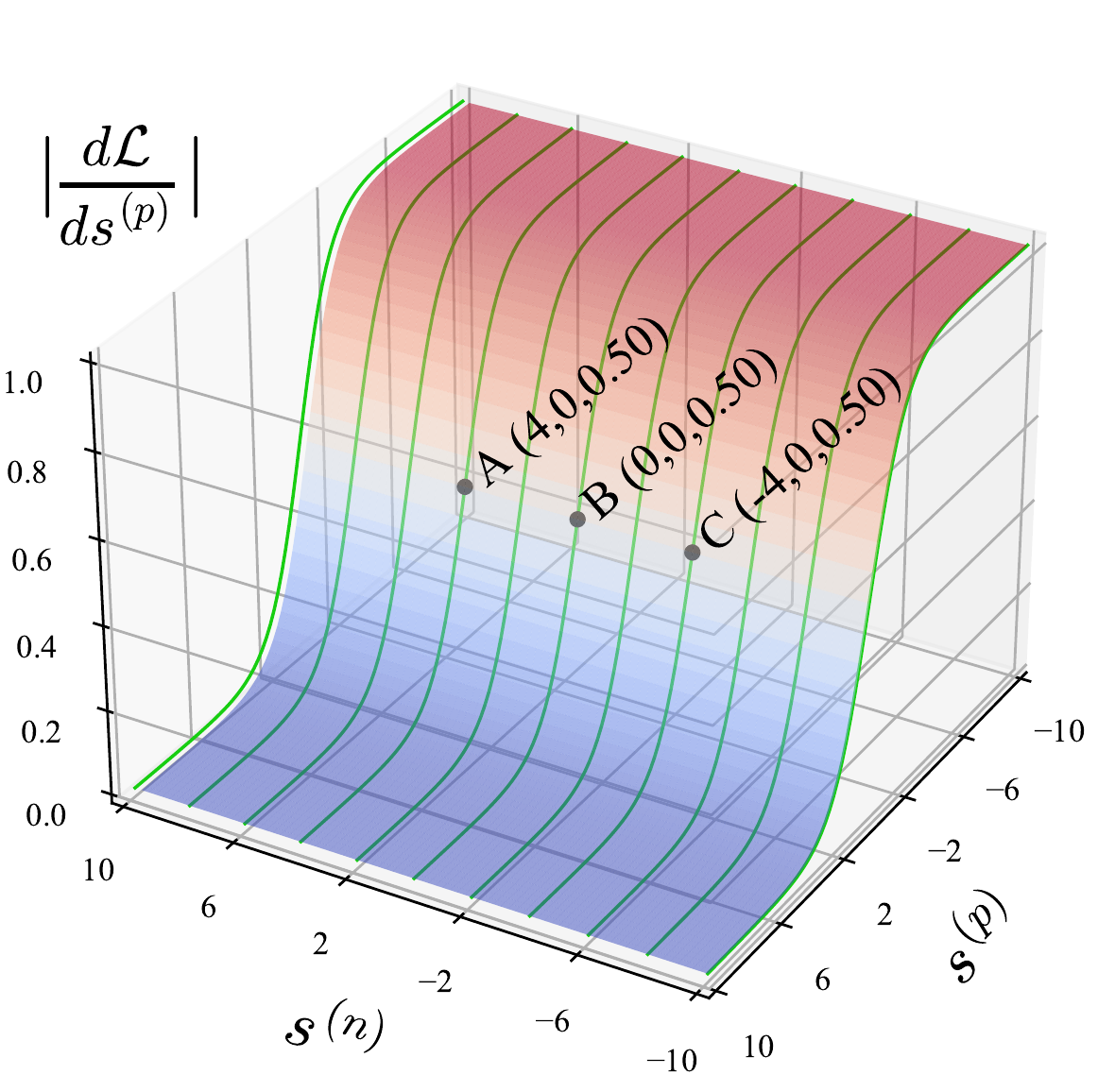}
    \caption{}
  \end{subfigure}
  \hspace{0.1cm}
  \begin{subfigure}{.5\textwidth}
    \centering
    \includegraphics[width=0.49\linewidth]{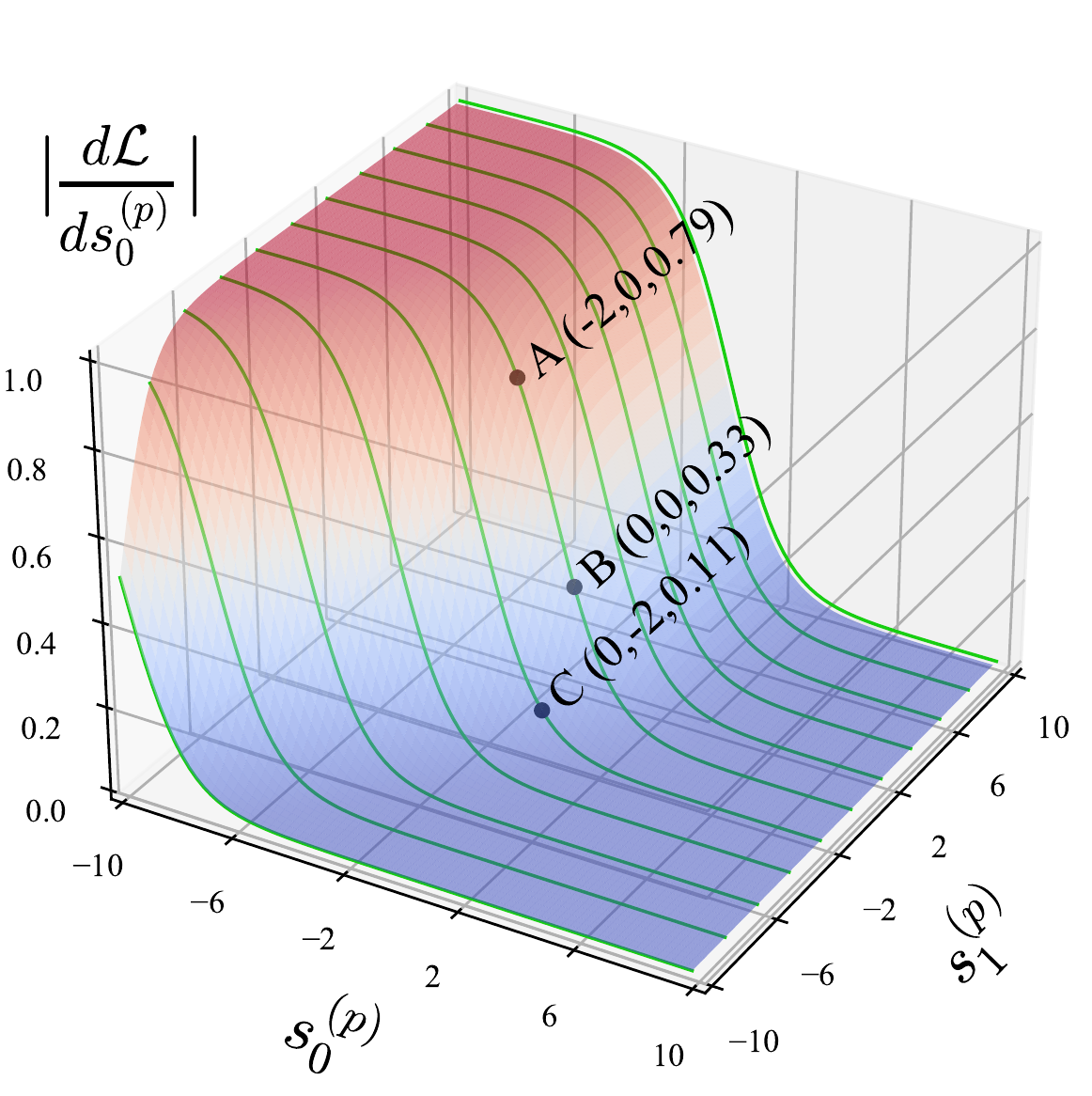}
    \includegraphics[width=0.49\linewidth]{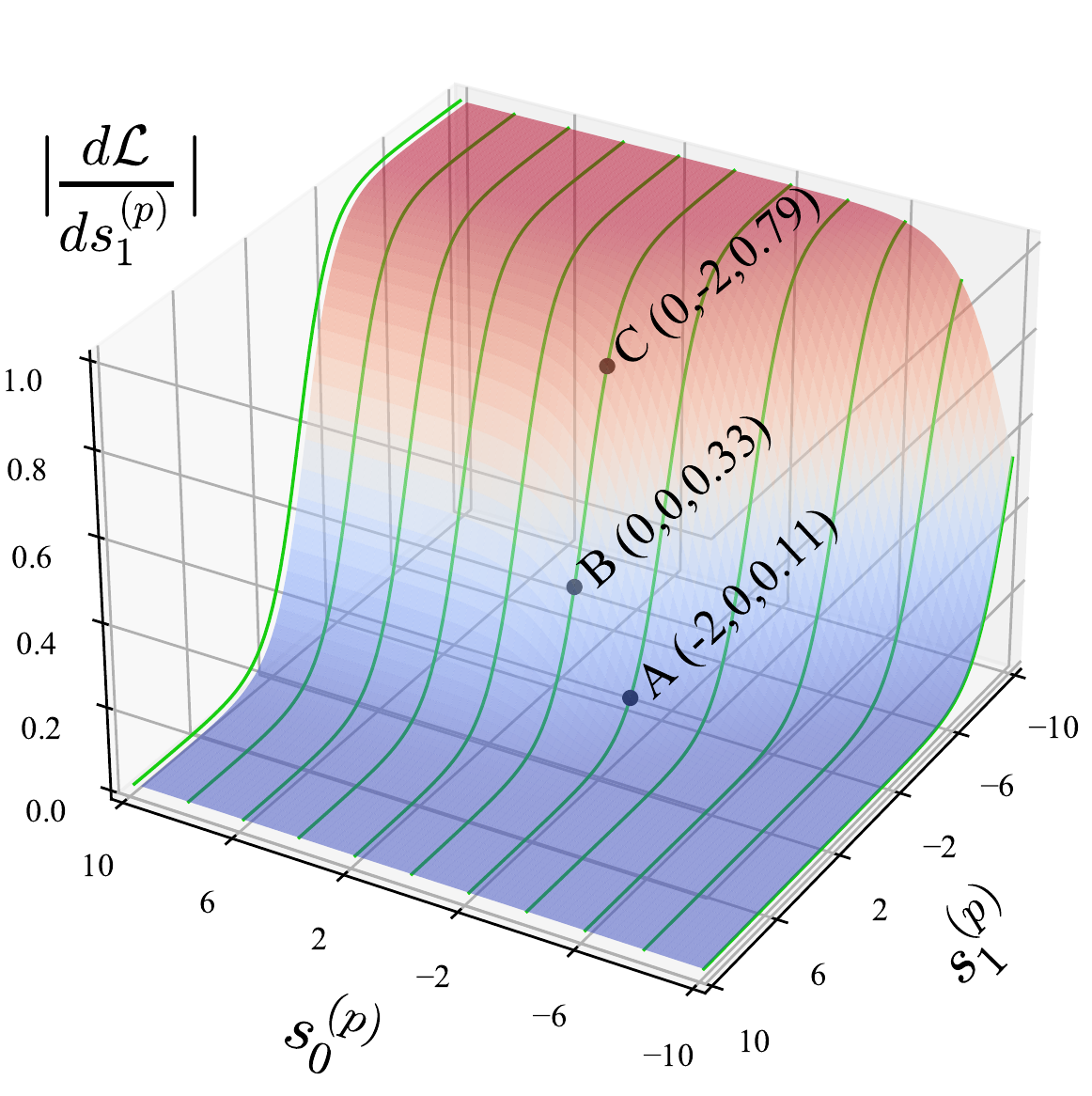}
    \caption{}
  \end{subfigure}
  \vspace{-0.3cm}
  \caption{Gradients of the omni-loss. (a) Gradients \wrt $s^{(n)}$ and $s^{(p)}$ are independent. (b) Gradients \wrt $s^{(p)}_{k}$, $\{k=0,1,\dots\}$, are automatically balanced. Please see the text in Sec.~\ref{sec:gradients} for details. This figure is inspired by \cite{sun2020Circle}.}
  \vspace{-0.3cm}
  \label{fig:gradient}
\end{figure*}

\section{Derivation from Unified Loss to Omni-loss.}
\label{apx:derivation_omniloss}

\subsection{Derivation of Omni-loss}

The unified loss~\cite{sun2020Circle} is defined as
\begin{equation}
  \resizebox{\hsize}{!}{$
      \begin{aligned}
        \mathcal{L}_{\text {uni}} & =\log \left[1+\sum_{s_{i}^{(n)} \in \sS_{\text{neg}}} \sum_{s_{j}^{(p)} \in \sS_{\text{pos}}} e^ {\left(\gamma\left(s_{i}^{(n)}-s_{j}^{(p)}+m\right)\right)} \right]                                         \\
                                  & =  \log \left[1+\sum_{s_{i}^{(n)} \in \sS_{\text{neg}}} e^ {\left(\gamma\left(s_{i}^{(n)}+m\right)\right)} \sum_{s_{j}^{(p)} \in \sS_{\text{pos}}} e^ {\left(\gamma\left(-s_{j}^{(p)}\right)\right)}\right],
      \end{aligned}
    $}
  \label{apx:equ:unified_loss}
\end{equation}
where $\gamma$ stands for a scale factor, and $m$ for a margin between positive and negative scores. $\sS_{\text{pos}}=\{\evs_1^{(p)}, \cdots, \evs_K^{(p)}\}$ and $\sS_{\text{neg}}=\{\evs_1^{(n)}, \cdots, \evs_L^{(n)}\}$ denote positive score set and negative score set, respectively.
\myeqref{apx:equ:unified_loss} aims to maximize $s^{(p)}$ and to minimize $s^{(n)}$.

The omni-loss is defined as
\begin{equation}
  \resizebox{\hsize}{!}{$
      \begin{aligned}
        \mathcal{L}_{\text {omni}}\left(\vx, \vy\right)
        = & \log \left(1 + \sum_{i \in \sI_{\text{neg}}} e^{\evs_i(\vx)} \right)
        + \log \left(1 + \sum_{j \in \sI_{\text{pos}}} e^{-\evs_j(\vx)} \right),
      \end{aligned}
    $}
  \label{apx:equ:multi_label_loss}
\end{equation}
where $\sI_{\text{neg}}$ is a set consisting of indexes of negative scores ($|\sI_{\text{neg}}|=L$), and $\sI_{\text{pos}}$ consists of indexes of positive scores ($|\sI_{\text{pos}}|=K$).

\myeqref{apx:equ:multi_label_loss} is a special case of \myeqref{apx:equ:unified_loss}, which has been proved by ~\cite{su2020multilabelloss}. For the convenience of readers in the English community, we provide our proof here. Let $\gamma$ be $1$ and $m$ be $0$, then
\begin{align}
  \mathcal{L}_{\text {uni}}                                     = & \log \left[1+\sum_{s_{i}^{(n)} \in \sS_{\text{neg}}} e^ {s_{i}^{(n)}} \sum_{s_{j}^{(p)} \in \sS_{\text{pos}}} e^ {-s_{j}^{(p)}}\right]                       \label{apx:equ:unified_loss_r1_m0}                                                                                                                  \\
  =                                                               & \log \left[ 1 + e                                                                                                                      ^ { \log \left( \sum_{s_{i}^{(n)} \in \sS_{\text{neg}}} e^ {s_{i}^{(n)}} \sum_{s_{j}^{(p)} \in \sS_{\text{pos}}} e^ {-s_{j}^{(p)}} \right) } \right]   \nonumber          \\
  =                                                               & \text{softplus} \left[    {                                                                                                                 \log \left( \sum_{s_{i}^{(n)} \in \sS_{\text{neg}}} e^ {s_{i}^{(n)}} \sum_{s_{j}^{(p)} \in \sS_{\text{pos}}} e^ {-s_{j}^{(p)}} \right) } \right]           \nonumber \\
  =                                                               & \text{softplus} \left[    {                                                                                                                 \log \left( \sum_{s_{i}^{(n)} \in \sS_{\text{neg}}} \sum_{s_{j}^{(p)} \in \sS_{\text{pos}}} e^ {s_{i}^{(n)} - s_{j}^{(p)}} \right) } \right]               \nonumber \\
  \approx                                                         & \left[    {                                                                                                                 \log \left( \sum_{s_{i}^{(n)} \in \sS_{\text{neg}}} \sum_{s_{j}^{(p)} \in \sS_{\text{pos}}} e^ {s_{i}^{(n)} - s_{j}^{(p)}} \right) } \right]_{+}               \nonumber  ,
\end{align}
where $[\cdot]_+$ means $\max(\cdot, 0)$.

According to $\log \sum_{i=1}^{n} e^{x_i} \approx \max(x_1, x_2, \dots, x_n) $, we get
\begin{equation}
  \begin{aligned}
    \mathcal{L}_{\text {uni}}                                     \approx & \left[    {                                                                                                                 \max_{s_{i}^{(n)} \in \sS_{\text{neg}}, s_{j}^{(p)} \in \sS_{\text{pos}}} {s_{i}^{(n)} - s_{j}^{(p)}}  } \right]_{+}      ,
  \end{aligned}
  \label{apx:equ:unified_loss_approx}
\end{equation}
where minimizing \myeqref{apx:equ:unified_loss_approx} makes the smallest $s_{j}^{(p)}$ greater than the largest $s_{i}^{(n)}$.

Let $\sS^{(1)}_{\text{pos}}=\{0\}$ and $\sS^{(1)}_{\text{neg}}=\{\evs_1^{(n)}, \cdots, \evs_L^{(n)}\}$. According to \myeqref{apx:equ:unified_loss_r1_m0}, we get
\begin{equation}
  \begin{aligned}
    \mathcal{L}^{(1)}_{\text {uni}}                                     = & \log \left[1+\sum_{s_{i}^{(n)} \in \sS^{(1)}_{\text{neg}}} e^ {s_{i}^{(n)}} \sum_{s_{j}^{(p)} \in \{0\}} e^ {-s_{j}^{(p)}}\right] \\
    =                                                                     & \log \left[1+\sum_{s_{i}^{(n)} \in \sS^{(1)}_{\text{neg}}} e^ {s_{i}^{(n)}} e^ {0}\right]                                         \\
    =                                                                     & \log \left[1+\sum_{s_{i}^{(n)} \in \sS^{(1)}_{\text{neg}}} e^ {s_{i}^{(n)}}\right],
  \end{aligned}
  \label{apx:equ:omni_loss_l1}
\end{equation}
where from \myeqref{apx:equ:unified_loss_approx} we know that minimizing \myeqref{apx:equ:omni_loss_l1} makes $s_{i}^{(n)}$ less than $0$.

Let $\sS^{(2)}_{\text{pos}}=\{\evs_1^{(p)}, \cdots, \evs_K^{(p)}\}$ and $\sS^{(2)}_{\text{neg}}=\{0\}$. According to \myeqref{apx:equ:unified_loss_r1_m0}, we get
\begin{equation}
  \begin{aligned}
    \mathcal{L}^{(2)}_{\text {uni}}
    = & \log \left[1+\sum_{s_{i}^{(n)} \in \{0\} } e^ {s_{i}^{(n)}} \sum_{s_{j}^{(p)} \in \sS^{(2)}_{\text{pos}}} e^ {-s_{j}^{(p)}}\right] \\
    = & \log \left[1+ e^ {0} \sum_{s_{j}^{(p)} \in \sS^{(2)}_{\text{pos}}} e^ {-s_{j}^{(p)}}\right]                                        \\
    = & \log \left[1+ \sum_{s_{j}^{(p)} \in \sS^{(2)}_{\text{pos}}} e^ {-s_{j}^{(p)}}\right]                                               \\
  \end{aligned}
  \label{apx:equ:omni_loss_l2}
\end{equation}
where minimizing \myeqref{apx:equ:omni_loss_l2} makes $s_{j}^{(p)}$ greater than $0$.

Adding \myeqref{apx:equ:omni_loss_l1} and \myeqref{apx:equ:omni_loss_l2}, we get
\begin{equation}
  \resizebox{\hsize}{!}{$
      \begin{aligned}
        \mathcal{L}_{\text {omni}}
        = & \log \left[1+\sum_{s_{i}^{(n)} \in \sS^{(1)}_{\text{neg}}} e^ {s_{i}^{(n)}}\right] + \log \left[1+ \sum_{s_{j}^{(p)} \in \sS^{(2)}_{\text{pos}}} e^ {-s_{j}^{(p)}}\right],
      \end{aligned}
    $}
  \label{apx:equ:omni_loss}
\end{equation}
where minimizing \myeqref{apx:equ:omni_loss} makes $s_{i}^{(n)}$ less than $0$ and $s_{j}^{(p)}$ greater than $0$.
We finish the derivation.

\subsection{Gradient Analysis}
\label{sec:gradients}

The gradients of omni-loss have two properties: on one hand, the gradients \wrt $s^{(n)}$ and $s^{(p)}$ are independent; on the other hand, the gradients \wrt $s^{(p)}_{k}$ (or $s^{(n)}_k$), $\{k=0,1,\dots\}$, are automatically balanced. To illustrate these properties, we visualize the gradients of omni-loss. Fig.~\ref{fig:gradient}a shows a case that only contains one $s^{(n)}$ and one $s^{(p)}$. $A$, $B$, and $C$ have the same $s^{(p)}$, which is $0$, but different $s^{(n)}$ (\ie, $4, 0, -4$, respectively). As a result, the gradients \wrt $s^{(p)}$ at these three points are the same (\ie, $0.5$). Nevertheless, the gradients \wrt $s^{(n)}$ at these three points are different. For example, the gradient \wrt $s^{(n)}$ at $A$ is largest (equal to $0.98$).  The reason for this is that the objective of omni-loss is to minimize $s^{(n)}$. Thus the larger the $s^{(n)}$, the larger the gradient \wrt $s^{(n)}$.

In Fig.~\ref{fig:gradient}b, we show the ability of omni-loss to automatically balance gradients. We consider a case with only two positive labels, namely $s^{(p)}_0$ and $s^{(p)}_1$. We can observe that for $A$, its $s^{(p)}_0$ is smaller than $s^{(p)}_1$ (\ie, -2 vs. 0). As a result, the gradients \wrt $s^{(p)}_0$ is larger than that \wrt $s^{(p)}_1$ (\ie, 0.79 vs. 0.11), meaning that the omni-loss try to increase $s^{(p)}_0$ with higher superiority. A similar analysis applies to $C$ as well. For $B$, since $s^{(p)}_0$ and $s^{(p)}_1$ are equal, the gradients of them are also equal ($0.33$).

\section{Gradient Penalty for Classification-based cGANs}
\label{apx:sec:gp}

\begin{figure}[t]
  \centering
  \begin{subfigure}{0.49\linewidth}
    \centering
    \includegraphics[width=\linewidth,height=3.3cm]{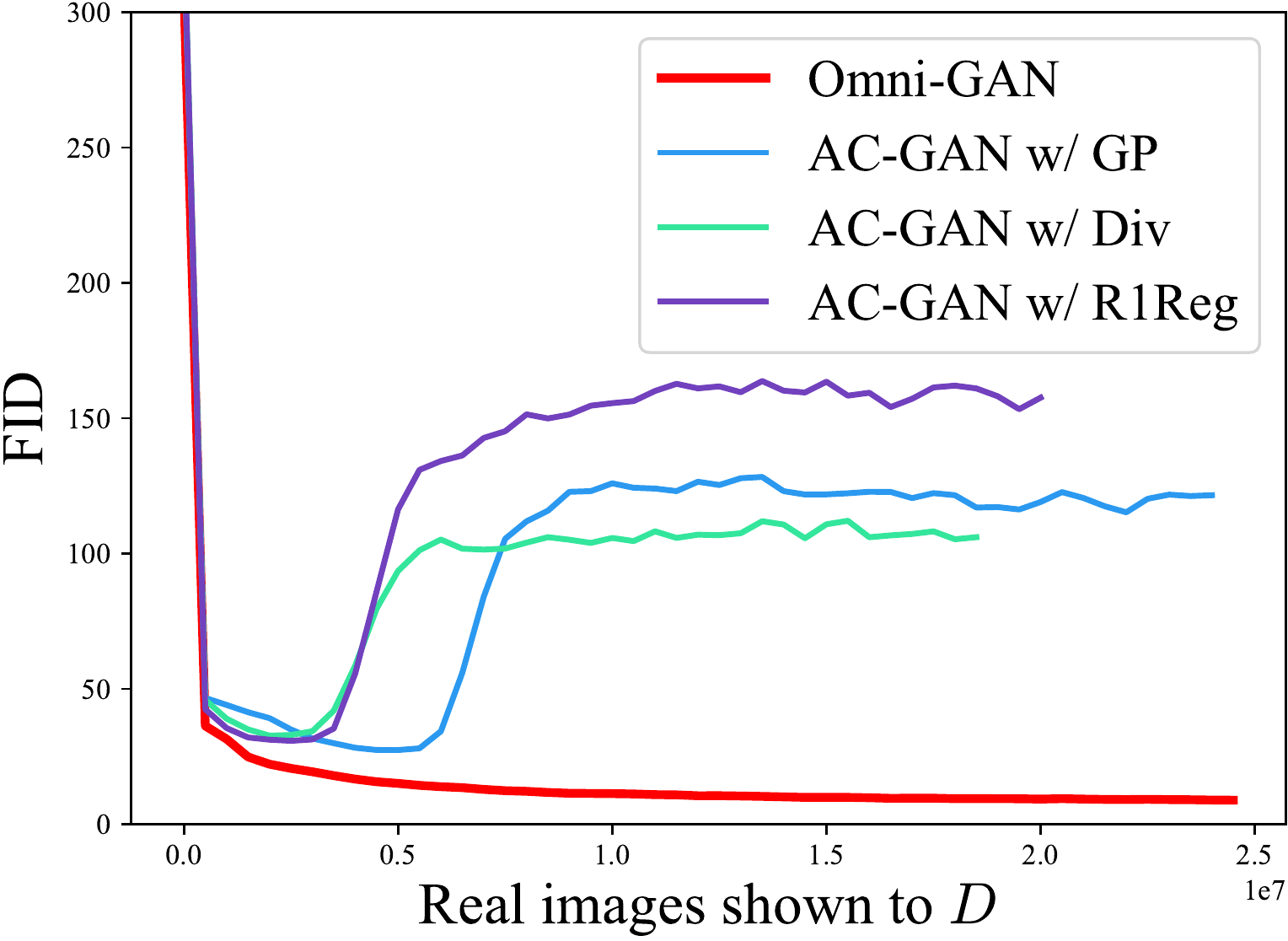}
    \caption{FID on CIFAR100}
  \end{subfigure}
  \begin{subfigure}{0.49\linewidth}
    \centering
    \includegraphics[width=\linewidth,height=3.3cm]{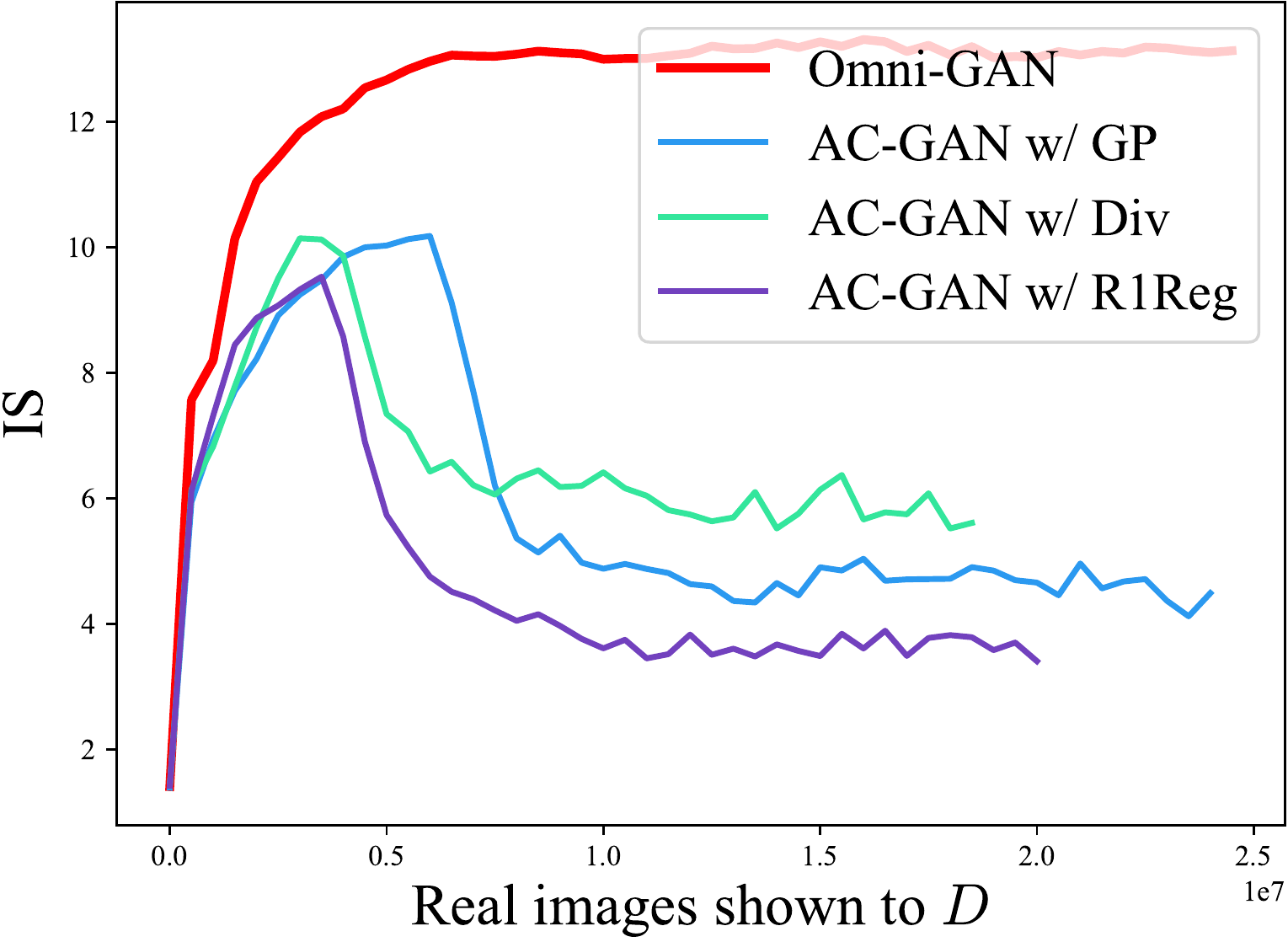}
    \caption{IS on CIFAR100}
  \end{subfigure}
  \vspace{-0.2cm}
  \caption{FID and IS on CIFAR100. We test three gradient penalty methods (\ie, WGAN-GP, WGAN-div, and R1 regularization), none of which can alleviate the collapse issue of AC-GAN.}
  \label{apx:fig:acgan_gp}
\end{figure}

We investigate whether gradient penalty will alleviate early collapse. We chose AC-GAN~\cite{odena2017Conditional}, the currently widely known classification-based cGAN, as the testbed, and evaluate three gradient penalty methods: WGAN-GP~\cite{gulrajani2017Improved}, WGAN-div~\cite{mescheder2018Which}, and R1 regularization~\cite{wu2018Wasserstein}. Because cGANs are more likely to collapse when the number of categories is large, we evaluate them on CIFAR100 instead of CIFAR10. As shown in Fig.~\ref{apx:fig:acgan_gp}, none of the three gradient penalty methods can prevent AC-GAN from collapsing. We emphasize that computing gradient penalties will introduce additional computational overhead during GAN's training, which is very unfriendly to large-scale datasets such as ImageNet. However, weight decay effectively alleviates the collapse problem without adding any additional training overhead.

\section{Improved AC-GAN (ImAC-GAN)}
\label{apx:sec:imacgan}

Auxiliary classifier GAN (AC-GAN)~\cite{odena2017Conditional} uses an auxiliary classifier to enhance the standard GAN model. Its objective function consists of tow parts: the GAN loss, $\mathcal{L}_{\text{GAN}}$, and the classification loss, $\mathcal{L}_{\text{cls}}$:
\begin{equation}
  \begin{aligned}
    \mathcal{L}_{\text{GAN}}=
     & \E \left[ \log P\left(\rg=\text{real} \mid \vx_{\text{real}}\right)\right] + \\
     & \E \left[ \log P\left(\rg=\text{fake} \mid \vx_{\text{fake}}\right)\right],
  \end{aligned}
\end{equation}
\begin{equation}
  \begin{array}{r}
    \mathcal{L}_{\text{cls}}=
    \E \left[ \log P\left(\rg=c \mid \vx_{\text{real}}\right) \right]+
    \E \left[ \log P\left(\rg=c \mid \vx_{\text{fake}}\right) \right],
  \end{array}
\end{equation}
where $\rg$ is a random variable denoting the class label and $c$ is the ground truth label of $\vx$. $\vx_{\text{real}}$ and $\vx_{\text{fake}}$ represent a real image and a generated image respectively. The discriminator $D$ of AC-GAN is trained to maximize $\mathcal{L}_\text{GAN}+\mathcal{L}_\text{cls}$, and the generator is trained to maximize $\mathcal{L}_{\text{cls}}-\mathcal{L}_\text{GAN}$.

The discriminator loss of AC-GAN is not optimal. We give a slightly modified version below. Suppose the dataset owns $C$ categories, then the discriminator is trained to maximize
\begin{equation}
  \begin{aligned}
    \mathcal{L}_{{D}} = & \mathcal{L}_{\text{GAN}} +                                         \\
                        & \E \left[ \log P\left(\rg=c \mid \vx_{\text{real}}\right) \right]+
    \E \left[ \log P\left(\rg=C \mid \vx_{\text{fake}}\right) \right],
  \end{aligned}
\end{equation}
where $c \in \{0, 1, \dots, C-1\}$ is the ground truth class label of $\vx_{\text{real}}$, and $\rg=C$ means that $\vx_{\text{fake}}$ belongs to the fake class. To sum up, we use an additional class to represent the generated image. In practice, this is achieved by setting the dimension of the fully connected layer of the auxiliary classification layer to be $C+1$ rather than $C$.

The objective function of the generator is consistent with that of the original AC-GAN, \ie, maximizing
\begin{equation}
  \begin{aligned}
    \mathcal{L}_{{G}} =  - \mathcal{L}_{\text{GAN}} +
    \E \left[ \log P\left(\rg=c_{\text{fake}} \mid \vx_{\text{fake}}\right) \right],
  \end{aligned}
\end{equation}
where $c_{\text{fake}}$ is the class label used by the generator to generate $\vx_{\text{fake}}$.

We name this improved version of AC-GAN ImAC-GAN. As shown in the paper, ImAC-GAN is comparable to Omni-GAN, both of which achieve superior performance compared to projection-based cGANs. However, because ImAC-GAN uses cross-entropy as the loss function for classification, it can only handle the case where the sample has a positive label. Omni-GAN uses omni-loss, essentially a multi-label classification loss, which naturally supports handling samples with one positive label or multiple positive labels. We will give an example of generating images with multiple positive labels in Sec.~\ref{apx:sec:multi_label_D}.

\section{Technical Details of Omni-INR-GAN}
\label{apx:sec:omni_inr_gan}


Learning image prior model is helpful for image restoration and manipulation, such as denoising, inpainting, and harmonizing. Deep generative prior (DGP)~\cite{pan2020Exploiting} showed the potential of employing the generator prior captured by a pre-trained GAN model (\ie, a BigGAN model trained on a large-scale image dataset, ImageNet). However, BigGAN can only output images with a fixed aspect ratio, limiting the practical application of DGP. To make the pre-trained GAN model more flexible for downstream tasks, we propose a new GAN named Omni-INR-GAN, which can output images with any aspect ratio and any resolution.

\begin{figure}[t]
  \begin{center}
    \includegraphics[width=\linewidth]{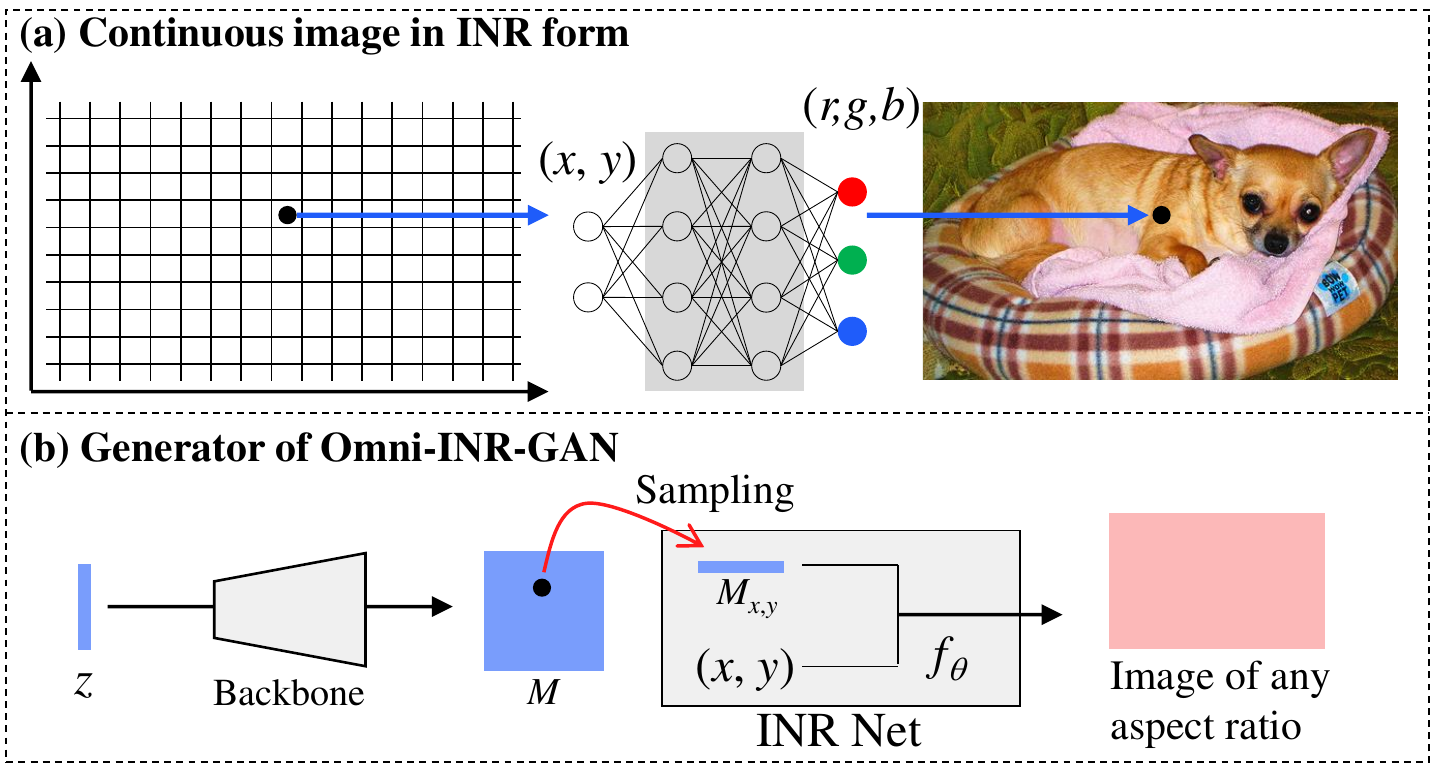}
  \end{center}
  \vspace{-0.5cm}
  \caption{(a) An example of an image represented in INR form. A fully connected network maps coordinates $(x, y)$ to pixel values $(r, g, b)$. (b) Using an INR network to enhance the generator so that the generator can output images with any resolution and any aspect ratio.}
  \label{apx:fig:omni_inr_gan}
\end{figure}

Images are usually represented by a set of pixels with fixed resolution. A popular method named implicit neural representation (INR) is prevalent in the 3D field~\cite{park2019DeepSDF,mescheder2019Occupancy,chen2019Learninga}. Recently, people introduced the INR method to 2D images~\cite{chen2020Learning,skorokhodov2020Adversarial}. As shown in Fig.~\ref{apx:fig:omni_inr_gan} (a), the INR of an image directly maps ($x$, $y$) coordinates to image's RGB pixel values. Since the coordinates are continuous, once we get the INR of an image, we can get images of arbitrary resolutions by sampling different numbers of coordinates.

Inspired by the local implicit image function (LIIF)~\cite{chen2020Learning}, we use INR to enhance Omni-GAN, with the goal of enabling the generator to output images with any aspect ratios and any resolution. We name our method Omni-INR-GAN.
As shown in Fig.~\ref{apx:fig:omni_inr_gan} (b), we keep the backbone of the generator network unchanged and employ an INR network for the output layer. Let $\mM \in \mathbb{R}^{C \times H \times W}$ represent the output feature map of the backbone, $f_\theta$ be the implicit neural function. Then the RGB signal at ($x$, $y$) coordinate is given by $s=f_\theta(\mM_{x, y}, x, y)$, where $\mM_{x, y}$ stands for the feature vector at ($x$, $y$). Note that since $x$ and $y$ can be any real numbers, $\mM_{x, y}$ may not exist in $\mM$. In such a case, we adopt the bilinear interpolation of the four feature vectors near ($x$, $y$) as the feature at ($x$, $y$).

Omni-INR-GAN can generate images with any aspect ratio, so as to be more friendly to downstream tasks like image restoration and manipulation. After trained on the large-scale dataset ImageNet, Omni-INR-GAN can be combined with DGP to do restoration tasks. Omni-INR-GAN eliminates cropping operations before image restoration, making it possible to repair the entire image directly. Since the generator has seen considerable natural images, utilizing the generator prior can facilitate downstream tasks significantly.

\section{An Example of Multi-label Discriminator}
\label{apx:sec:multi_label_D}

\begin{figure}[tbp]
  \centering
  \begin{minipage}[t]{\linewidth}
    \begin{subfigure}{0.49\linewidth}
      \centering
      \includegraphics[width=\linewidth]{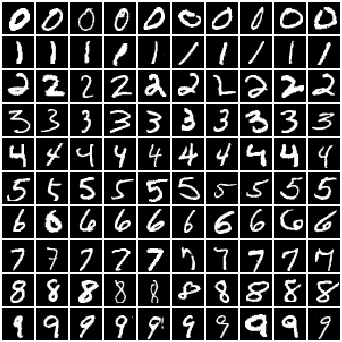}
      \caption{MNIST}
    \end{subfigure}
    \begin{subfigure}{0.49\linewidth}
      \centering
      \includegraphics[width=\linewidth]{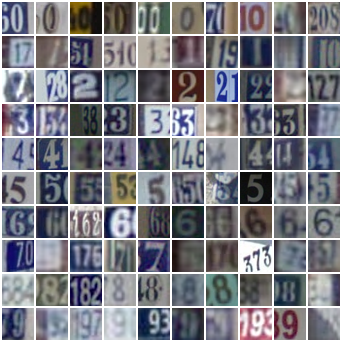}
      \caption{SVHN}
    \end{subfigure}
    \vspace{-0.3cm}
    \caption{Real images sampled from the dataset.}
    \vspace{0.5cm}
    \label{apx:fig:mnist_svhn}
  \end{minipage}\\
  \begin{minipage}[t]{\linewidth}
    \begin{subfigure}{0.49\linewidth}
      \centering
      \includegraphics[width=\linewidth]{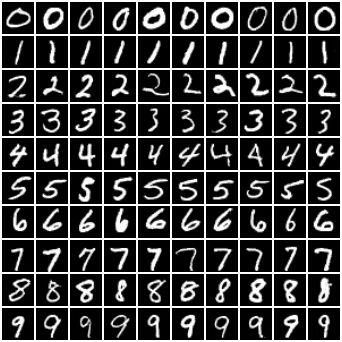}
      \caption{MNIST}
    \end{subfigure}
    \begin{subfigure}{0.49\linewidth}
      \centering
      \includegraphics[width=\linewidth]{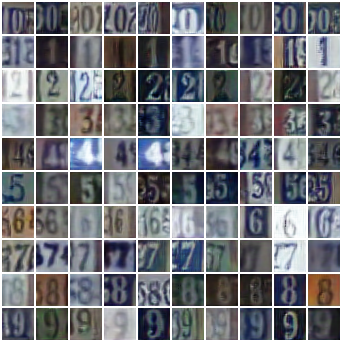}
      \caption{SVHN}
    \end{subfigure}
    \vspace{-0.3cm}
    \caption{Images generated by a generator which is guided by a multi-label discriminator.}
    \vspace{0.5cm}
    \label{apx:fig:generated_mnist_svhn}
  \end{minipage}
\end{figure}

Omni-loss is essentially a multi-label classification loss and naturally supports classification with multiple positive labels. To verify the ability of Omni-GAN for generating samples with multiple positive labels, we construct a mixed dataset containing images of digits from two distinct domains, namely MNIST~\cite{lecun1998gradient} of handwritten digits and SVHN~\cite{netzer2011reading} of house numbers. Some example images from the datasets are shown in Fig.~\ref{apx:fig:mnist_svhn}. In this setting, the discriminator needs to predict three attributes, class (recognizing digits), domain, and reality.

Let us take images of MNIST as an example, and show how to set the loss for the discriminator. As for SVHN, the case is analogous. Suppose $\vx_{\text{real}}$ is an image sampled from MNIST, its multi-label vector is given by
\begin{equation}
  \vy_{\text{real}} = [\underbrace{-1, \dots, 1_{\text{gt}}, \dots, -1}_{\text{class}}, \underbrace{1_{\text{mnist}}, -1}_{\text{domain}}, \underbrace{1_{\text{real}}, -1}_{\text{reality}}],
  \label{apx:equ:oneside-multi_label_real}
\end{equation}
where $-1$ means the corresponding score belongs to the negative set, and $1$ to the positive set. As can be seen, $\vy_{\text{real}}$ possesses three positive labels. The multi-label vector for $\vx_{\text{fake}}$ is then given by
\begin{equation}
  \vy_{\text{fake}} = [\underbrace{-1, \dots, -1, \dots, -1}_{\text{class}}, \underbrace{-1, -1}_{\text{domain}}, \underbrace{-1, 1_{\text{fake}}}_{\text{reality}}],
  \label{apx:equ:oneside-multi_label_fake}
\end{equation}
which is a one-hot vector with the last element being $1$. The discriminator loss is given by
\begin{equation}
  \begin{aligned}
    \mathcal{L}_{D}
    = & \E_{\vx_{\text{real}} \sim p_{\text{d}}} \left[\mathcal{L}_{\text {omni}}\left(\vx_{\text{real}}, \vy_{\text{real}}\right)\right]    \\
      & + \E_{\vx_{\text{fake}} \sim p_{\text{g}}} \left[\mathcal{L}_{\text {omni}}\left(\vx_{\text{fake}}, \vy_{\text{fake}}\right)\right].
  \end{aligned}
  \label{apx:equ:D_loss}
\end{equation}

For generator, its goal is to cheat the discriminator. The multi-label vector for $\vx_{\text{fake}}$ is given by
\begin{equation}
  \vy_{\text{fake}}^{(\text{G})} = [\underbrace{-1, \dots, 1_{\text{G}}, \dots, -1}_{\text{class}}, \underbrace{1_{\text{mnist}}, -1}_{\text{domain}}, \underbrace{1_{\text{real}}, -1}_{\text{reality}}],
  \label{apx:equ:oneside-multi_label_G}
\end{equation}
where $1_{\text{G}}$ is $1$ if its index in the vector is equal to the label adopted by the generator to generate $\vx_{\text{fake}}$, otherwise $-1$. The generator loss is given by.
\begin{equation}
  \begin{aligned}
    \mathcal{L}_{G}
    = \E_{\vx_{\text{fake}} \sim p_{\text{g}}} \left[\mathcal{L}_{\text {omni}}\left(\vx_{\text{fake}}, \vy_{\text{fake}}^{(\text{G})}\right)\right].
  \end{aligned}
  \label{apx:equ:G_loss}
\end{equation}

We experimentally found that this multi-label discriminator can instruct the generator to generate images from different domains. Some generated images are shown in Fig.~\ref{apx:fig:generated_mnist_svhn}. We must emphasize that this is only a preliminary experiment to verify the function of the multi-label discriminator. We look forward to applying the multi-label discriminator to other tasks in the future, such as translation between images in different domains, domain adaptation, \etc.

\section{Additional Results on CIFAR}
\label{apx:sec:results_cifar}

\subsection{Over-fitting of the Discriminator}

\begin{figure}[!t]
  \begin{subfigure}{0.49\linewidth}
    \centering
    \includegraphics[width=\linewidth]{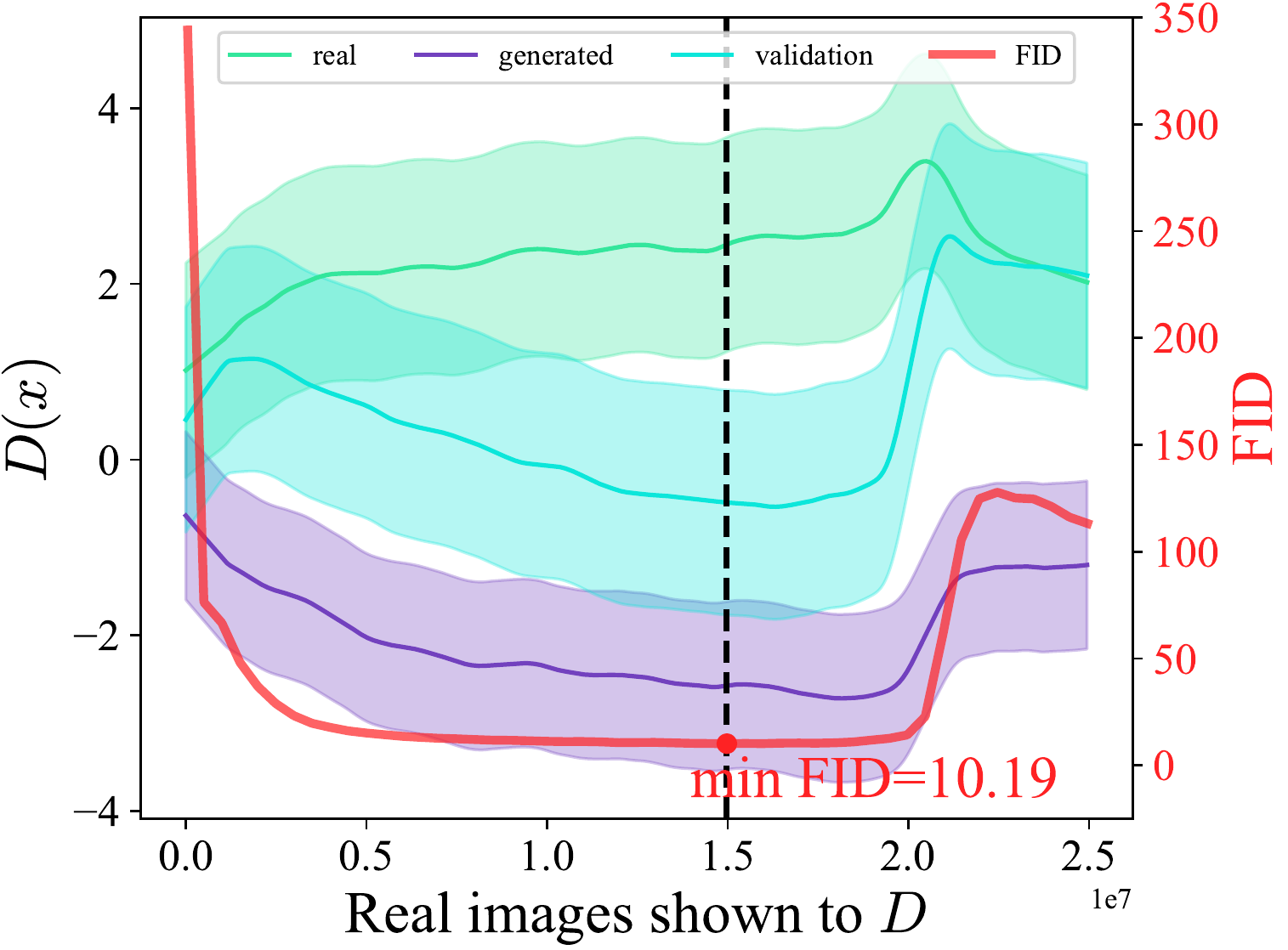}
    \caption{}
  \end{subfigure}
  \begin{subfigure}{.49\linewidth}
    \centering
    \includegraphics[width=\linewidth]{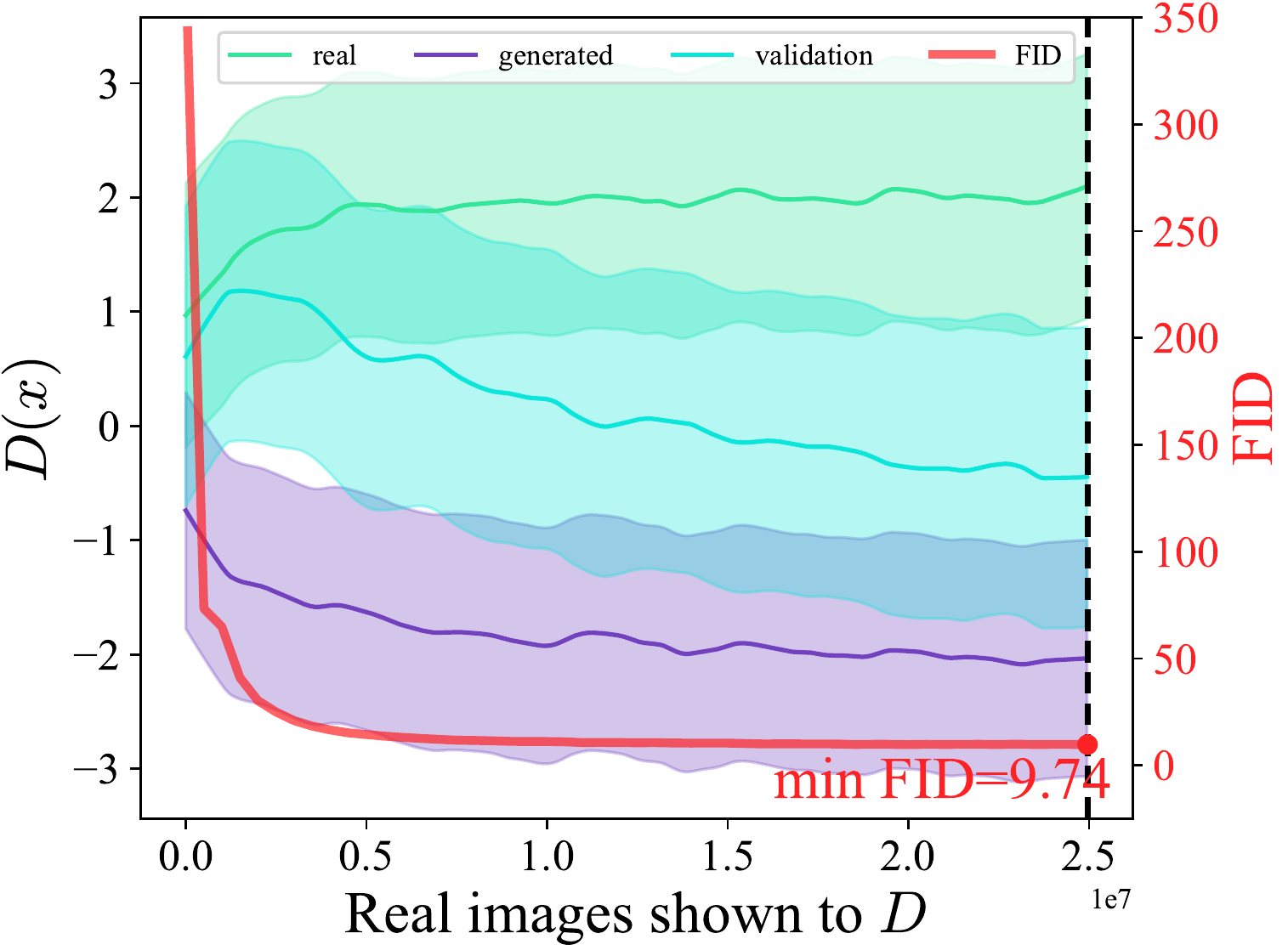}
    \caption{}
  \end{subfigure}
  \vspace{-0.3cm}
  \caption{The raw logits of $D(\vx)$ and the corresponding FID score of a projection-based cGAN are plotted in the same figure. The black dashed line indicates where the minimum FID is reached. (a) training without weight decay. (b) training with weight decay.  The figures of $D(\vx)$ are inspired by~\cite{karras2020Training}.}
  \label{apx:fig:early_collapse}
\end{figure}

Karras \etal~\cite{karras2020Training} found that the discriminator overfits the training dataset, which will lead to incorrect gradients provided to the generator. Thus the training diverges. To verify that the collapse of the projection-based cGAN is due to the over-fitting of the discriminator, we plotted the scalar output of the discriminator, $D(\vx)$, over the course of training. We utilized the test set of CIFAR100 containing $10,000$ images as the verification set, which was not used in the training.

As shown in Fig.~\ref{apx:fig:early_collapse}a, obviously, as training progresses, the $D(\vx)$ of the validation set tends to that of the generated images, substantiating that the discriminator overfits the training data. We also plotted the FID curve in the same figure. We can see that the training commences diverging when showing about $20$M real images (\ie, around $400$ epoch) to the discriminator. The best FID is obtained when approximately $15$M real images are shown to the discriminator.

In Fig.~\ref{apx:fig:early_collapse}b, we show the $D(\vx)$ and FID after applying weight decay to the projection-based discriminator. We can find that although the discriminator still overfits the training data, the training dose not collapse during the whole training process (the minimum FID, $9.74$, is reached at the end of the training).


\subsection{Comparison of One-sided Omni-GAN and Projection-based GAN on CIFAR10}
\label{apx:sec:one_side}

\begin{figure}[!t]
  \begin{subfigure}{0.49\linewidth}
    \centering
    \includegraphics[width=\linewidth]{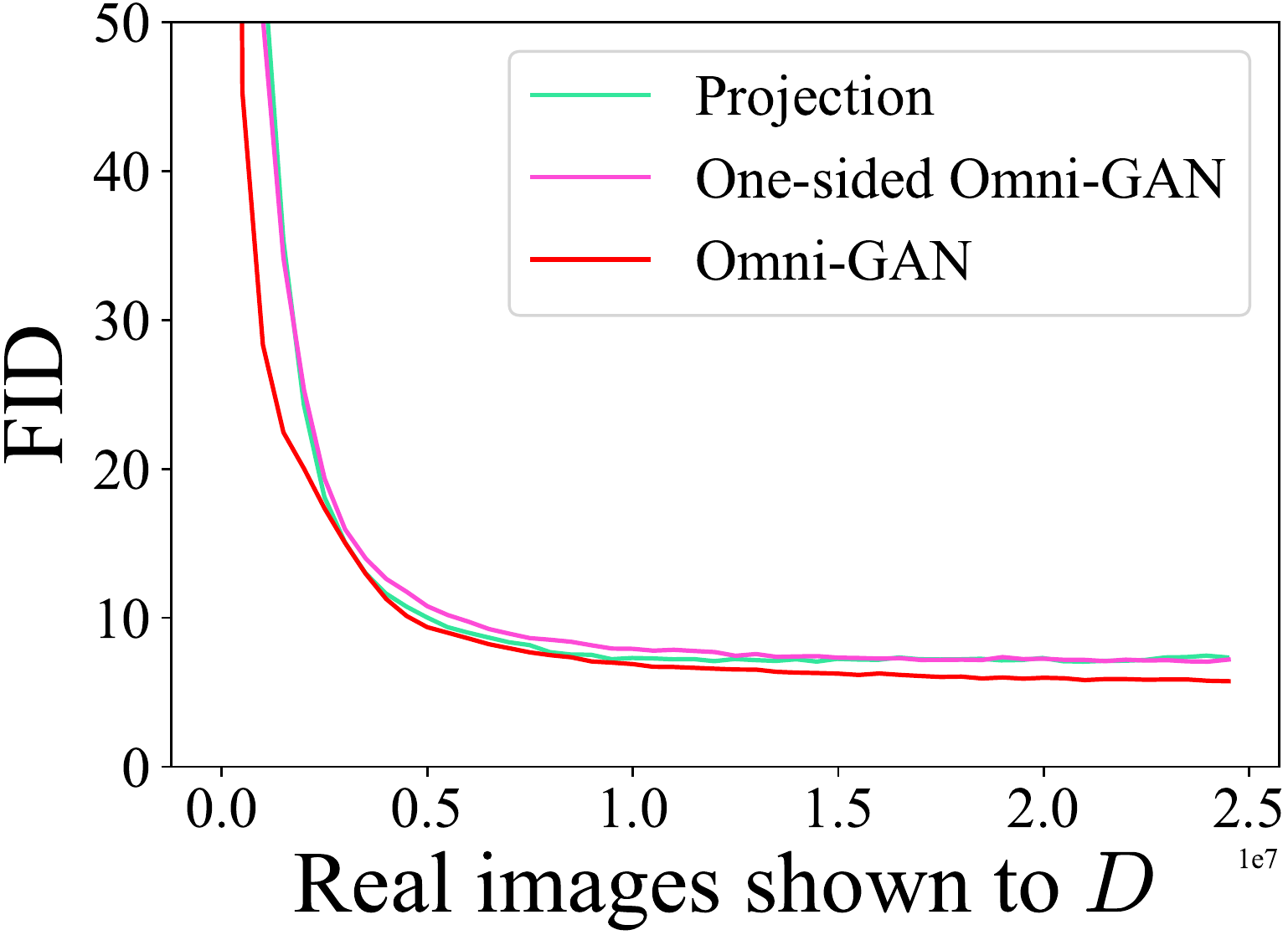}
    \caption{}
  \end{subfigure}
  \begin{subfigure}{0.49\linewidth}
    \centering
    \includegraphics[width=\linewidth]{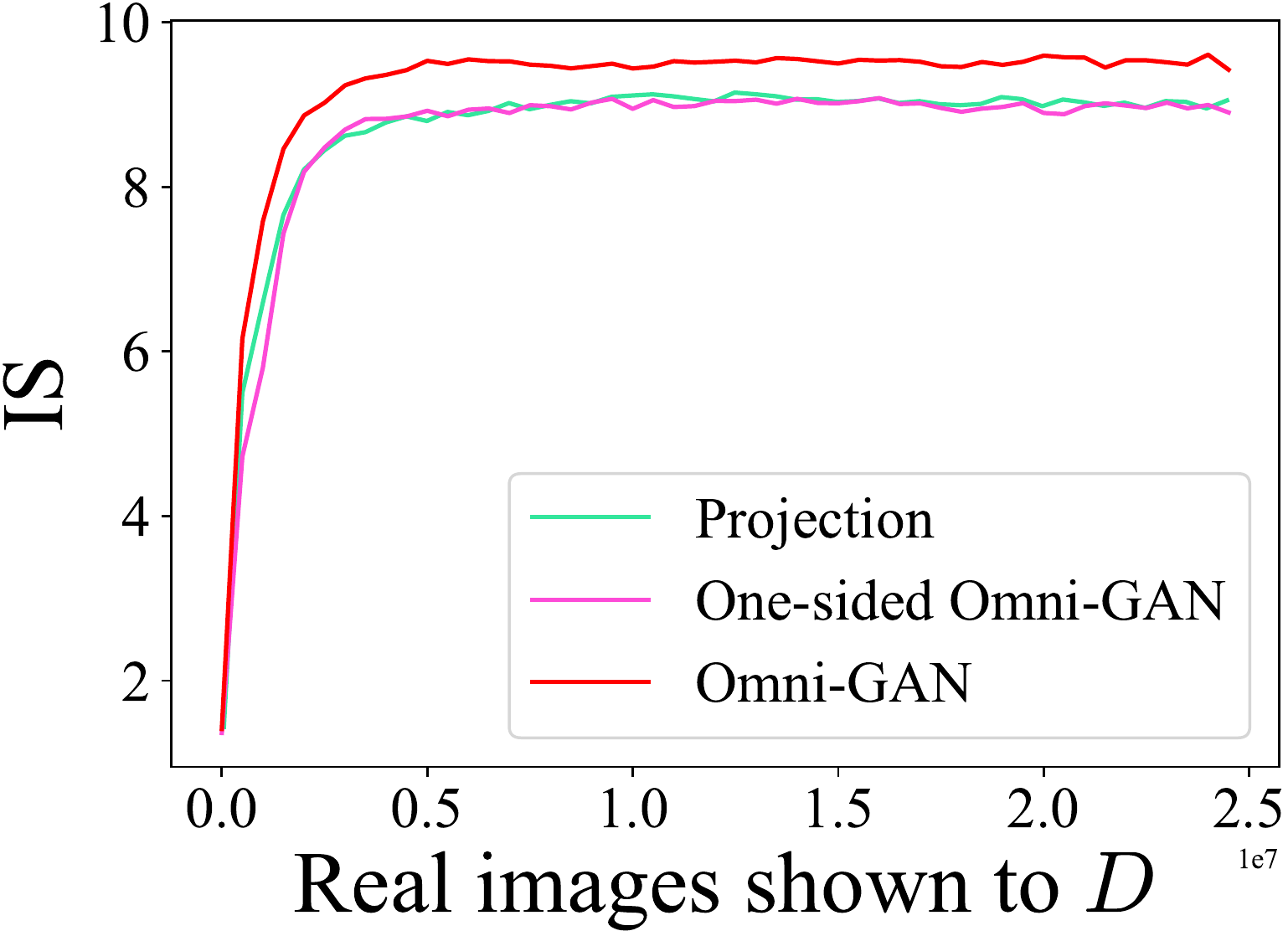}
    \caption{}
  \end{subfigure}
  \caption{One-sided Omni-GAN is on par with projection-based BigGAN on CIFAR10, proving that one-sided Omni-GAN indeed belongs to projection-based cGANs. Both of them are inferior to the classification-based cGAN, Omni-GAN.}
  \label{apx:fig:oneside_c10}
\end{figure}

We provide the results of one-sided Omni-GAN and projection-based BigGAN on CIFAR10. As shown in Fig.~\ref{apx:fig:oneside_c10}, one-sided Omni-GAN is comparable to the projection-based BigGAN in terms of both FID and IS. This proves that one-sided Omni-GAN indeed belongs to projection-based cGANs. Both one-sided Omni-GAN and projection-based BigGAN are inferior to Omni-GAN. Because the only difference between one-sided Omni-GAN and Omni-GAN is whether the supervision is fully utilized, we conclude that the superiority of Omni-GAN lies in the full use of supervision.

\subsection{Comparison with Multi-hinge GAN}
\label{sec:comparison_multihinge}

\begin{figure}[t]
  \centering
  \begin{subfigure}{0.49\linewidth}
    \includegraphics[width=\linewidth]{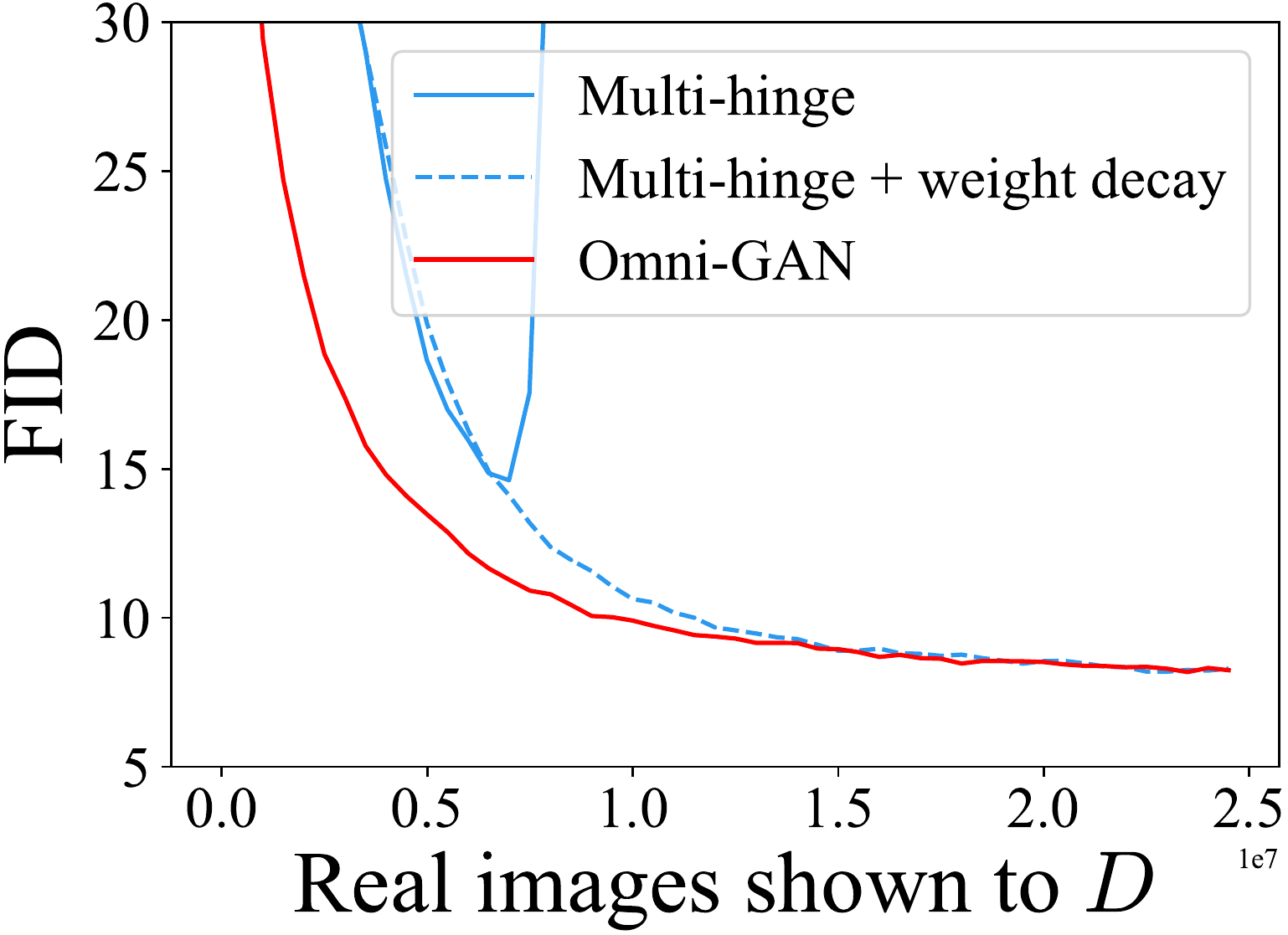}
    \caption{}
  \end{subfigure}
  \begin{subfigure}{0.49\linewidth}
    \centering
    \includegraphics[width=\linewidth]{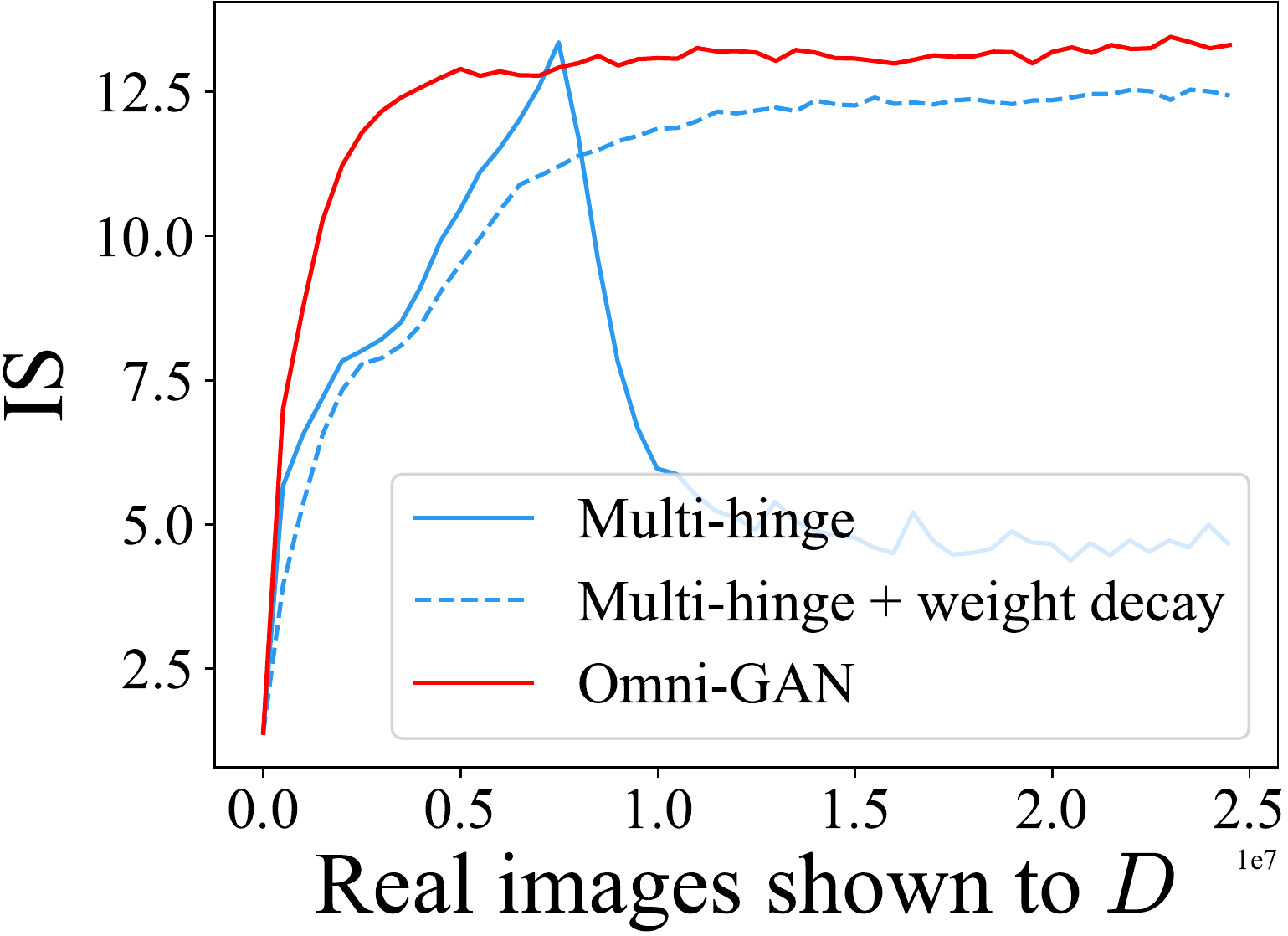}
    \caption{}
  \end{subfigure}
  \caption{FID and IS on CIFAR100. Weight decay can eliminate the early collapse problem of Multi-hinge GAN on CIFAR100.}
  \label{apx:fig:multihinge_wd_c100}
\end{figure}

Multi-hinge GAN belongs to classification-based cGANs, and also suffers from the early collapse issue. We study whether weight decay is effective for Multi-hinge GAN. As shown in Fig.~\ref{apx:fig:multihinge_wd_c100}, original Multi-hinge GAN suffers a severe early collapse issue. After equipped with weight decay, Multi-hinge GAN enjoys a safe optimization and its FID is even comparable to that of Omni-GAN. However, its IS is worse than that of Omni-GAN.

\begin{figure}[t]
  \begin{subfigure}{0.49\linewidth}
    \centering
    \includegraphics[width=\linewidth]{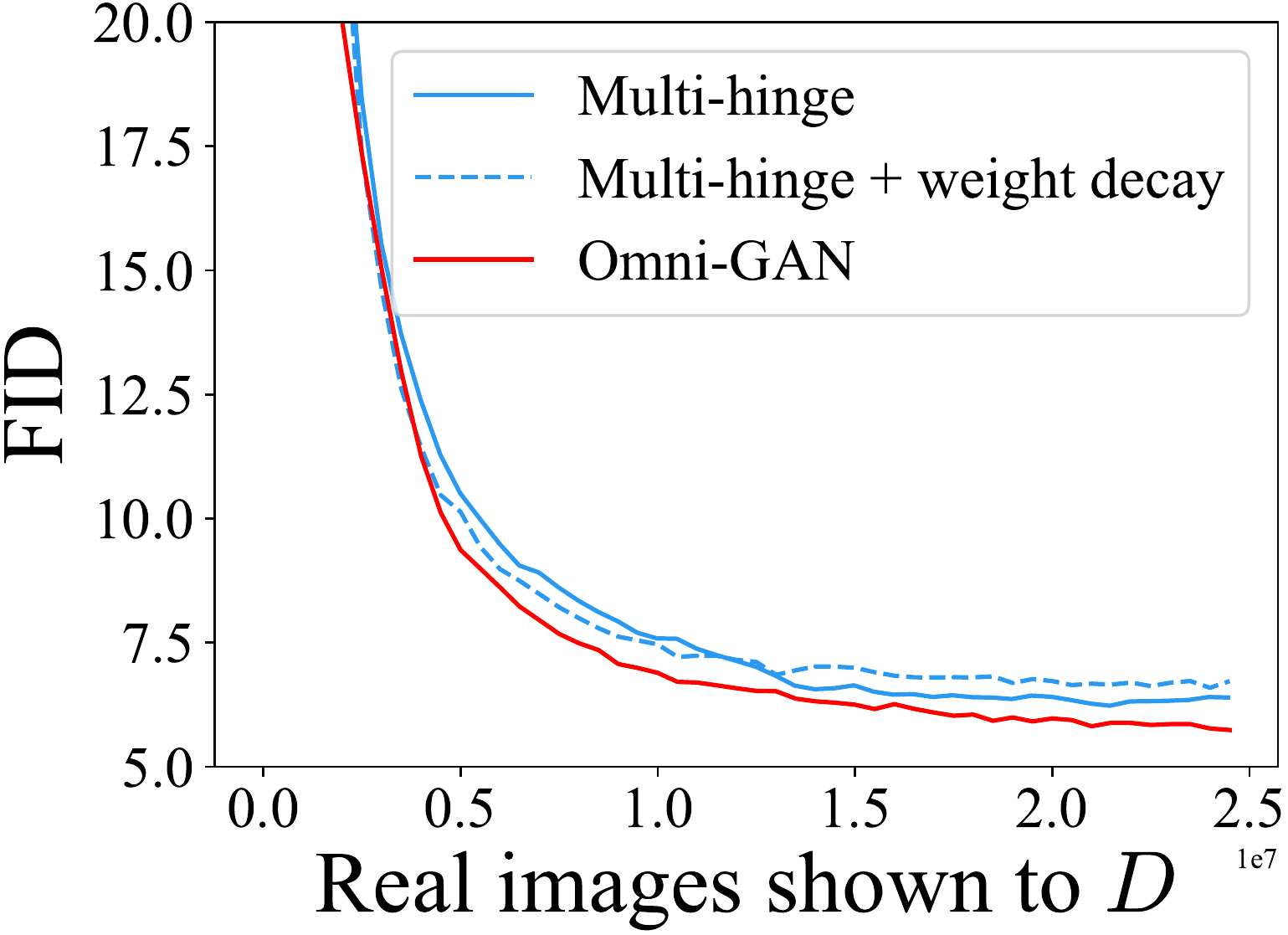}
    \caption{}
  \end{subfigure}
  \begin{subfigure}{0.49\linewidth}
    \centering
    \includegraphics[width=\linewidth]{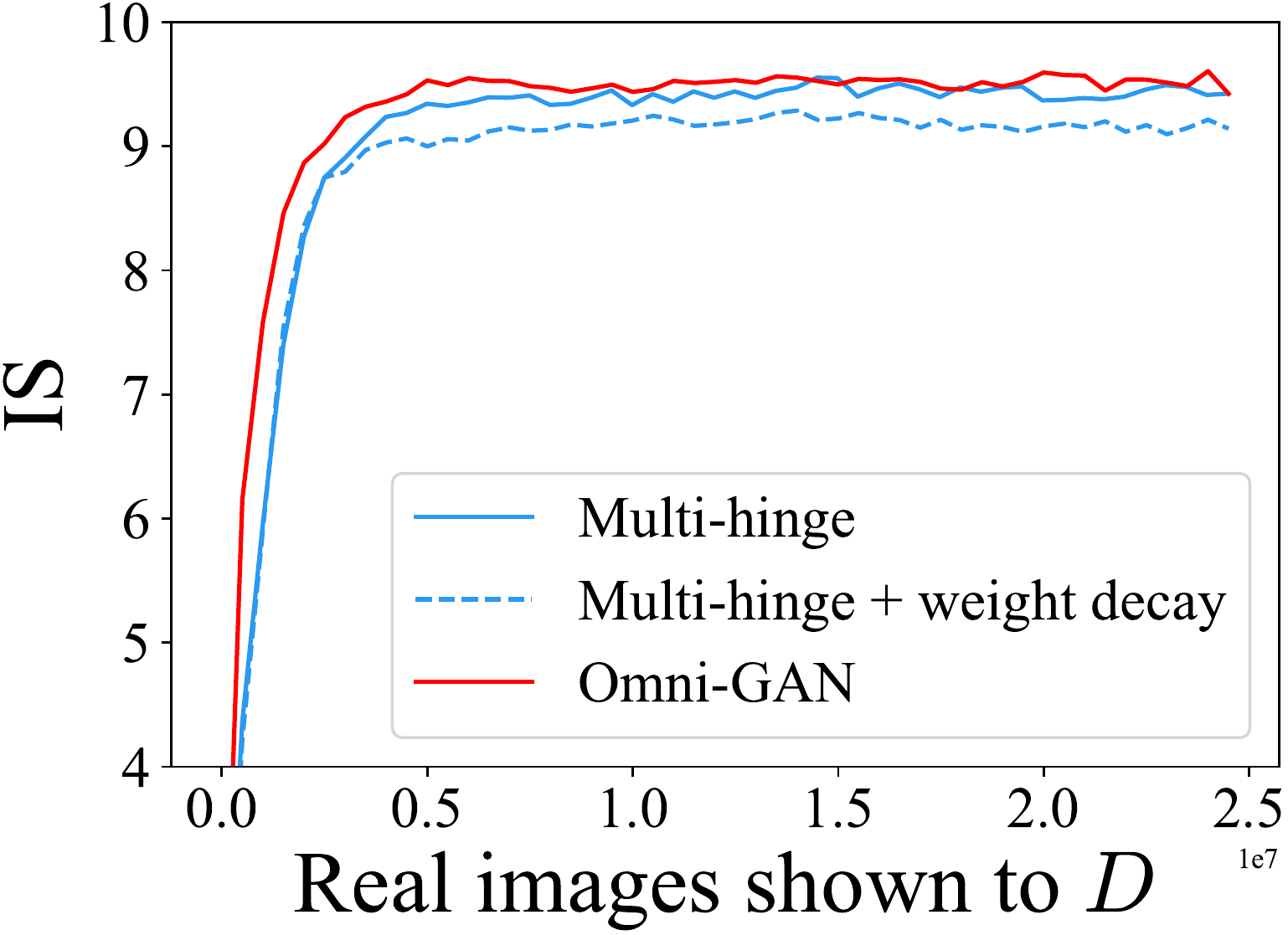}
    \caption{}
  \end{subfigure}
  \caption{FID and IS on CIFAR10. Weight decay deteriorates Multi-hinge GAN.}
  \label{apx:fig:multihinge_wd_c10}
\end{figure}

Multi-hinge GAN combined with weight decay does not always perform well. The results on CIFAR10 are shown in Fig.~\ref{apx:fig:multihinge_wd_c10}. Weight decay deteriorates Multi-hinge GAN in terms of both FID and IS. However, Omni-GAN outperforms Multi-hinge GAN. In addition, omni-loss is more flexible than multi-hinge loss. It supports implementing a multi-label discriminator. As a result, we suggest first considering using Omni-GAN when choosing cGANs.


\subsection{Applying Weight Decay to the Generator}
\label{apx:sec:weight_decay_generator}

\begin{figure}[t]
  \centering
  \centering
  \includegraphics[width=0.7\linewidth]{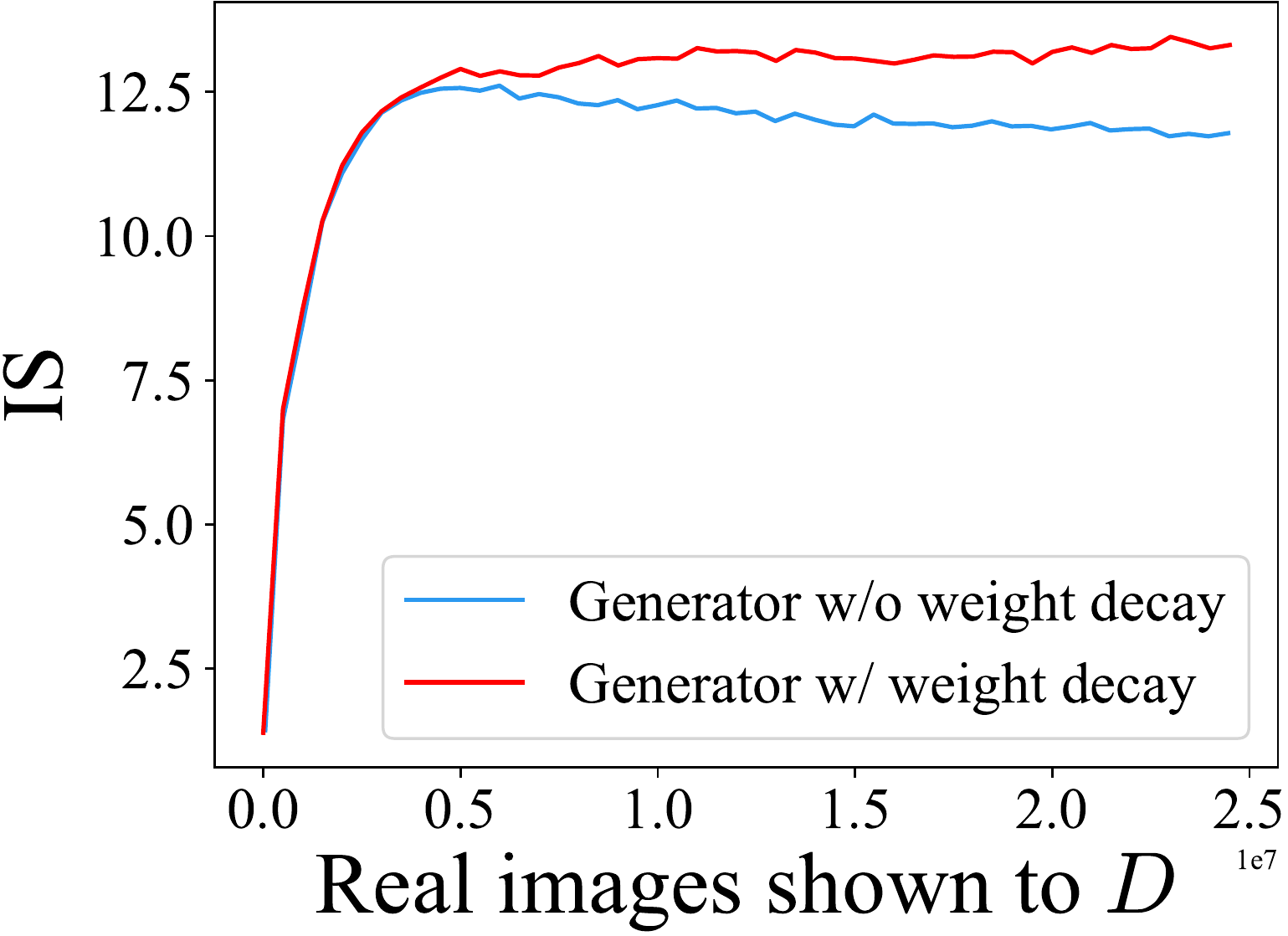}
  \caption{Applying weight decay to the generator. Applying weight decay to the discriminator helps alleviate the collapse issue, but the IS gradually decreases as the training progresses. Applying weight decay to the generator simultaneously solves this problem. Experiments are conducted on CIFAR100.}
  \label{apx:fig:generator_wd}
\end{figure}


We found empirically that applying weight decay also to the generator can make training more stable. As shown in Fig.~\ref{apx:fig:generator_wd}, although only applying weight decay to the discriminator can avoid the risk of collapse earlier, the IS has a trend of gradually decreasing as the training progresses. Fortunately, applying weight decay (set to be $0.001$ in our most experiments) to the generator can solve this problem. This phenomenon seems to indicate that the generator is also at a risk of over-fitting.



\subsection{How to Set the Weight Decay?}

We did a grid search for the weight decay on CIFAR and found that its value is related to the size of the training dataset. For CIFAR100, there are only $500$ images per class, and the weight decay is set to be $0.0005$. For CIFAR10, there are $5000$ images per class, and the weight decay is set to be $0.0001$. For ImageNet, it is a large dataset with a considerable number of training data (approximate $1.2$M). The weight decay is set to 0.00001. The conclusion is that the smaller the dataset, the higher the risk of over-fitting for the discriminator. Then weight decay should be larger.

\section{Additional Results on ImageNet}
\label{apx:sec:curves_imagenet}

\begin{figure}[t]
  \centering
  \begin{subfigure}{0.49\linewidth}
    \centering
    \includegraphics[width=\linewidth,height=3.3cm]{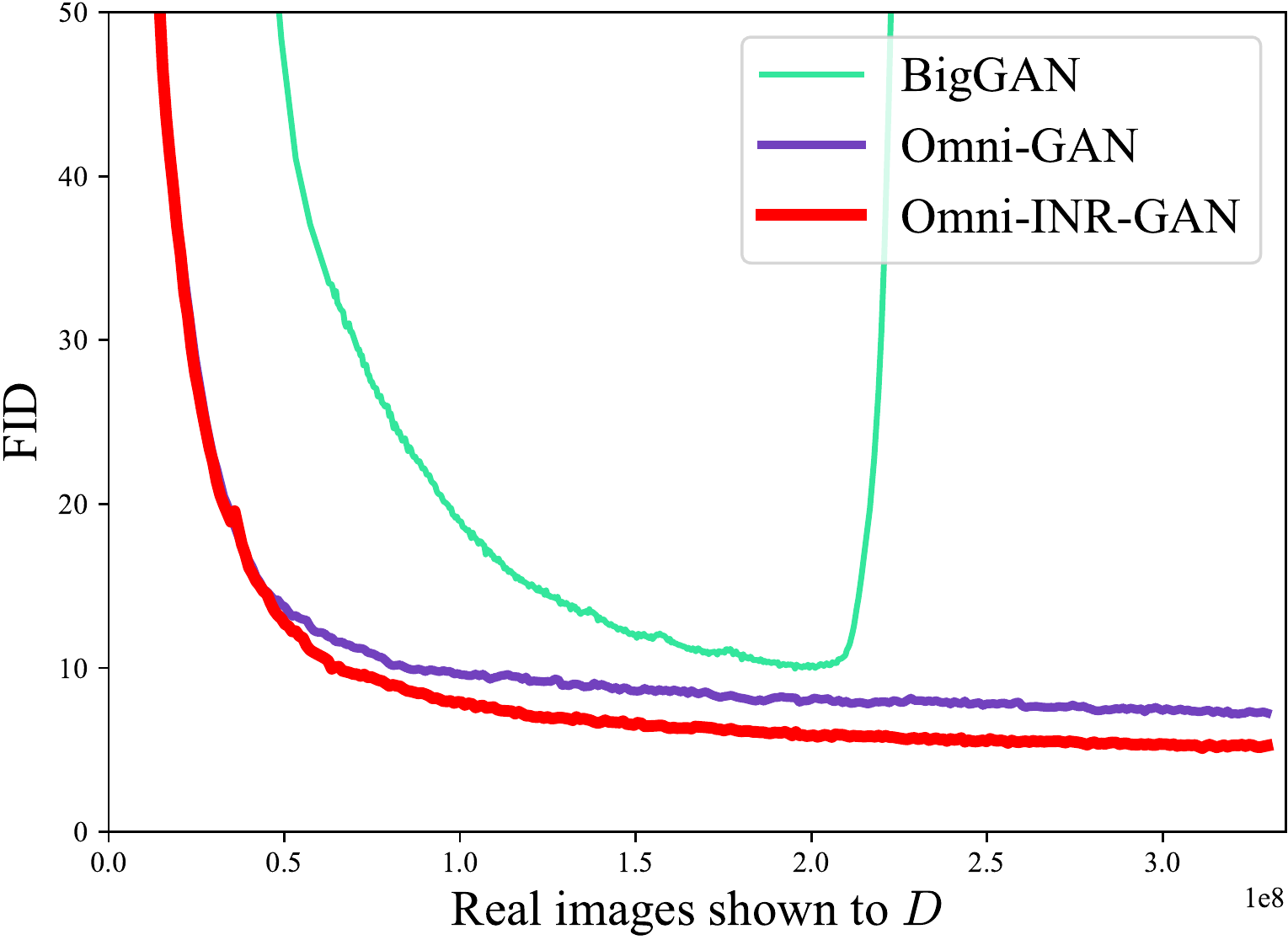}
    \caption{FID on ImageNet $256\times256$}
  \end{subfigure}
  \begin{subfigure}{0.49\linewidth}
    \centering
    \includegraphics[width=\linewidth,height=3.3cm]{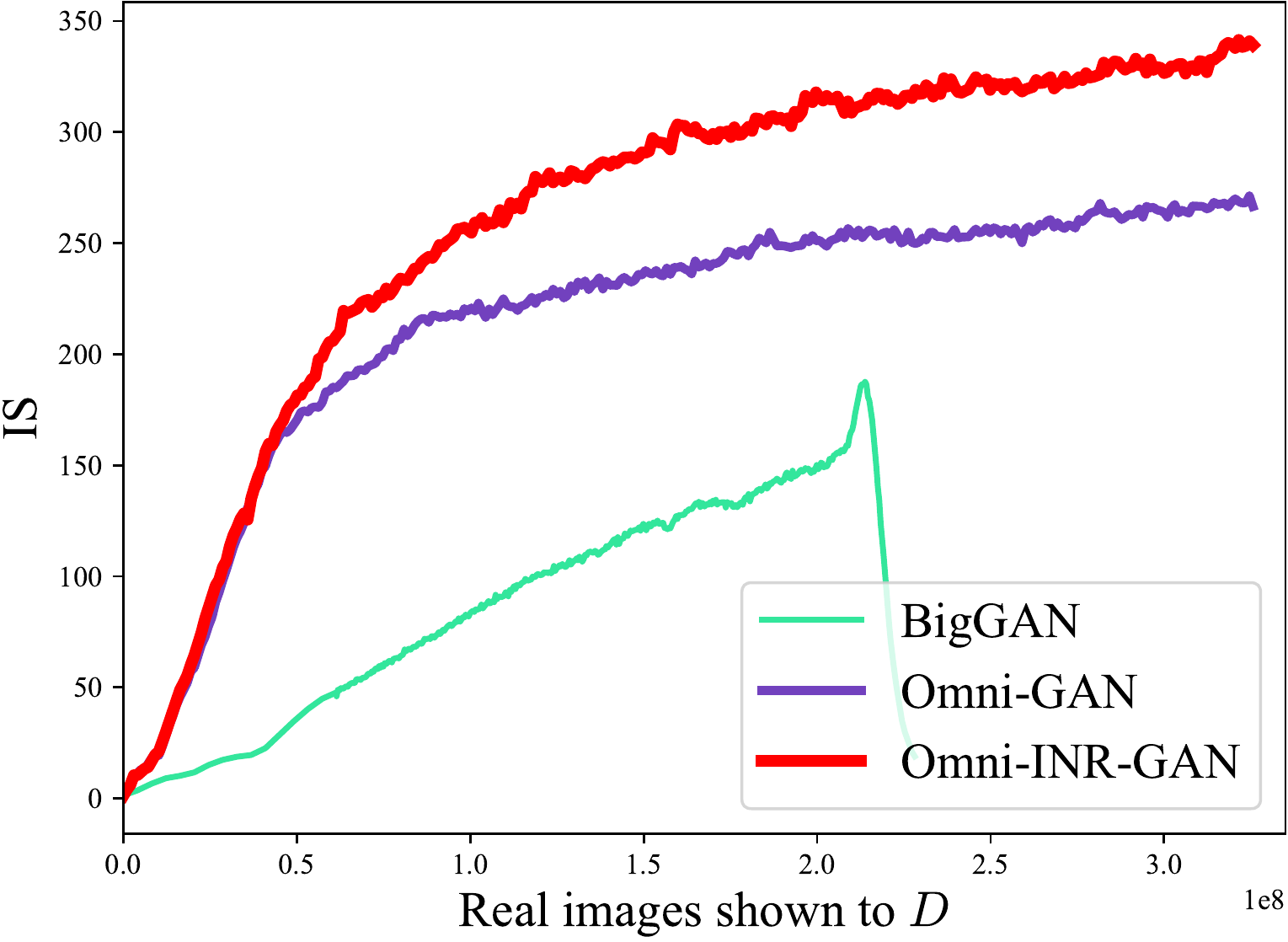}
    \caption{IS on ImageNet $256\times256$}
  \end{subfigure}
  \vspace{-0.2cm}
  \caption{FID and IS on ImgeNet $256\times256$. Omni-GAN and Omni-INR-GAN converge faster than the projection-based BigGAN. Omni-INR-GAN clearly outperforms Omni-GAN, showing its significant potential for future applications.}
  \label{apx:fig:imagenet256}
\end{figure}

\begin{figure}[t]
  \centering
  \centering
  \includegraphics[width=0.7\linewidth]{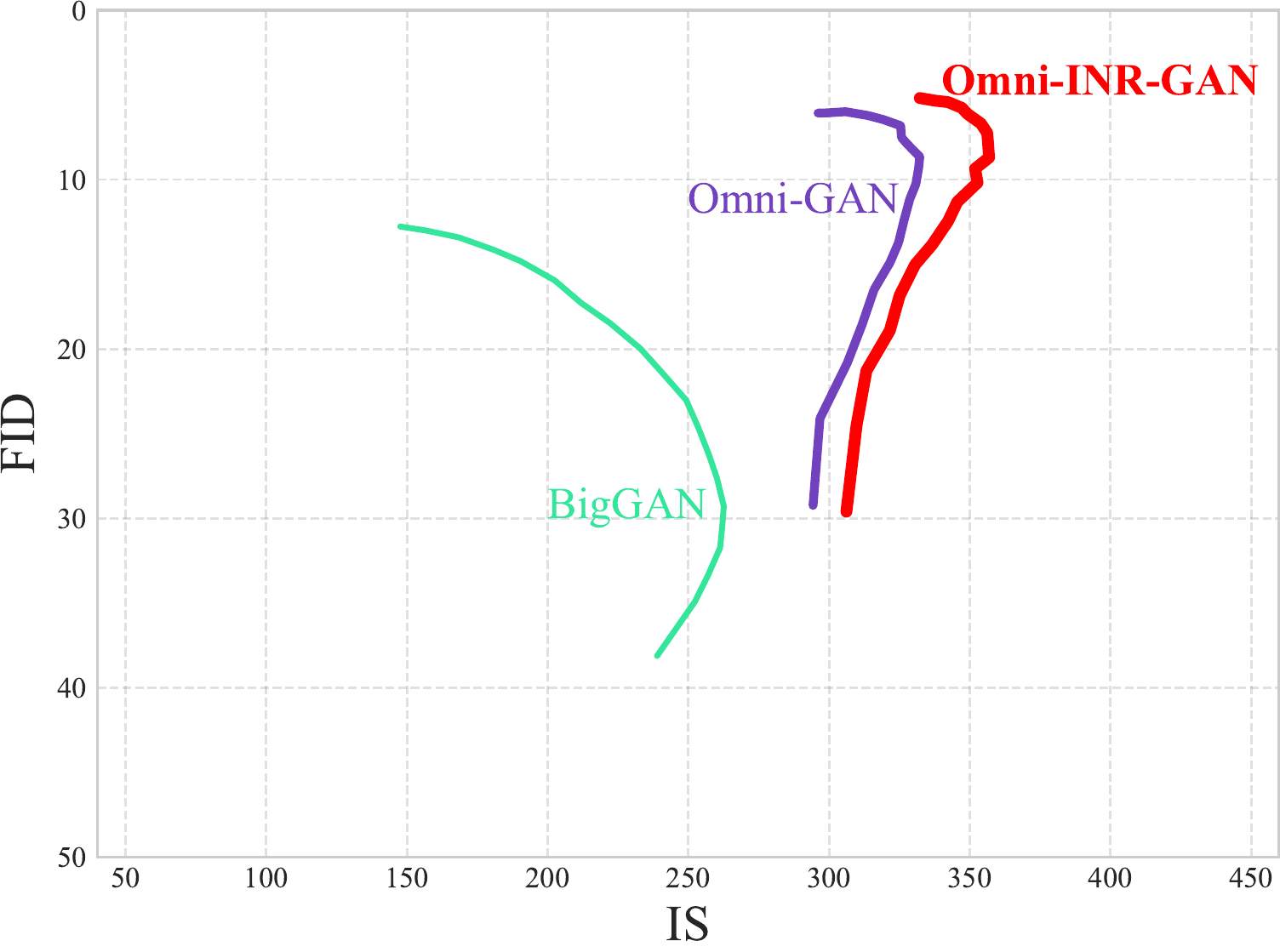}
  \caption{Tradeoff curves using truncation trick on ImageNet $256\times256$. We show truncation values from $\sigma=0.05$ to $\sigma=1$ with step being $0.05$. Omni-INR-GAN outperforms Omni-GAN and BigGAN.}
  \label{apx:fig:truncation_imagenet256.}
\end{figure}

We provide convergence curves on ImageNet $256\times256$. As shown in Fig.~\ref{apx:fig:imagenet256}, both Omni-GAN and Omni-INR-GAN converge faster than BigGAN, proving the effectiveness of combining strong supervision and weight decay. Omni-INR-GAN clearly outperforms Omni-GAN, showing its significant potential for future applications. In Fig.~\ref{apx:fig:truncation_imagenet256.}, we show the tradeoff curve of these methods using the truncation trick on ImageNet $256\times256$. Omni-INR-GAN is consistently superior to Omni-GAN and BigGAN.

\section{Application to Image-to-Image Translation}
\label{apx:sec:imagetoimage}

\begin{table*}[t]
  \centering
  \resizebox{0.9\textwidth}{!}{%
    \begin{tabular}{ccccccccccc}
      \toprule
      \multirow{4}{*}{SPADE~\cite{park2019Semantic}} & \textit{road} & \textit{sidewalk} & \textit{building} & \textit{wall} & \textit{fence} & \textit{pole} & \textit{traffic light} & \textit{traffic sign} & \textit{vegetation} & \textit{terrain} \\
                                                     & $97.44$       & $79.89$           & $87.86$           & $50.57$       & $47.21$        & $35.90$       & $38.97$                & $44.67$               & $88.15$             & $66.14$          \\
      \cmidrule{2-11}
                                                     & \textit{sky}  & \textit{person}   & \textit{rider}    & \textit{car}  & \textit{truck} & \textit{bus}  & \textit{train}         & \textit{motorcycle}   & \textit{bicycle}    & mIoU             \\
                                                     & $91.61$       & $62.27$           & $38.67$           & $88.68$       & $64.96$        & $70.17$       & $41.42$                & $28.58$               & $58.86$             & $62.21$          \\
      \midrule
      \multirow{4}{*}{+ Omni-GAN}                    & \textit{road} & \textit{sidewalk} & \textit{building} & \textit{wall} & \textit{fence} & \textit{pole} & \textit{traffic light} & \textit{traffic sign} & \textit{vegetation} & \textit{terrain} \\
                                                     & $97.57$       & $81.62$           & $88.58$           & $53.39$       & $50.47$        & $35.88$       & $41.08$                & $46.75$               & $89.31$             & $67.00$
      \\
      \cmidrule{2-11}
                                                     & \textit{sky}  & \textit{person}   & \textit{rider}    & \textit{car}  & \textit{truck} & \textit{bus}  & \textit{train}         & \textit{motorcycle}   & \textit{bicycle}    & mIoU             \\
                                                     & $92.14$       & $63.97$           & $41.99$           & $89.91$       & $71.06$        & $74.21$       & $56.16$                & $33.99$               & $61.23$             & $\mathbf{65.07}$ \\
      \bottomrule
    \end{tabular}%
  }
  \vspace{-0.2cm}
  \caption{Semantic image synthesis using SPADE. Replacing the GAN used by SPADE with Omni-GAN can improve the quality of synthesized images.}
  \vspace{-0.4cm}
  \label{tab:spade}
\end{table*}

\begin{figure}[t]
  \begin{center}
    \includegraphics[width=\linewidth]{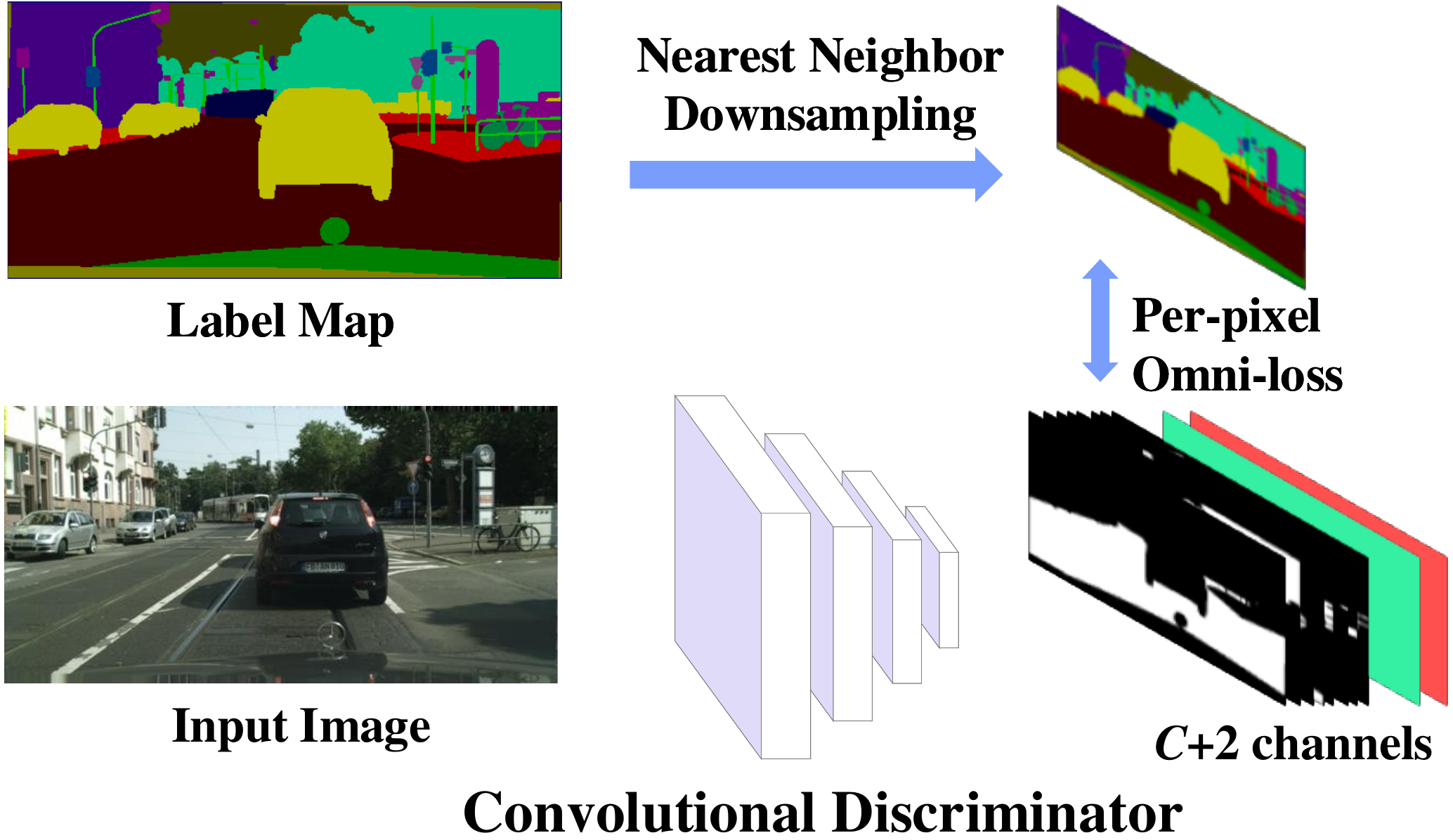}
  \end{center}
  \vspace{-0.5cm}
  \caption{Combine omni-loss with a fully convolutional discriminator whose outputs are feature maps. In the figure, the green and red feature maps represent scores that the input images are real and fake, respectively. Omni-loss is applied to the output feature maps pixel-by-pixel. }
  \label{apx:fig:per_pixel_omni_loss}
\end{figure}

Omni-GAN can be used for image-to-image translation tasks. We verify the effectiveness of Omni-GAN on semantic image synthesis~\cite{wang2018VideotoVideo,qi2018Semiparametric}. In particular, we replace the GAN loss of SPADE~\cite{park2019Semantic} with Omni-GAN's loss, and keep other hyper-parameters unchanged.
The discriminator is a fully convolutional network, which is widely adopted by image-to-image translation tasks~\cite{park2019Semantic,isola2017ImagetoImage,wang2018HighResolution}. As shown in Fig.~\ref{apx:fig:per_pixel_omni_loss}, the discriminator takes images as input and outputs feature maps with the number of channels being $C+2$. $C$ represents the number of classes which is analogous to that of the semantic segmentation task. $2$ indicates there are two extra feature maps representing to what extent the input image is real or fake. We adopt nearest neighbor downsampling to downsample the label map to the same resolution as the output feature maps of the discriminator. Then we use the downsampled label map as the ground truth label, and apply a per-pixel omni-loss to the output feature maps of the discriminator.

We use Cityscapes dataset~\cite{cordts2016Cityscapes} as a testbed, and train models on the training set with size of $2,975$. The images is resized to $256\times512$. Models are evaluated by the mIoU of the generated images on the test set with $500$ images. We use a pre-trained DRN-D-105~\cite{yu2017Dilated} as the segmentation model for the sake of evaluation. As shown in Table~\ref{tab:spade}, Omni-GAN improves the mIoU score of SPADE from $62.21$ to $65.07$, substantiating that the synthesized images possess more semantic information. We believe that the improvement comes from the improved ability of the discriminator in distinguishing different classes, so that the generator receives better guidance and thus produces images with richer semantic information.

\section{Application to Downstream Tasks}
\label{apx:sec:downstream_tasks}


\subsection{Colorization and Super-resolution}

Deep generative prior (DGP)~\cite{pan2020Exploiting} showed the potential of employing image prior captured by a pre-trained GAN model. Our colorization and super-resolution schemes are based on DGP. We first introduce the preliminary knowledge of DGP.

Suppose $\vx$ is a natural image and $\phi$ is a degradation function, \eg, gray transform for colorization and down-sampling for super-resolution. Then $\hat{\vx}=\phi{(\vx)}$ represents the degraded image, \ie, a partial observation of the original image, $\vx$. The goal of image restoration is to recover $\vx$ from $\hat{\vx}$ with the help of some statistical image prior of $\vx$. DGP proposes employing the image prior stored in a pre-trained GAN's generator. The objective is defined as
\begin{equation}
  \begin{aligned}
    \boldsymbol{\theta}^{*}, \vz^{*}=
    \underset{\boldsymbol{\theta}, \vz}{\arg \min } \mathcal{L}(\hat{\vx}, \phi(G(\vz ; \boldsymbol{\theta}))),
  \end{aligned}
  \label{apx:equ:dgp_objective}
\end{equation}
where $\vz$ is a noise vector. $G$ represents the generator in GAN and is parameterized by $\boldsymbol{\theta}$. $\mathcal{L}$ is a discriminator-based distance metric: $\mathcal{L}\left(\mathbf{x}_{1}, \mathbf{x}_{2}\right)=\sum_{i \in \mathcal{I}}\left\|D\left(\mathbf{x}_{1}, i\right), D\left(\mathbf{x}_{2}, i\right)\right\|_{1}$. $D$ is the discriminator coupled with $G$. $\mathcal{I}$ is a index set for feature maps of different blocks of $D$. Note that both $G$ and $D$ have been trained on a large-scale natural image dataset. DGP employs the prior of $G$ by fine-tuning $\boldsymbol{\theta}$ and $\vz$. After fine-tuning, we get the restored image $\vx^{*}=G(\vz^{*} ; \boldsymbol{\theta}^{*})$.

Although DGP has achieved noteworthy results in image restoration and manipulation, it has limitations due to the inflexibility of the pre-trained GAN model. For example, if DGP adopts a $128\times128$ BigGAN model, DGP must first crop the original image into image patch of size $128\times128$ before restoration, restricting its practical application. However, because Omni-INR-GAN can output images of any resolution, combining it with DGP can directly restore the original image.

We use Omni-INR-GAN pre-trained on ImageNet $256\times256$ for colorization and super-resolution. \myeqref{apx:equ:dgp_objective} is the objective. For colorization, $\hat{\vx}$ is a grayscale image, and for super-resolution, $\hat{\vx}$ is a low-resolution image. We resize the input image's short edge to $256$ and keep the aspect ratio of the image unchanged.  After fine-tuning, $\vx^{*}=G(\vz^{*} ; \boldsymbol{\theta}^{*})$ is the restored image. Because $G(\vz^{*}; \boldsymbol{\theta}^{*})$ represents $\vx^{*}$ in the INR form, we can get the restored image at any resolution through $G(\vz^{*}; \boldsymbol{\theta}^{*})$. Therefore, Omni-INR-GAN is more friendly to downstream tasks.


\subsection{Reconstruction}

We compare pre-trained GAN models for image reconstruction tasks. Specifically, we finetune the parameters of the generator to make it reconstruct given images. Note that we do not use mse or L1 loss, because these loss functions make it easy for the generator to overfit the given image, as long as the training iterations are enough. Instead, we only use the discriminator feature loss, because it has been proven to be very effective for utilizing the prior of the generator. For the dataset, we use 1k images sampled from the ImageNet validation set, which is the same as DGP's choice. Note that these data have not been used in GAN's training. We adopt the progressive reconstruction strategy of DGP~\cite{pan2020Exploiting}, and finetune each GAN model for the same number of iterations.

\section{Implementation Details}

We adopt BigGAN architectures of $128\times128$ and $256\times256$ in our experiments. Table~\ref{apx:tab:omnigan_imagenet128}, \ref{apx:tab:omnigan_imagenet256}, \ref{apx:tab:omniinrgan_imagenet128}, and \ref{apx:tab:omniinrgan_imagenet256} show the architectural details. Each experiment is conducted on eight v100 GPUs. Training Omni-GAN on ImageNet $128\times128$ and $256\times256$ took $25$ days and $60$ days, respectively. Training Omni-INR-GAN on ImageNet $128\times128$ and $256\times256$ took $27$ days and $87$ days, respectively. No collapse occurred during the entire training process. We have found experimentally that classification-based cGANs cannot set a large batch size like projection-based BigGAN. For all experiments of Omni-GAN and Omni-INR-GAN, the batch size is set to $256$. We adopt the ADAM optimizer in all experiments, with betas being $0$ and $0.999$. The learning rates of the generator and discriminator are set to $0.0001$ and $0.0004$, respectively.

For $128\times128$ experiments, the generator and discriminator use non-local block at $64\times64$ resolution. The generator is updated once every time the discriminator is updated. The weight decay of the generator and discriminator are set to $0.001$ and $0.00001$, respectively. For Omni-INR-GAN, we removed the non-local block at $64\times64$ resolution of the discriminator. Because when Omni-INR-GAN is used for downstream tasks, the input image of the discriminator may be of any size, so the middle layer of the discriminator may not output $64\times64$ resolution features. Moreover, although Omni-INR-GAN can generate images of any resolution, we did not adopt a multi-scale training strategy. We found that multi-scale training led to training collapse. We think that a possible reason is that multi-scale training enhances the discriminator, resulting in the ability of the generator and the discriminator to be out of balance. Thus we only generate $128\times128$ images during training, and the real images are also resized to $128\times128$.

For $256\times256$ experiments, the weight decay of the generator and discriminator are set to $0.0001$ and $0.00001$, respectively. The generator is updated once every time the discriminator is updated twice. We have found experimentally that this will make training more stable. The generator and discriminator use non-local block at $64\times64$ resolution rather than $128\times128$ due to limited GPU memory. For Omni-INR-GAN, in order to support downstream tasks friendly, we do not use non-local block in the discriminator. Moreover, due to GPU memory limitation, we reduce the batch size to $128$ and accumulate the gradient twice to approximate the gradient when the batch size is $256$. We did not adopt a multi-scale training strategy. Only $256\times256$ images are generated during training, and the real images are also resized to $256\times256$.

\begin{table*}[htbp]
  \begin{subtable}[t]{0.48\linewidth}
    \centering
    \resizebox{0.8\linewidth}{!}{%
      \centering
      \begin{tabular}{c}
        \toprule
        $\vz\in \mathbb{R}^{120} \sim \mathcal{N}(0, I), \text{embed}(y) \in \mathbb{R}^{128}$ \\
        \midrule
        Linear $20$ $\rightarrow$ $4\times4\times16 ch$                                        \\         \midrule
        ResBlock up $16ch \rightarrow 16ch$                                                    \\         \midrule
        ResBlock up $16ch \rightarrow 8ch$                                                     \\         \midrule
        ResBlock up $8ch \rightarrow 4ch$                                                      \\         \midrule
        ResBlock up $4ch \rightarrow 2ch$                                                      \\         \midrule
        Non-local Block  ($64 \times 64$)                                                      \\         \midrule
        ResBlock up $2ch \rightarrow ch$                                                       \\         \midrule
        BN, ReLU, $3\times3$ Conv $ch \rightarrow 3$                                           \\         \midrule
        Tanh                                                                                   \\
        \bottomrule
      \end{tabular}%
    }
    \caption{Generator}
  \end{subtable}
  \centering
  \begin{subtable}[t]{0.48\linewidth}
    \centering
    \resizebox{.6\linewidth}{!}{%
      \begin{tabular}{c}
        \toprule
        RGB image $\vx \in \mathbb{R}^{128 \times 128 \times 3}$ \\ \midrule
        ResBlock down $3 \rightarrow ch$                         \\ \midrule
        Non-local Block ($64 \times 64$)                         \\ \midrule
        ResBlock down $ch \rightarrow 2ch$                       \\ \midrule
        ResBlock down $2ch \rightarrow 4ch$                      \\ \midrule
        ResBlock down $4ch \rightarrow 8ch$                      \\ \midrule
        ResBlock down $8ch \rightarrow 16ch$                     \\ \midrule
        ResBlock $16ch \rightarrow 16ch$                         \\ \midrule
        ReLU, global sum pooling                                 \\ \midrule
        Linear $\rightarrow 1002$                                \\
        \bottomrule
      \end{tabular}%
    }
    \caption{Discriminator}
  \end{subtable}
  \caption{Omni-GAN architecture on ImageNet $128\times128$. $ch$ is set to be $96$.}
  \label{apx:tab:omnigan_imagenet128}
\end{table*}

\begin{table*}[htbp]
  \begin{subtable}[t]{0.48\linewidth}
    \centering
    \resizebox{0.8\linewidth}{!}{%
      \centering
      \begin{tabular}{c}
        \toprule
        $\vz\in \mathbb{R}^{120} \sim \mathcal{N}(0, I), \text{embed}(y) \in \mathbb{R}^{128}$ \\
        \midrule
        Linear $17$ $\rightarrow$ $4\times4\times16 ch$                                        \\         \midrule
        ResBlock up $16ch \rightarrow 16ch$                                                    \\         \midrule
        ResBlock up $16ch \rightarrow 8ch$                                                     \\         \midrule
        ResBlock up $8ch \rightarrow 8ch$                                                      \\         \midrule
        ResBlock up $8ch \rightarrow 4ch$                                                      \\         \midrule
        Non-local Block  ($64 \times 64$)                                                      \\         \midrule
        ResBlock up $4ch \rightarrow 2ch$                                                      \\         \midrule
        ResBlock up $2ch \rightarrow ch$                                                       \\         \midrule
        BN, ReLU, $3\times3$ Conv $ch \rightarrow 3$                                           \\         \midrule
        Tanh                                                                                   \\
        \bottomrule
      \end{tabular}%
    }
    \caption{Generator}
  \end{subtable}
  \centering
  \begin{subtable}[t]{0.48\linewidth}
    \centering
    \resizebox{.6\linewidth}{!}{%
      \begin{tabular}{c}
        \toprule
        RGB image $\vx \in \mathbb{R}^{256 \times 256 \times 3}$ \\ \midrule
        ResBlock down $3 \rightarrow ch$                         \\ \midrule
        ResBlock down $ch \rightarrow 2ch$                       \\ \midrule
        Non-local Block ($64 \times 64$)                         \\ \midrule
        ResBlock down $2ch \rightarrow 4ch$                      \\ \midrule
        ResBlock down $4ch \rightarrow 8ch$                      \\ \midrule
        ResBlock down $8ch \rightarrow 8ch$                      \\ \midrule
        ResBlock down $8ch \rightarrow 16ch$                     \\ \midrule
        ResBlock $16ch \rightarrow 16ch$                         \\ \midrule
        ReLU, global sum pooling                                 \\ \midrule
        Linear $\rightarrow 1002$                                \\
        \bottomrule
      \end{tabular}%
    }
    \caption{Discriminator}
  \end{subtable}
  \caption{Omni-GAN architecture on Imagenet $256\times256$. $ch$ is set to be $96$.}
  \label{apx:tab:omnigan_imagenet256}
\end{table*}

\begin{table*}[htbp]
  \begin{subtable}[t]{0.48\linewidth}
    \centering
    \resizebox{0.8\linewidth}{!}{%
      \centering
      \begin{tabular}{c}
        \toprule
        $\vz\in \mathbb{R}^{120} \sim \mathcal{N}(0, I), \text{embed}(y) \in \mathbb{R}^{128}$ \\
        \midrule
        Linear $20$ $\rightarrow$ $4\times4\times16 ch$                                        \\         \midrule
        ResBlock up $16ch \rightarrow 16ch$                                                    \\         \midrule
        ResBlock up $16ch \rightarrow 8ch$                                                     \\         \midrule
        ResBlock up $8ch \rightarrow 4ch$                                                      \\         \midrule
        ResBlock up $4ch \rightarrow 2ch$                                                      \\         \midrule
        Non-local Block  ($64 \times 64$)                                                      \\         \midrule
        ResBlock up $2ch \rightarrow ch$                                                       \\         \midrule
        Unfold(kernel\_size=3) $ch \rightarrow 9ch$                                            \\         \midrule
        Grid\_sample($x, y$), Concat feature and $(x, y)$                                      \\         \midrule
        Linear, Relu $9ch + 2 \rightarrow ch$                                                  \\         \midrule
        Linear, Relu $ch \rightarrow ch$                                                       \\         \midrule
        Linear $ch \rightarrow 3$                                                              \\         \midrule
        Tanh                                                                                   \\
        \bottomrule
      \end{tabular}%
    }
    \caption{Generator}
  \end{subtable}
  \centering
  \begin{subtable}[t]{0.48\linewidth}
    \centering
    \resizebox{.6\linewidth}{!}{%
      \begin{tabular}{c}
        \toprule
        RGB image $\vx \in \mathbb{R}^{128 \times 128 \times 3}$ \\ \midrule
        ResBlock down $3 \rightarrow ch$                         \\ \midrule
        ResBlock down $ch \rightarrow 2ch$                       \\ \midrule
        ResBlock down $2ch \rightarrow 4ch$                      \\ \midrule
        ResBlock down $4ch \rightarrow 8ch$                      \\ \midrule
        ResBlock down $8ch \rightarrow 16ch$                     \\ \midrule
        ResBlock $16ch \rightarrow 16ch$                         \\ \midrule
        ReLU, global sum pooling                                 \\ \midrule
        Linear $\rightarrow 1002$                                \\
        \bottomrule
      \end{tabular}%
    }
    \caption{Discriminator}
  \end{subtable}
  \caption{Omni-INR-GAN architecture on ImageNet $128\times128$. $ch$ is set to be $96$.}
  \label{apx:tab:omniinrgan_imagenet128}
\end{table*}

\begin{table*}[htbp]
  \begin{subtable}[t]{0.48\linewidth}
    \centering
    \resizebox{0.8\linewidth}{!}{%
      \centering
      \begin{tabular}{c}
        \toprule
        $\vz\in \mathbb{R}^{120} \sim \mathcal{N}(0, I), \text{embed}(y) \in \mathbb{R}^{128}$ \\
        \midrule
        Linear $17$ $\rightarrow$ $4\times4\times16 ch$                                        \\         \midrule
        ResBlock up $16ch \rightarrow 16ch$                                                    \\         \midrule
        ResBlock up $16ch \rightarrow 8ch$                                                     \\         \midrule
        ResBlock up $8ch \rightarrow 8ch$                                                      \\         \midrule
        ResBlock up $8ch \rightarrow 4ch$                                                      \\         \midrule
        Non-local Block  ($64 \times 64$)                                                      \\         \midrule
        ResBlock up $4ch \rightarrow 2ch$                                                      \\         \midrule
        ResBlock up $2ch \rightarrow ch$                                                       \\         \midrule
        Unfold(kernel\_size=3) $ch \rightarrow 9ch$                                            \\         \midrule
        Grid\_sample($x, y$), Concat feature and $(x, y)$                                      \\         \midrule
        Linear, Relu $9ch + 2 \rightarrow ch$                                                  \\         \midrule
        Linear, Relu $ch \rightarrow ch$                                                       \\         \midrule
        Linear $ch \rightarrow 3$                                                              \\         \midrule
        Tanh                                                                                   \\
        \bottomrule
      \end{tabular}%
    }
    \caption{Generator}
  \end{subtable}
  \centering
  \begin{subtable}[t]{0.48\linewidth}
    \centering
    \resizebox{.6\linewidth}{!}{%
      \begin{tabular}{c}
        \toprule
        RGB image $\vx \in \mathbb{R}^{256 \times 256 \times 3}$ \\ \midrule
        ResBlock down $3 \rightarrow ch$                         \\ \midrule
        ResBlock down $ch \rightarrow 2ch$                       \\ \midrule
        ResBlock down $2ch \rightarrow 4ch$                      \\ \midrule
        ResBlock down $4ch \rightarrow 8ch$                      \\ \midrule
        ResBlock down $8ch \rightarrow 8ch$                      \\ \midrule
        ResBlock down $8ch \rightarrow 16ch$                     \\ \midrule
        ResBlock $16ch \rightarrow 16ch$                         \\ \midrule
        ReLU, global sum pooling                                 \\ \midrule
        Linear $\rightarrow 1002$                                \\
        \bottomrule
      \end{tabular}%
    }
    \caption{Discriminator}
  \end{subtable}
  \caption{Omni-INR-GAN architecture on Imagenet $256\times256$. $ch$ is set to be $96$.}
  \label{apx:tab:omniinrgan_imagenet256}
\end{table*}



\section{Additional Results}

\subsection{Generated Images on CIFAR}

\begin{figure}[t]
  \centering
  \includegraphics[width=\linewidth]{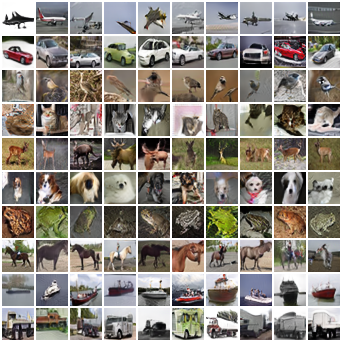}
  \caption{Randomly generated image by Omni-GAN for CIFAR10}
  \label{apx:fig:cifar10}
\end{figure}

\begin{figure}[t]
  \centering
  \includegraphics[width=\linewidth]{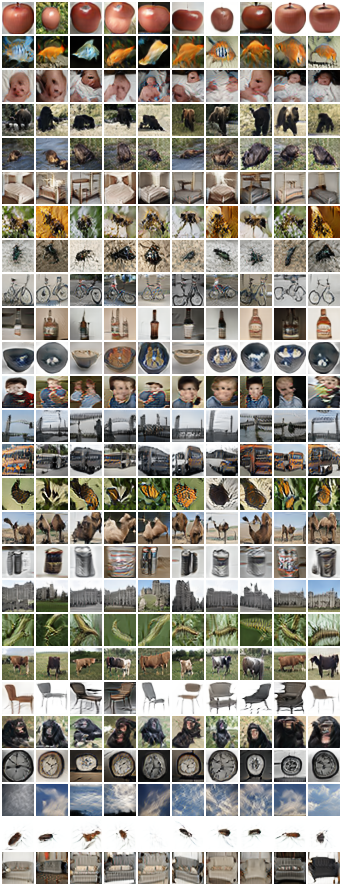}
  \caption{Randomly generated image by Omni-GAN for CIFAR100}
  \label{apx:fig:cifar100}
\end{figure}

In Fig.~\ref{apx:fig:cifar10} and \ref{apx:fig:cifar100}, we show generated images from Omni-GAN on CIFAR10, CIFAR100 respectively. Due to limited space, we only show images of some categories on CIFAR100.

\subsection{Generated Images on ImageNet}

Omni-INR-GAN inherently supports generating images of arbitrary resolution. We adopt the Omni-INR-GAN $256\times256$ model to generate some images with different resolutions, \eg, Fig.~\ref{apx:fig:omniinrgan_sample_c1}, ~\ref{apx:fig:omniinrgan_sample_c10}, etc.


\begin{figure*}[t]
  \footnotesize
  \centering
  \renewcommand{\tabcolsep}{1pt} \renewcommand{\arraystretch}{0.8}
  \graphicspath{{figures/results/OmniINGAN_samples/class_0001_20210324_113415_932/}}
  \resizebox{0.7\linewidth}{!}{%
    \begin{tabular}{cccc}
      \includegraphics[width=0.25\linewidth]{0032.jpg}                    &
      \includegraphics[width=0.25\linewidth]{0064.jpg}                    &
      \includegraphics[width=0.25\linewidth]{0128.jpg}                    &
      \includegraphics[width=0.25\linewidth]{0256.jpg}                                                                                             \\
      $32\times32$                                                        & $64\times64$                         & $128\times128$ & $256\times256$ \\
      \multicolumn{2}{c}{\includegraphics[width=0.5\linewidth]{0512.jpg}} &
      \multicolumn{2}{c}{\includegraphics[width=0.5\linewidth]{1024.jpg}}                                                                          \\
      \multicolumn{2}{c}{$512\times512$}                                  & \multicolumn{2}{c}{$1024\times1024$}
    \end{tabular}
  }
  \vspace{-5pt}
  \caption{Samples generated by our Omni-INR-GAN $256\times256$ model. Omni-INR-GAN has the ability to generate images of any resolution.}
  \label{apx:fig:omniinrgan_sample_c1}
  \vspace{-5pt}
\end{figure*}

\begin{figure*}[t]
  \footnotesize
  \centering
  \renewcommand{\tabcolsep}{1pt} \renewcommand{\arraystretch}{0.8}
  \graphicspath{{figures/results/OmniINGAN_samples/class_0010_20210324_114434_254/}}
  \resizebox{0.7\linewidth}{!}{%
    \begin{tabular}{cccc}
      \includegraphics[width=0.25\linewidth]{0032.jpg}                    &
      \includegraphics[width=0.25\linewidth]{0064.jpg}                    &
      \includegraphics[width=0.25\linewidth]{0128.jpg}                    &
      \includegraphics[width=0.25\linewidth]{0256.jpg}                                                                                             \\
      $32\times32$                                                        & $64\times64$                         & $128\times128$ & $256\times256$ \\
      \multicolumn{2}{c}{\includegraphics[width=0.5\linewidth]{0512.jpg}} &
      \multicolumn{2}{c}{\includegraphics[width=0.5\linewidth]{1024.jpg}}                                                                          \\
      \multicolumn{2}{c}{$512\times512$}                                  & \multicolumn{2}{c}{$1024\times1024$}
    \end{tabular}
  }
  \vspace{-5pt}
  \caption{Samples generated by our Omni-INR-GAN $256\times256$ model. Omni-INR-GAN has the ability to generate images of any resolution.}
  \label{apx:fig:omniinrgan_sample_c10}
  \vspace{-5pt}
\end{figure*}

\begin{figure*}[t]
  \footnotesize
  \centering
  \renewcommand{\tabcolsep}{1pt} \renewcommand{\arraystretch}{0.8}
  \graphicspath{{figures/results/OmniINGAN_samples/class_0011_20210324_114622_559/}}
  \resizebox{0.7\linewidth}{!}{%
    \begin{tabular}{cccc}
      \includegraphics[width=0.25\linewidth]{0032.jpg}                    &
      \includegraphics[width=0.25\linewidth]{0064.jpg}                    &
      \includegraphics[width=0.25\linewidth]{0128.jpg}                    &
      \includegraphics[width=0.25\linewidth]{0256.jpg}                                                                                             \\
      $32\times32$                                                        & $64\times64$                         & $128\times128$ & $256\times256$ \\
      \multicolumn{2}{c}{\includegraphics[width=0.5\linewidth]{0512.jpg}} &
      \multicolumn{2}{c}{\includegraphics[width=0.5\linewidth]{1024.jpg}}                                                                          \\
      \multicolumn{2}{c}{$512\times512$}                                  & \multicolumn{2}{c}{$1024\times1024$}
    \end{tabular}
  }
  \vspace{-5pt}
  \caption{Samples generated by our Omni-INR-GAN $256\times256$ model. Omni-INR-GAN has the ability to generate images of any resolution.}
  \vspace{-5pt}
\end{figure*}

\begin{figure*}[t]
  \footnotesize
  \centering
  \renewcommand{\tabcolsep}{1pt} \renewcommand{\arraystretch}{0.8}
  \graphicspath{{figures/results/OmniINGAN_samples/class_0014_20210324_114836_556/}}
  \resizebox{0.7\linewidth}{!}{%
    \begin{tabular}{cccc}
      \includegraphics[width=0.25\linewidth]{0032.jpg}                    &
      \includegraphics[width=0.25\linewidth]{0064.jpg}                    &
      \includegraphics[width=0.25\linewidth]{0128.jpg}                    &
      \includegraphics[width=0.25\linewidth]{0256.jpg}                                                                                             \\
      $32\times32$                                                        & $64\times64$                         & $128\times128$ & $256\times256$ \\
      \multicolumn{2}{c}{\includegraphics[width=0.5\linewidth]{0512.jpg}} &
      \multicolumn{2}{c}{\includegraphics[width=0.5\linewidth]{1024.jpg}}                                                                          \\
      \multicolumn{2}{c}{$512\times512$}                                  & \multicolumn{2}{c}{$1024\times1024$}
    \end{tabular}
  }
  \vspace{-5pt}
  \caption{Samples generated by our Omni-INR-GAN $256\times256$ model. Omni-INR-GAN has the ability to generate images of any resolution.}
  \vspace{-5pt}
\end{figure*}

\begin{figure*}[t]
  \footnotesize
  \centering
  \renewcommand{\tabcolsep}{1pt} \renewcommand{\arraystretch}{0.8}
  \graphicspath{{figures/results/OmniINGAN_samples/class_0143_20210324_142740_544/}}
  \resizebox{0.7\linewidth}{!}{%
    \begin{tabular}{cccc}
      \includegraphics[width=0.25\linewidth]{0032.jpg}                    &
      \includegraphics[width=0.25\linewidth]{0064.jpg}                    &
      \includegraphics[width=0.25\linewidth]{0128.jpg}                    &
      \includegraphics[width=0.25\linewidth]{0256.jpg}                                                                                             \\
      $32\times32$                                                        & $64\times64$                         & $128\times128$ & $256\times256$ \\
      \multicolumn{2}{c}{\includegraphics[width=0.5\linewidth]{0512.jpg}} &
      \multicolumn{2}{c}{\includegraphics[width=0.5\linewidth]{1024.jpg}}                                                                          \\
      \multicolumn{2}{c}{$512\times512$}                                  & \multicolumn{2}{c}{$1024\times1024$}
    \end{tabular}
  }
  \vspace{-5pt}
  \caption{Samples generated by our Omni-INR-GAN $256\times256$ model. Omni-INR-GAN has the ability to generate images of any resolution.}
  \vspace{-5pt}
\end{figure*}

\begin{figure*}[t]
  \footnotesize
  \centering
  \renewcommand{\tabcolsep}{1pt} \renewcommand{\arraystretch}{0.8}
  \graphicspath{{figures/results/OmniINGAN_samples/class_0153_20210324_131040_858/}}
  \resizebox{0.7\linewidth}{!}{%
    \begin{tabular}{cccc}
      \includegraphics[width=0.25\linewidth]{0032.jpg}                    &
      \includegraphics[width=0.25\linewidth]{0064.jpg}                    &
      \includegraphics[width=0.25\linewidth]{0128.jpg}                    &
      \includegraphics[width=0.25\linewidth]{0256.jpg}                                                                                             \\
      $32\times32$                                                        & $64\times64$                         & $128\times128$ & $256\times256$ \\
      \multicolumn{2}{c}{\includegraphics[width=0.5\linewidth]{0512.jpg}} &
      \multicolumn{2}{c}{\includegraphics[width=0.5\linewidth]{1024.jpg}}                                                                          \\
      \multicolumn{2}{c}{$512\times512$}                                  & \multicolumn{2}{c}{$1024\times1024$}
    \end{tabular}
  }
  \vspace{-5pt}
  \caption{Samples generated by our Omni-INR-GAN $256\times256$ model. Omni-INR-GAN has the ability to generate images of any resolution.}
  \vspace{-5pt}
\end{figure*}

\begin{figure*}[t]
  \footnotesize
  \centering
  \renewcommand{\tabcolsep}{1pt} \renewcommand{\arraystretch}{0.8}
  \graphicspath{{figures/results/OmniINGAN_samples/class_0156_20210324_131419_171/}}
  \resizebox{0.7\linewidth}{!}{%
    \begin{tabular}{cccc}
      \includegraphics[width=0.25\linewidth]{0032.jpg}                    &
      \includegraphics[width=0.25\linewidth]{0064.jpg}                    &
      \includegraphics[width=0.25\linewidth]{0128.jpg}                    &
      \includegraphics[width=0.25\linewidth]{0256.jpg}                                                                                             \\
      $32\times32$                                                        & $64\times64$                         & $128\times128$ & $256\times256$ \\
      \multicolumn{2}{c}{\includegraphics[width=0.5\linewidth]{0512.jpg}} &
      \multicolumn{2}{c}{\includegraphics[width=0.5\linewidth]{1024.jpg}}                                                                          \\
      \multicolumn{2}{c}{$512\times512$}                                  & \multicolumn{2}{c}{$1024\times1024$}
    \end{tabular}
  }
  \vspace{-5pt}
  \caption{Samples generated by our Omni-INR-GAN $256\times256$ model. Omni-INR-GAN has the ability to generate images of any resolution.}
  \vspace{-5pt}
\end{figure*}

\begin{figure*}[t]
  \footnotesize
  \centering
  \renewcommand{\tabcolsep}{1pt} \renewcommand{\arraystretch}{0.8}
  \graphicspath{{figures/results/OmniINGAN_samples/class_0278_20210324_132057_656/}}
  \resizebox{0.7\linewidth}{!}{%
    \begin{tabular}{cccc}
      \includegraphics[width=0.25\linewidth]{0032.jpg}                    &
      \includegraphics[width=0.25\linewidth]{0064.jpg}                    &
      \includegraphics[width=0.25\linewidth]{0128.jpg}                    &
      \includegraphics[width=0.25\linewidth]{0256.jpg}                                                                                             \\
      $32\times32$                                                        & $64\times64$                         & $128\times128$ & $256\times256$ \\
      \multicolumn{2}{c}{\includegraphics[width=0.5\linewidth]{0512.jpg}} &
      \multicolumn{2}{c}{\includegraphics[width=0.5\linewidth]{1024.jpg}}                                                                          \\
      \multicolumn{2}{c}{$512\times512$}                                  & \multicolumn{2}{c}{$1024\times1024$}
    \end{tabular}
  }
  \vspace{-5pt}
  \caption{Samples generated by our Omni-INR-GAN $256\times256$ model. Omni-INR-GAN has the ability to generate images of any resolution.}
  \vspace{-5pt}
\end{figure*}

\begin{figure*}[t]
  \footnotesize
  \centering
  \renewcommand{\tabcolsep}{1pt} \renewcommand{\arraystretch}{0.8}
  \graphicspath{{figures/results/OmniINGAN_samples/class_0323_20210324_132407_993/}}
  \resizebox{0.7\linewidth}{!}{%
    \begin{tabular}{cccc}
      \includegraphics[width=0.25\linewidth]{0032.jpg}                    &
      \includegraphics[width=0.25\linewidth]{0064.jpg}                    &
      \includegraphics[width=0.25\linewidth]{0128.jpg}                    &
      \includegraphics[width=0.25\linewidth]{0256.jpg}                                                                                             \\
      $32\times32$                                                        & $64\times64$                         & $128\times128$ & $256\times256$ \\
      \multicolumn{2}{c}{\includegraphics[width=0.5\linewidth]{0512.jpg}} &
      \multicolumn{2}{c}{\includegraphics[width=0.5\linewidth]{1024.jpg}}                                                                          \\
      \multicolumn{2}{c}{$512\times512$}                                  & \multicolumn{2}{c}{$1024\times1024$}
    \end{tabular}
  }
  \vspace{-5pt}
  \caption{Samples generated by our Omni-INR-GAN $256\times256$ model. Omni-INR-GAN has the ability to generate images of any resolution.}
  \vspace{-5pt}
\end{figure*}

\begin{figure*}[t]
  \footnotesize
  \centering
  \renewcommand{\tabcolsep}{1pt} \renewcommand{\arraystretch}{0.8}
  \graphicspath{{figures/results/OmniINGAN_samples/class_0326_20210324_132706_954/}}
  \resizebox{0.7\linewidth}{!}{%
    \begin{tabular}{cccc}
      \includegraphics[width=0.25\linewidth]{0032.jpg}                    &
      \includegraphics[width=0.25\linewidth]{0064.jpg}                    &
      \includegraphics[width=0.25\linewidth]{0128.jpg}                    &
      \includegraphics[width=0.25\linewidth]{0256.jpg}                                                                                             \\
      $32\times32$                                                        & $64\times64$                         & $128\times128$ & $256\times256$ \\
      \multicolumn{2}{c}{\includegraphics[width=0.5\linewidth]{0512.jpg}} &
      \multicolumn{2}{c}{\includegraphics[width=0.5\linewidth]{1024.jpg}}                                                                          \\
      \multicolumn{2}{c}{$512\times512$}                                  & \multicolumn{2}{c}{$1024\times1024$}
    \end{tabular}
  }
  \vspace{-5pt}
  \caption{Samples generated by our Omni-INR-GAN $256\times256$ model. Omni-INR-GAN has the ability to generate images of any resolution.}
  \vspace{-5pt}
\end{figure*}

\begin{figure*}[t]
  \footnotesize
  \centering
  \renewcommand{\tabcolsep}{1pt} \renewcommand{\arraystretch}{0.8}
  \graphicspath{{figures/results/OmniINGAN_samples/class_0331_20210324_133147_310/}}
  \resizebox{0.7\linewidth}{!}{%
    \begin{tabular}{cccc}
      \includegraphics[width=0.25\linewidth]{0032.jpg}                    &
      \includegraphics[width=0.25\linewidth]{0064.jpg}                    &
      \includegraphics[width=0.25\linewidth]{0128.jpg}                    &
      \includegraphics[width=0.25\linewidth]{0256.jpg}                                                                                             \\
      $32\times32$                                                        & $64\times64$                         & $128\times128$ & $256\times256$ \\
      \multicolumn{2}{c}{\includegraphics[width=0.5\linewidth]{0512.jpg}} &
      \multicolumn{2}{c}{\includegraphics[width=0.5\linewidth]{1024.jpg}}                                                                          \\
      \multicolumn{2}{c}{$512\times512$}                                  & \multicolumn{2}{c}{$1024\times1024$}
    \end{tabular}
  }
  \vspace{-5pt}
  \caption{Samples generated by our Omni-INR-GAN $256\times256$ model. Omni-INR-GAN has the ability to generate images of any resolution.}
  \vspace{-5pt}
\end{figure*}

\begin{figure*}[t]
  \footnotesize
  \centering
  \renewcommand{\tabcolsep}{1pt} \renewcommand{\arraystretch}{0.8}
  \graphicspath{{figures/results/OmniINGAN_samples/class_0358_20210324_133427_825/}}
  \resizebox{0.7\linewidth}{!}{%
    \begin{tabular}{cccc}
      \includegraphics[width=0.25\linewidth]{0032.jpg}                    &
      \includegraphics[width=0.25\linewidth]{0064.jpg}                    &
      \includegraphics[width=0.25\linewidth]{0128.jpg}                    &
      \includegraphics[width=0.25\linewidth]{0256.jpg}                                                                                             \\
      $32\times32$                                                        & $64\times64$                         & $128\times128$ & $256\times256$ \\
      \multicolumn{2}{c}{\includegraphics[width=0.5\linewidth]{0512.jpg}} &
      \multicolumn{2}{c}{\includegraphics[width=0.5\linewidth]{1024.jpg}}                                                                          \\
      \multicolumn{2}{c}{$512\times512$}                                  & \multicolumn{2}{c}{$1024\times1024$}
    \end{tabular}
  }
  \vspace{-5pt}
  \caption{Samples generated by our Omni-INR-GAN $256\times256$ model. Omni-INR-GAN has the ability to generate images of any resolution.}
  \vspace{-5pt}
\end{figure*}

\begin{figure*}[t]
  \footnotesize
  \centering
  \renewcommand{\tabcolsep}{1pt} \renewcommand{\arraystretch}{0.8}
  \graphicspath{{figures/results/OmniINGAN_samples/class_0425_20210324_140054_456/}}
  \resizebox{0.7\linewidth}{!}{%
    \begin{tabular}{cccc}
      \includegraphics[width=0.25\linewidth]{0032.jpg}                    &
      \includegraphics[width=0.25\linewidth]{0064.jpg}                    &
      \includegraphics[width=0.25\linewidth]{0128.jpg}                    &
      \includegraphics[width=0.25\linewidth]{0256.jpg}                                                                                             \\
      $32\times32$                                                        & $64\times64$                         & $128\times128$ & $256\times256$ \\
      \multicolumn{2}{c}{\includegraphics[width=0.5\linewidth]{0512.jpg}} &
      \multicolumn{2}{c}{\includegraphics[width=0.5\linewidth]{1024.jpg}}                                                                          \\
      \multicolumn{2}{c}{$512\times512$}                                  & \multicolumn{2}{c}{$1024\times1024$}
    \end{tabular}
  }
  \vspace{-5pt}
  \caption{Samples generated by our Omni-INR-GAN $256\times256$ model. Omni-INR-GAN has the ability to generate images of any resolution.}
  \vspace{-5pt}
\end{figure*}

\begin{figure*}[t]
  \footnotesize
  \centering
  \renewcommand{\tabcolsep}{1pt} \renewcommand{\arraystretch}{0.8}
  \graphicspath{{figures/results/OmniINGAN_samples/class_0449_20210324_140507_216/}}
  \resizebox{0.7\linewidth}{!}{%
    \begin{tabular}{cccc}
      \includegraphics[width=0.25\linewidth]{0032.jpg}                    &
      \includegraphics[width=0.25\linewidth]{0064.jpg}                    &
      \includegraphics[width=0.25\linewidth]{0128.jpg}                    &
      \includegraphics[width=0.25\linewidth]{0256.jpg}                                                                                             \\
      $32\times32$                                                        & $64\times64$                         & $128\times128$ & $256\times256$ \\
      \multicolumn{2}{c}{\includegraphics[width=0.5\linewidth]{0512.jpg}} &
      \multicolumn{2}{c}{\includegraphics[width=0.5\linewidth]{1024.jpg}}                                                                          \\
      \multicolumn{2}{c}{$512\times512$}                                  & \multicolumn{2}{c}{$1024\times1024$}
    \end{tabular}
  }
  \vspace{-5pt}
  \caption{Samples generated by our Omni-INR-GAN $256\times256$ model. Omni-INR-GAN has the ability to generate images of any resolution.}
  \vspace{-5pt}
\end{figure*}

\begin{figure*}[t]
  \footnotesize
  \centering
  \renewcommand{\tabcolsep}{1pt} \renewcommand{\arraystretch}{0.8}
  \graphicspath{{figures/results/OmniINGAN_samples/class_0946_20210324_141318_570/}}
  \resizebox{0.7\linewidth}{!}{%
    \begin{tabular}{cccc}
      \includegraphics[width=0.25\linewidth]{0032.jpg}                    &
      \includegraphics[width=0.25\linewidth]{0064.jpg}                    &
      \includegraphics[width=0.25\linewidth]{0128.jpg}                    &
      \includegraphics[width=0.25\linewidth]{0256.jpg}                                                                                             \\
      $32\times32$                                                        & $64\times64$                         & $128\times128$ & $256\times256$ \\
      \multicolumn{2}{c}{\includegraphics[width=0.5\linewidth]{0512.jpg}} &
      \multicolumn{2}{c}{\includegraphics[width=0.5\linewidth]{1024.jpg}}                                                                          \\
      \multicolumn{2}{c}{$512\times512$}                                  & \multicolumn{2}{c}{$1024\times1024$}
    \end{tabular}
  }
  \vspace{-5pt}
  \caption{Samples generated by our Omni-INR-GAN $256\times256$ model. Omni-INR-GAN has the ability to generate images of any resolution.}
  \vspace{-5pt}
\end{figure*}

\begin{figure*}[t]
  \footnotesize
  \centering
  \renewcommand{\tabcolsep}{1pt} \renewcommand{\arraystretch}{0.8}
  \graphicspath{{figures/results/OmniINGAN_samples/class_0970_20210324_141828_735/}}
  \resizebox{0.7\linewidth}{!}{%
    \begin{tabular}{cccc}
      \includegraphics[width=0.25\linewidth]{0032.jpg}                    &
      \includegraphics[width=0.25\linewidth]{0064.jpg}                    &
      \includegraphics[width=0.25\linewidth]{0128.jpg}                    &
      \includegraphics[width=0.25\linewidth]{0256.jpg}                                                                                             \\
      $32\times32$                                                        & $64\times64$                         & $128\times128$ & $256\times256$ \\
      \multicolumn{2}{c}{\includegraphics[width=0.5\linewidth]{0512.jpg}} &
      \multicolumn{2}{c}{\includegraphics[width=0.5\linewidth]{1024.jpg}}                                                                          \\
      \multicolumn{2}{c}{$512\times512$}                                  & \multicolumn{2}{c}{$1024\times1024$}
    \end{tabular}
  }
  \vspace{-5pt}
  \caption{Samples generated by our Omni-INR-GAN $256\times256$ model. Omni-INR-GAN has the ability to generate images of any resolution.}
  \vspace{-5pt}
\end{figure*}

\begin{figure*}[t]
  \footnotesize
  \centering
  \renewcommand{\tabcolsep}{1pt} \renewcommand{\arraystretch}{0.8}
  \graphicspath{{figures/results/OmniINGAN_samples/class_0980_20210324_142034_652/}}
  \resizebox{0.7\linewidth}{!}{%
    \begin{tabular}{cccc}
      \includegraphics[width=0.25\linewidth]{0032.jpg}                    &
      \includegraphics[width=0.25\linewidth]{0064.jpg}                    &
      \includegraphics[width=0.25\linewidth]{0128.jpg}                    &
      \includegraphics[width=0.25\linewidth]{0256.jpg}                                                                                             \\
      $32\times32$                                                        & $64\times64$                         & $128\times128$ & $256\times256$ \\
      \multicolumn{2}{c}{\includegraphics[width=0.5\linewidth]{0512.jpg}} &
      \multicolumn{2}{c}{\includegraphics[width=0.5\linewidth]{1024.jpg}}                                                                          \\
      \multicolumn{2}{c}{$512\times512$}                                  & \multicolumn{2}{c}{$1024\times1024$}
    \end{tabular}
  }
  \vspace{-5pt}
  \caption{Samples generated by our Omni-INR-GAN $256\times256$ model. Omni-INR-GAN has the ability to generate images of any resolution.}
  \vspace{-5pt}
\end{figure*}

\begin{figure*}[t]
  \footnotesize
  \centering
  \renewcommand{\tabcolsep}{1pt} \renewcommand{\arraystretch}{0.8}
  \graphicspath{{figures/results/OmniINGAN_samples/class_0985_20210324_142348_264/}}
  \resizebox{0.7\linewidth}{!}{%
    \begin{tabular}{cccc}
      \includegraphics[width=0.25\linewidth]{0032.jpg}                    &
      \includegraphics[width=0.25\linewidth]{0064.jpg}                    &
      \includegraphics[width=0.25\linewidth]{0128.jpg}                    &
      \includegraphics[width=0.25\linewidth]{0256.jpg}                                                                                             \\
      $32\times32$                                                        & $64\times64$                         & $128\times128$ & $256\times256$ \\
      \multicolumn{2}{c}{\includegraphics[width=0.5\linewidth]{0512.jpg}} &
      \multicolumn{2}{c}{\includegraphics[width=0.5\linewidth]{1024.jpg}}                                                                          \\
      \multicolumn{2}{c}{$512\times512$}                                  & \multicolumn{2}{c}{$1024\times1024$}
    \end{tabular}
  }
  \vspace{-5pt}
  \caption{Samples generated by our Omni-INR-GAN $256\times256$ model. Omni-INR-GAN has the ability to generate images of any resolution.}
  \vspace{-5pt}
\end{figure*}

\begin{figure*}[!t]
  \footnotesize
  \centering
  \renewcommand{\tabcolsep}{1pt} \renewcommand{\arraystretch}{0.4}
  \centering
  \resizebox{\linewidth}{!}{%
    \centering
    \begin{tabular}{cc|ccc}
      \includegraphics[width=.19\linewidth]{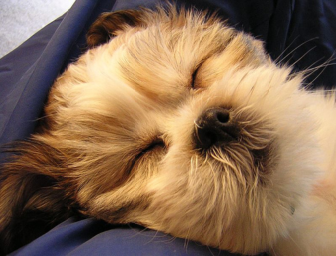}                         &
      \includegraphics[width=.19\linewidth]{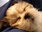}          \hspace{0.01cm} &
      \hspace{0.01cm}
      \includegraphics[width=.19\linewidth]{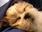}                    &
      \includegraphics[width=.19\linewidth]{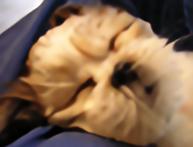}                &
      \includegraphics[width=.19\linewidth]{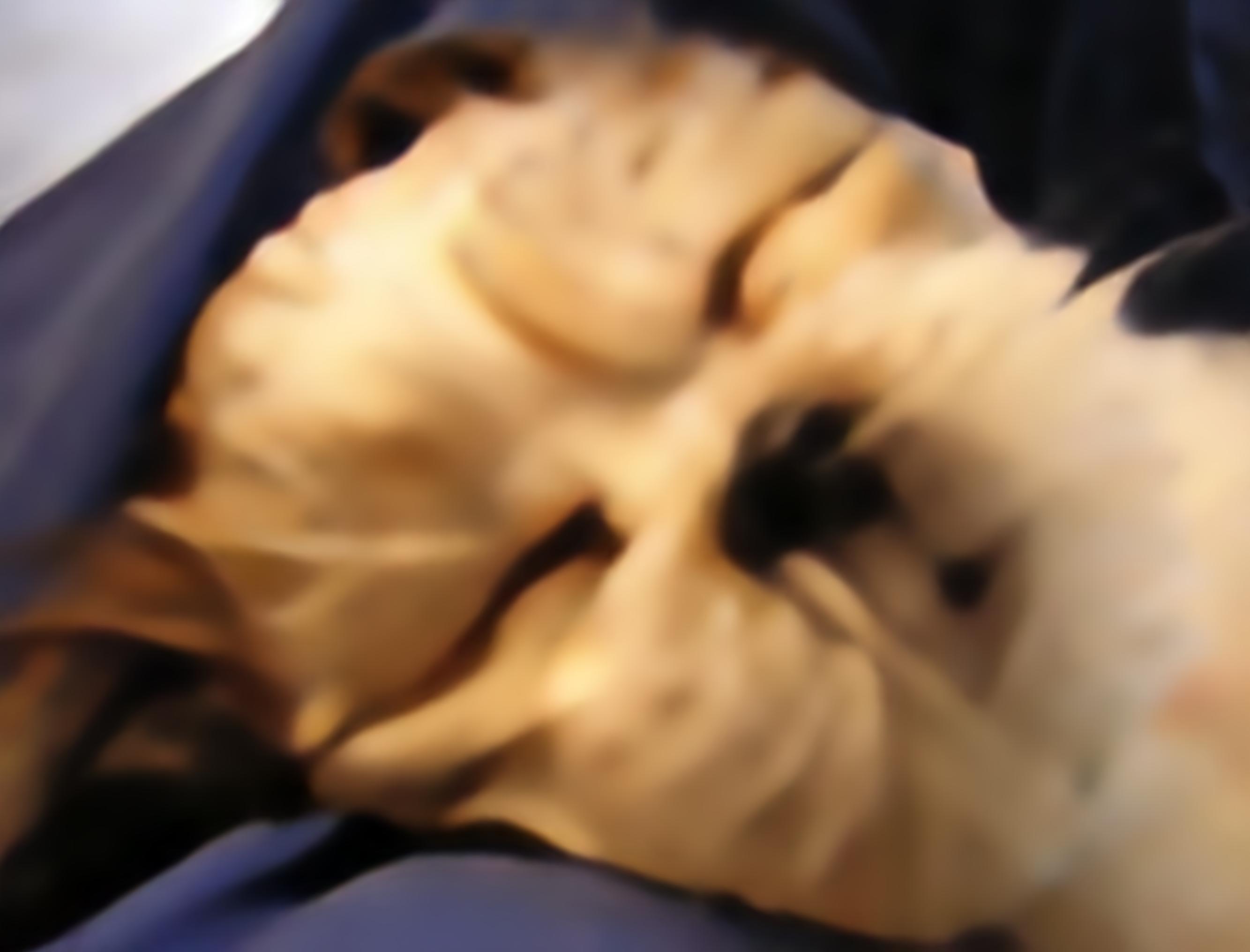}
      \\
      Size ($256\times336$)                                                                                                                       & Size ($32\times42$)          & $\times 1$ ($32\times42$)                          & $\times 4.6$ ($147\times193$) & $\times 63.5$ ($2032\times2667$)
      \\
      (a) Ground truth                                                                                                                            & (b) Input                    & \multicolumn{3}{c}{(c) LIIF}
      \\ \midrule
      \includegraphics[width=.19\linewidth]{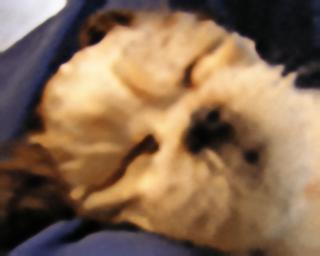}                       &
      \includegraphics[height=2.51cm]{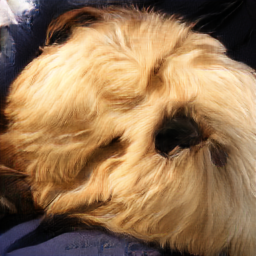}                                                                   &
      \hspace{0.01cm}
      \includegraphics[width=.19\linewidth]{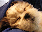}                                         &
      \includegraphics[width=.19\linewidth]{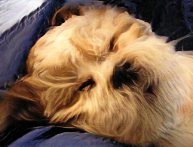}                                       &
      \includegraphics[width=.19\linewidth]{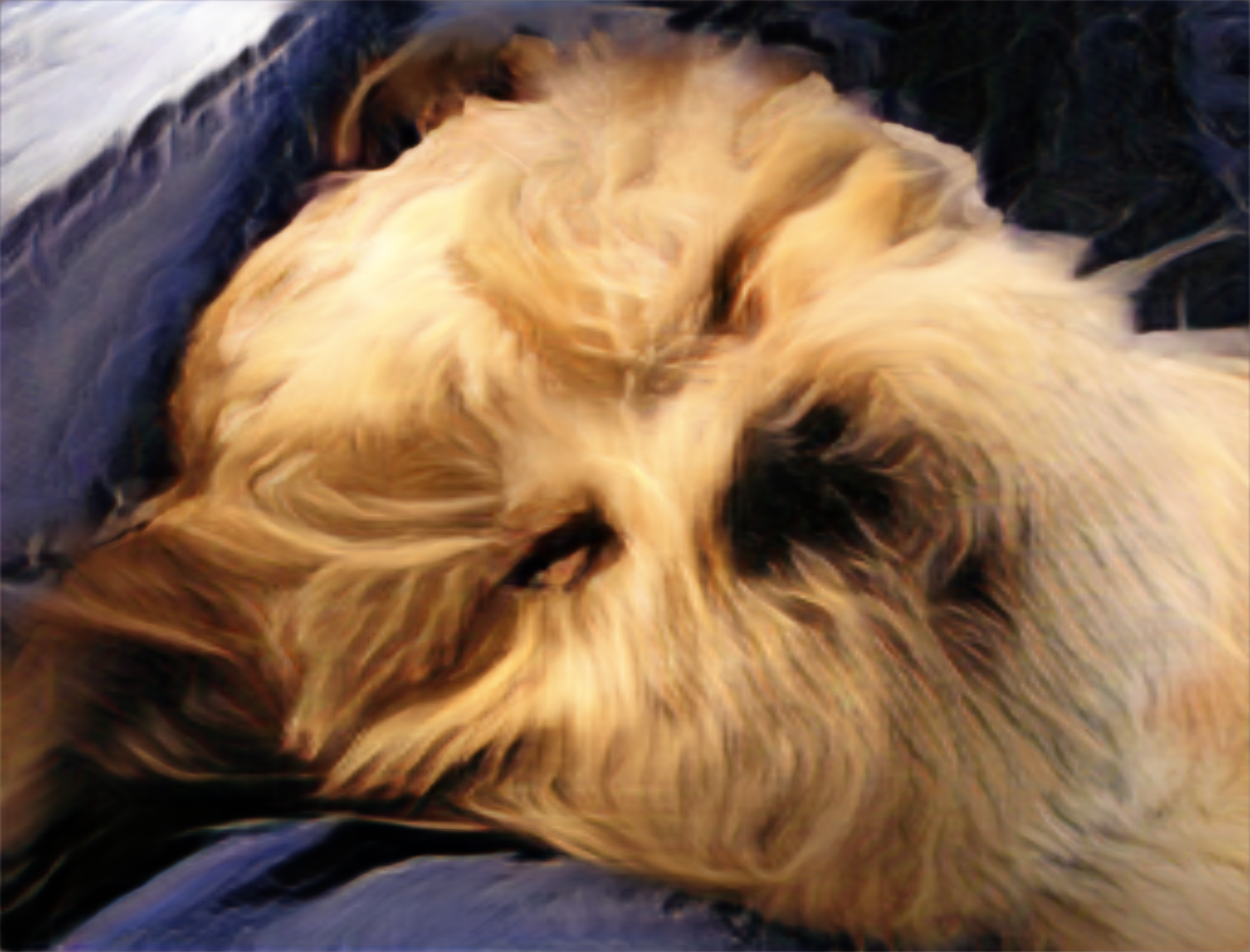}
      \\
      $\times 8$ ($256\times 320$)                                                                                                                & $\times 8$ ($256\times 256$) & $\times 1$ ($32\times42$)                          & $\times 4.6$ ($147\times193$) & $\times 63.5$ ($2032\times2667$)
      \\
      (d) DIP                                                                                                                                     & (e) DGP w/ BigGAN            & \multicolumn{3}{c}{(f) DGP w/ Omni-INR-GAN (ours)}                                                                    \\
      \midrule
    \end{tabular}
  }
  \vspace{-0.3cm}
  \caption{Super-resolution using Omni-INR-GAN's prior, at any scale ($\times1$-$\times60+$). (b) input image with low resolution. (c) LIIF~\cite{chen2020Learning} can extrapolate the input image to any scale, but it cannot add semantic details, so the result is still blurred. (d) DIP~\cite{ulyanov2018Deep} also failed because the input image resolution is too low. (e) DGP~\cite{pan2020Exploiting} with BigGAN must crop the input and upsamples the cropped patch to a fixed size, which is inflexible. (f) Omni-INR-GAN has the ability to upsample the input image to any scale and also adds rich semantic details.}
  \label{apx:fig:dgp_SR}
  \vspace{-.5cm}
\end{figure*}

\subsection{Results of Semantic Image Synthesis}

\begin{figure*}[htbp]
  \footnotesize
  \centering
  \renewcommand{\tabcolsep}{1pt} \renewcommand{\arraystretch}{0.4}
  \begin{tabular}{cccc}
    Label                                                                                                                                & Ground Truth                                                                                                                & SPADE                                                                                                                                      & Ours                                                                                                                                          \\
    \includegraphics[width=0.23\linewidth]{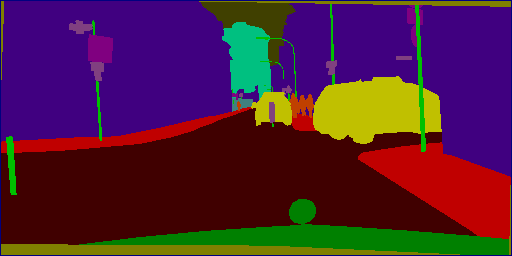} & \includegraphics[width=0.23\linewidth]{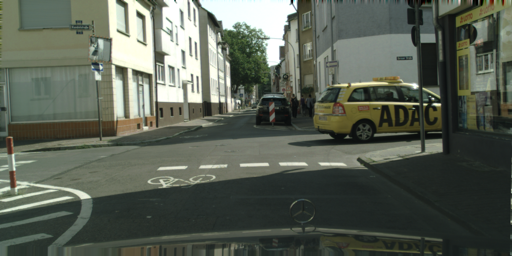} & \includegraphics[width=0.23\linewidth]{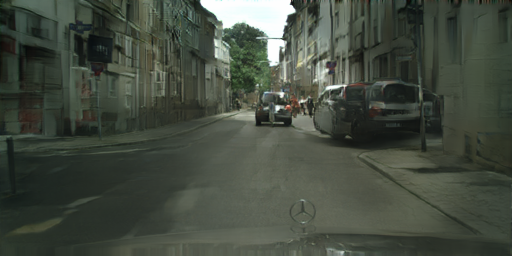} & \includegraphics[width=0.23\linewidth]{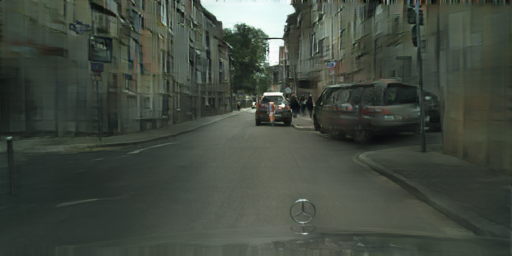} \\
    \includegraphics[width=0.23\linewidth]{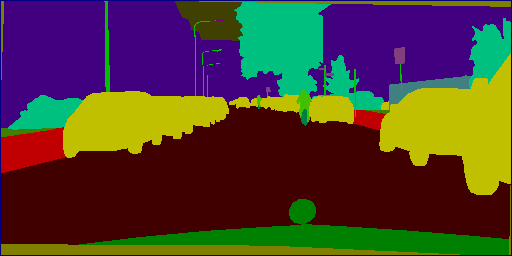} & \includegraphics[width=0.23\linewidth]{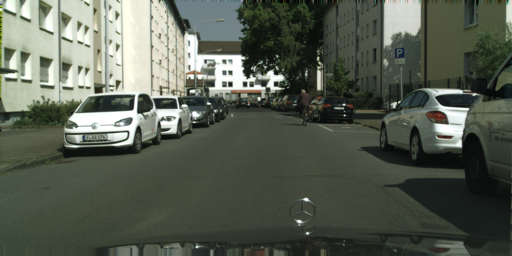} & \includegraphics[width=0.23\linewidth]{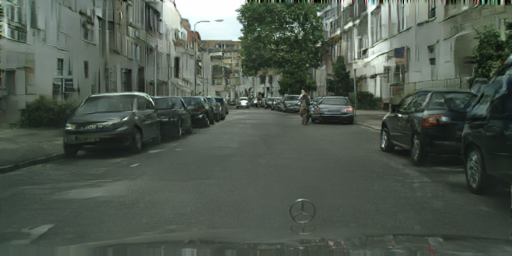} & \includegraphics[width=0.23\linewidth]{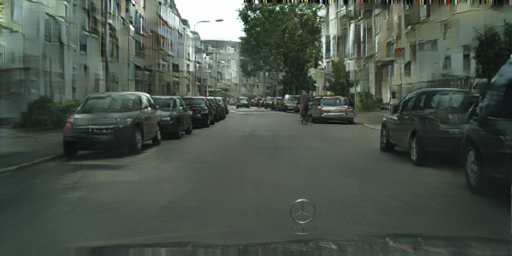} \\
    \includegraphics[width=0.23\linewidth]{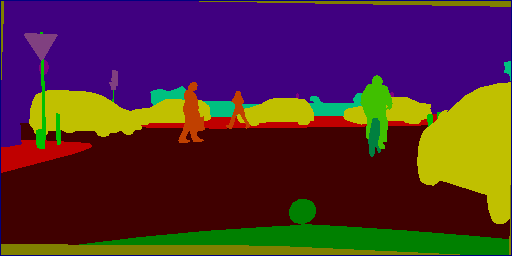} & \includegraphics[width=0.23\linewidth]{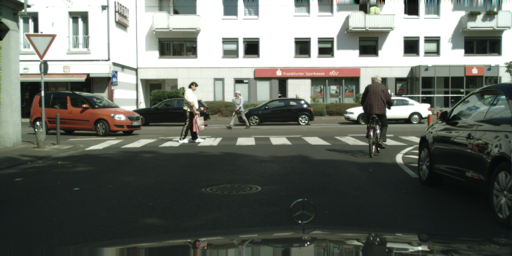} & \includegraphics[width=0.23\linewidth]{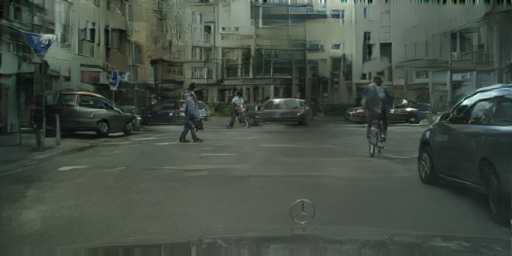} & \includegraphics[width=0.23\linewidth]{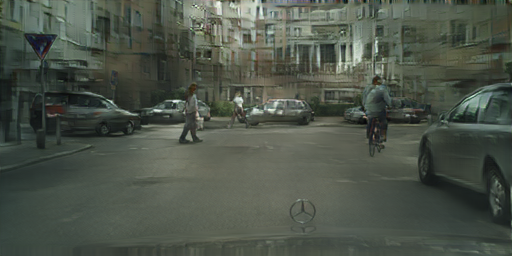} \\
    \includegraphics[width=0.23\linewidth]{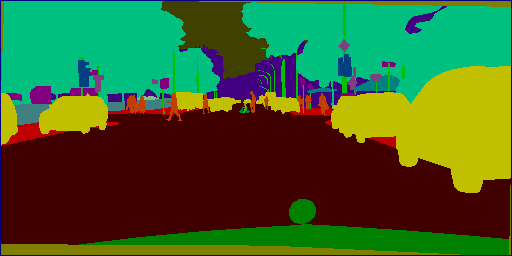} & \includegraphics[width=0.23\linewidth]{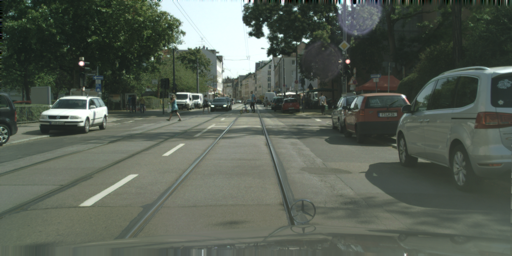} & \includegraphics[width=0.23\linewidth]{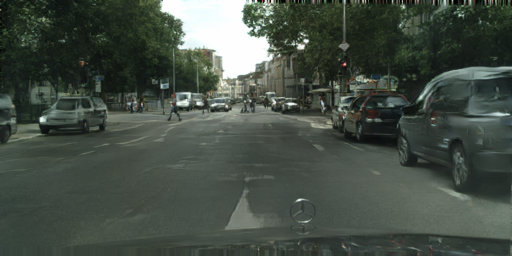} & \includegraphics[width=0.23\linewidth]{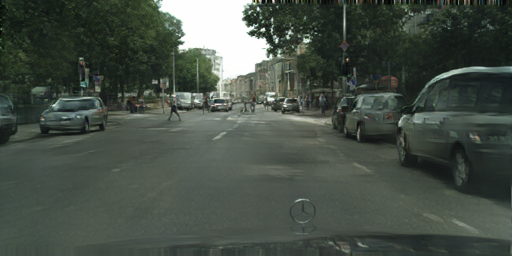} \\
    \includegraphics[width=0.23\linewidth]{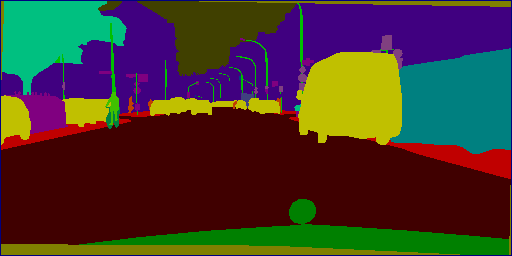} & \includegraphics[width=0.23\linewidth]{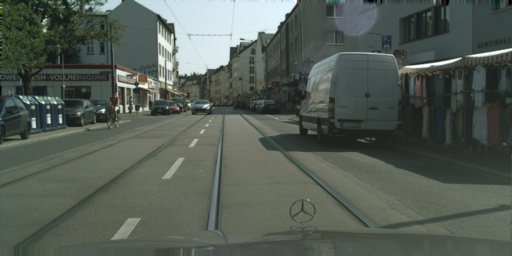} & \includegraphics[width=0.23\linewidth]{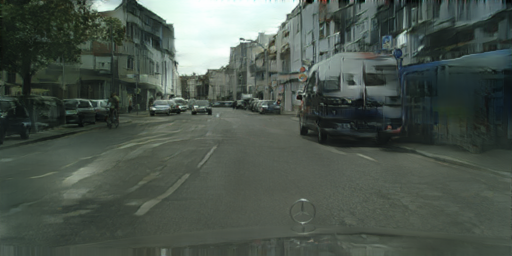} & \includegraphics[width=0.23\linewidth]{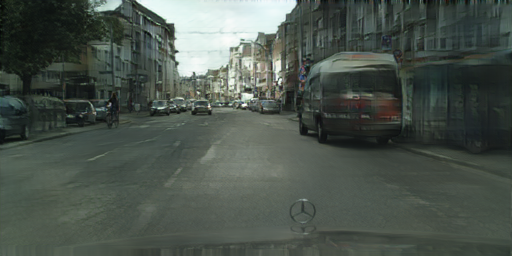} \\
    \includegraphics[width=0.23\linewidth]{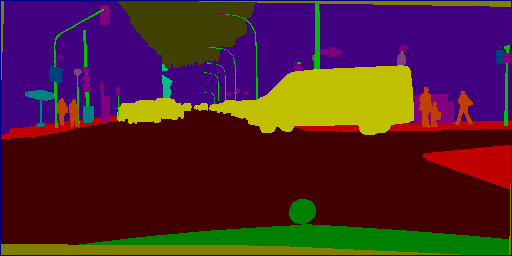} & \includegraphics[width=0.23\linewidth]{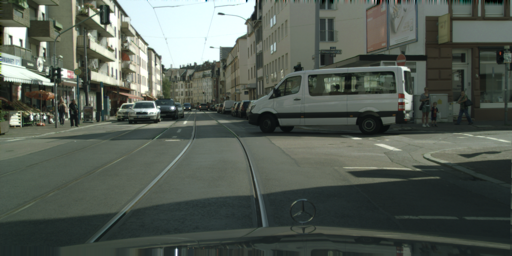} & \includegraphics[width=0.23\linewidth]{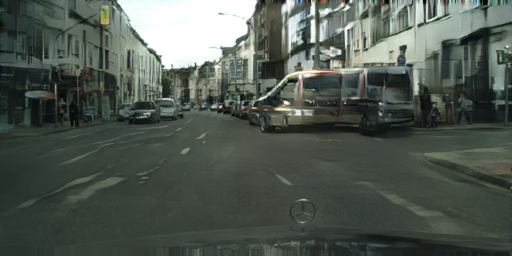} & \includegraphics[width=0.23\linewidth]{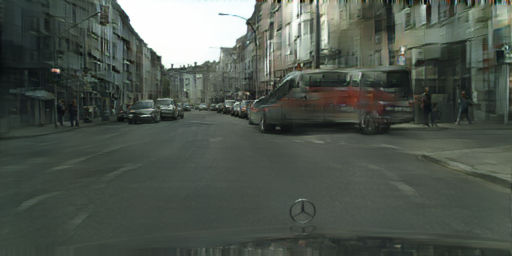} \\
    \includegraphics[width=0.23\linewidth]{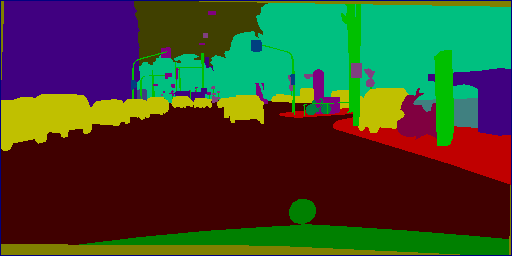} & \includegraphics[width=0.23\linewidth]{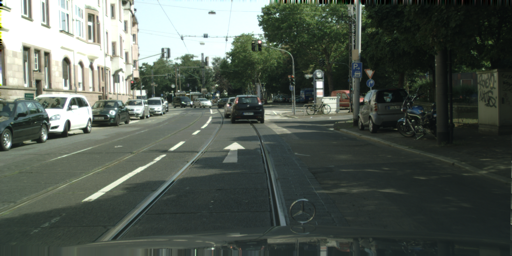} & \includegraphics[width=0.23\linewidth]{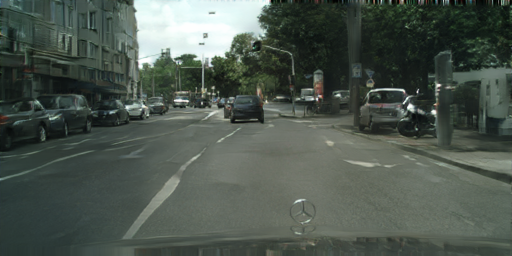} & \includegraphics[width=0.23\linewidth]{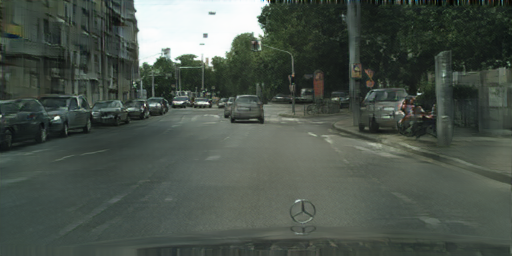} \\
    \includegraphics[width=0.23\linewidth]{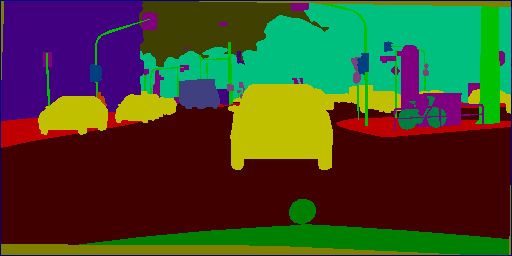} & \includegraphics[width=0.23\linewidth]{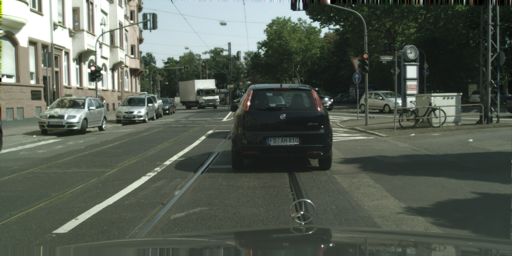} & \includegraphics[width=0.23\linewidth]{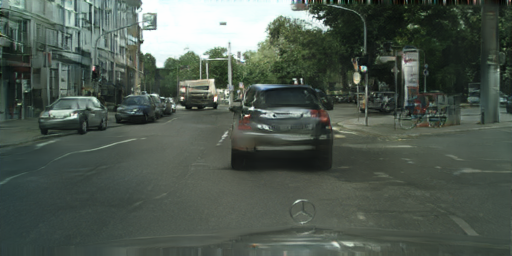} & \includegraphics[width=0.23\linewidth]{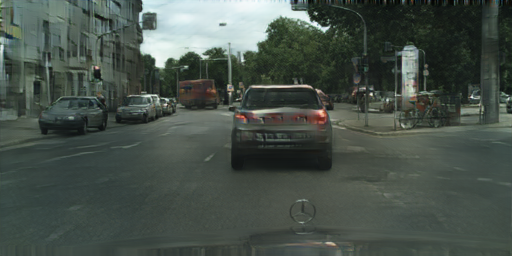} \\
    \includegraphics[width=0.23\linewidth]{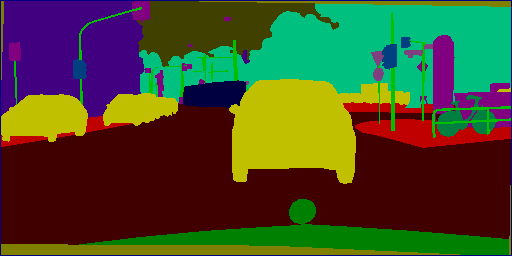} & \includegraphics[width=0.23\linewidth]{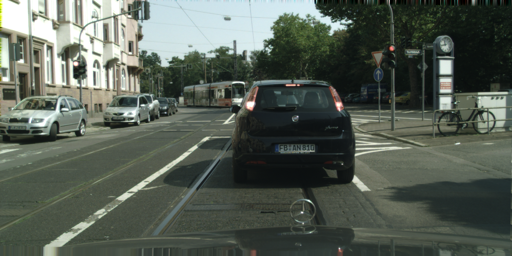} & \includegraphics[width=0.23\linewidth]{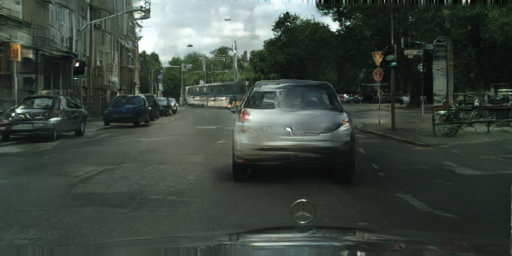} & \includegraphics[width=0.23\linewidth]{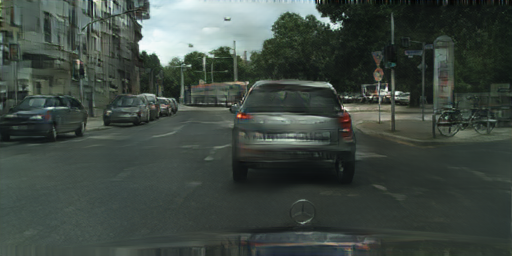} \\
    \includegraphics[width=0.23\linewidth]{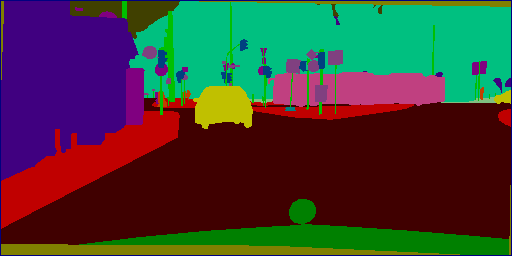} & \includegraphics[width=0.23\linewidth]{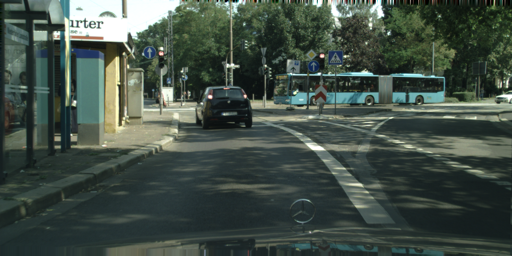} & \includegraphics[width=0.23\linewidth]{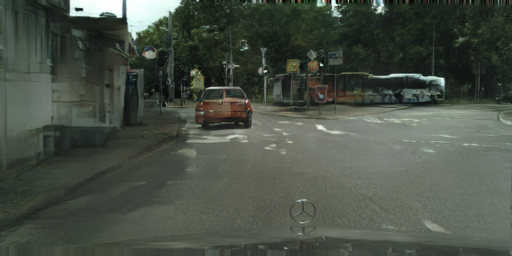} & \includegraphics[width=0.23\linewidth]{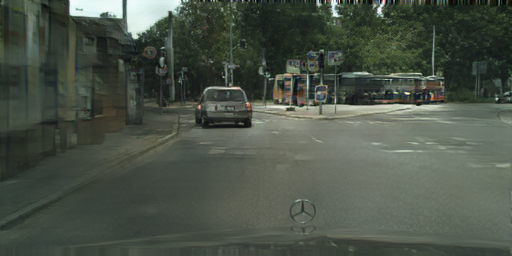} \\
  \end{tabular}
  \caption{Results of semantic image synthesis on Cityscapes.}
  \label{apx:fig:cityscapes}
  \vspace{-5pt}
\end{figure*}

In Fig.~\ref{apx:fig:cityscapes}, we show several results of Omni-GAN as well as those of SPADE for semantic image synthesis. The label maps and the ground truth images are from the first ten items in the test set of Cityscapes dataset, without cherry-picking.

\clearpage

\end{document}